\documentclass[preprint, 11pt]{article} 
\usepackage[utf8]{inputenc}
\usepackage{amsmath,bm}
\usepackage{amssymb}
\usepackage{centernot}
\input{epsf}
\usepackage{natbib}
\usepackage[margin=1in]{geometry}
\usepackage[raggedright]{titlesec}
\usepackage[hang,flushmargin]{footmisc} 
\usepackage{graphicx}
\usepackage{framed}
\usepackage{color}
\usepackage{array}
\usepackage{yfonts}
\usepackage[dvipsnames]{xcolor}
\usepackage{pifont}
\usepackage{amsmath,amssymb}
\usepackage{multirow}
\usepackage{amsfonts}
\usepackage{subcaption}
\usepackage{caption}
\usepackage{threeparttable}
\usepackage{esvect}
\usepackage{pgf}
\usepackage{tikz}
\usepackage{tikz-cd}
\usepackage{fancyvrb}
\usetikzlibrary{positioning}
\usepackage{bbm}
\usepackage{colortbl}
\usepackage{pdflscape}
\usepackage{booktabs}
\usepackage{amsmath}
\usepackage{mathdots}
\usepackage{yhmath}
\usepackage{cancel}
\usepackage{siunitx}
\usepackage{array}
\usepackage{multirow}
\usepackage{gensymb}
\usepackage{tabularx}
\usepackage{booktabs}
\usetikzlibrary{fadings}
\usetikzlibrary{patterns}
\usetikzlibrary{shadows.blur}
\usetikzlibrary{shapes}
\usepackage{placeins}
\usepackage{comment}
\usepackage{accents}
\newcommand{\ubar}[1]{\underaccent{\bar}{#1}}

\usepackage{hyperref}
\hypersetup{
    colorlinks=true,
    linkcolor=NavyBlue,
    filecolor=magenta,      
    urlcolor=cyan,
    citecolor=NavyBlue
}
 
\usepackage{rotating}
\definecolor{lightgray}{gray}{0.9}

\usepackage{theorem}
\theoremstyle{plain}
\theoremheaderfont{\scshape}
\newtheorem{assumption}{Assumption}

\newtheorem{proposition}{Proposition}
\newtheorem{theorem}{Theorem}
\newtheorem{lemma}{Lemma}
\allowdisplaybreaks

\newcommand{\E}{\mathbb{E}}
\newcommand{\bbone}{\mathbbm{1}}

\newcommand{\1}{\mathbbm{1}}

\DeclareMathOperator*{\argmin}{arg\,min}

\newcommand{\indep}{\raisebox{0.05em}{\rotatebox[origin=c]{90}{$\models$}}}
\definecolor{shadecolor}{gray}{0.9}
\newcommand{\qed}{\hfill \ensuremath{\Box}}

\tikzset{every picture/.style={line width=0.75pt}} 

\usepackage{enumitem}
\newlist{Step}{enumerate}{2}
\setlist[Step]{label={{Step \arabic*.}}, leftmargin=*}

\newcommand{\emp}[1]{{\textcolor{blue}{#1}}}

\usepackage{setspace}

\newcommand\spacingset[1]{\renewcommand{\baselinestretch}%
  {#1}\small\normalsize}

 \makeatletter
\newcommand\circled[1]{%
  \mathpalette\@circled{#1}%
}
\newcommand\@circled[2]{%
  \tikz[baseline=(math.base)] \node[draw,circle,inner sep=2pt] (math) {$\m@th#1#2$};%
}
\makeatother

 \makeatletter
\newcommand\circledblue[1]{%
  \mathpalette\@circledblue{#1}%
}
\newcommand\@circledblue[2]{%
  \tikz[baseline=(math.base)] \node[draw,circle, fill=blue!20, inner sep=2pt] (math) {$\m@th#1#2$};%
 }

\title{{\bf Does AI help humans make better decisions?}\\A
  statistical evaluation framework for experimental
  and observational studies} 
\author{Eli Ben-Michael\thanks{Assistant Professor, Department of Statistics \& Data Science and Heinz College of Information Systems \& Public Policy, Carnegie Mellon University. 4800 Forbes Avenue, Hamburg Hall, Pittsburgh PA 15213.  Email:
    \href{mailto:ebenmichael@cmu.edu}{ebenmichael@cmu.edu} URL: \href{https://ebenmichael.github.io}{ebenmichael.github.io}} \and D. James Greiner\thanks{Honorable S.  William Green Professor of Public Law, Harvard Law School, 1525 Massachusetts Avenue, Griswold 504, Cambridge, MA 02138.}  \and Melody Huang\thanks{Postdoctoral Researcher, Harvard University, Cambridge, MA 02138. Email: \href{mailto:melodyhuang@fas.harvard.edu}{melodyhuang@fas.harvard.edu} URL: \href{https://melodyyhuang.com}{https://melodyyhuang.com}} \and Kosuke Imai\thanks{Corresponding author. Professor, Department of Government and Department of Statistics, Harvard University.  1737 Cambridge Street, Institute for Quantitative Social Science, Cambridge MA 02138.  Email: \href{mailto:imai@harvard.edu}{imai@harvard.edu} URL:
      \href{https://imai.fas.harvard.edu}{https://imai.fas.harvard.edu}} \and Zhichao Jiang\thanks{Professor, School of Mathematics, Sun Yat-sen University, Guangzhou,  Guangdong 510275, China. Email: \href{mailto:jiangzhch7@mail.sysu.edu.cn}{jiangzhch7@mail.sysu.edu.cn}} \and Sooahn Shin\thanks{Ph.D. candidate, Department of Government, Harvard University, Cambridge, MA 02138. Email: \href{mailto:sooahnshin@g.harvard.edu}{sooahnshin@g.harvard.edu}
      URL: \href{https://sooahnshin.com}{https://sooahnshin.com}}}
\date{\today}

\begin{document}
\maketitle

\begin{abstract}
  The use of Artificial Intelligence (AI), or more generally
  data-driven algorithms, has become ubiquitous in today's society.
  Yet, in many cases and especially when stakes are high, humans still
  make final decisions.  The critical question, therefore, is whether
  AI helps humans make better decisions compared to a human-alone or
  AI-alone system.  We introduce a new methodological framework to
  empirically answer this question with a minimal set of assumptions.
  We measure a decision maker's ability to make correct decisions
  using standard classification metrics based on the baseline
  potential outcome.  We consider a single-blinded and unconfounded
  treatment assignment, where the provision of AI-generated
  recommendations is assumed to be randomized across cases with humans
  making final decisions.  Under this study design, we show how to
  compare the performance of three alternative decision-making systems
  --- human-alone, human-with-AI, and AI-alone.  Importantly, the
  AI-alone system includes any individualized treatment assignment,
  including those that are not used in the original study.  We also
  show when AI recommendations should be provided to a human-decision
  maker, and when one should follow such recommendations.  We apply
  the proposed methodology to our own randomized controlled trial
  evaluating a pretrial risk assessment instrument.  We find that the
  risk assessment recommendations do not improve the classification
  accuracy of a judge's decision to impose cash bail.  Furthermore, we
  find that replacing a human judge with algorithms --- the risk
  assessment score and a large language model in particular --- leads
  to a worse classification performance.

\end{abstract}

\clearpage 
\doublespacing
\section{Introduction} 

Artificial Intelligence (AI), or more broadly data-driven algorithms,
have found a wide range of applications in today's society, including
judicial decisions in the criminal justice system, diagnosis and
treatment decisions in medicine, and recommendations in online
advertisement and shopping.  And yet, in many settings and especially
when stakes are high, humans still make final decisions.  The critical
question, therefore, is whether AI recommendations help humans make
better decisions as compared to a human alone or AI alone
\citep{imai2023experimental}.

Recent literature has largely focused on questions of whether AI
recommendations themselves are accurate or biased
\citep[e.g.,][]{barocas2017fairness,corbett2018measure,
  chouldechova2020snapshot, mitchell2021algorithmic,
  imai2023principal}. However, AI recommendations may not improve the
accuracy of human decisions if, for example, the human decision-maker
selectively ignores AI recommendations
\citep[e.g.,][]{hoffman2018discretion, lai2021human,
  cheng2022heterogeneity}. Similarly, the fairness of such an
AI-assisted human decision-making system depends on how the bias of
the AI system interacts with that of the human decisions.

In this paper, we introduce a new methodological framework for
researchers to evaluate experimentally whether the provision of AI
recommendations helps humans make better decisions. We formulate the
notion of a decision-maker's ``ability'' as a classification problem
by combining a standard confusion matrix with the potential outcomes
framework of causal inference \citep{neyman1923application,
  rubin1974estimating}.

For example, when deciding whether to impose cash bail on an arrestee
or to release them on their own recognizance (i.e., an arrestee can be
released without depositing money with the court), a judge must
balance public safety and efficient court administration against
various costs of incarceration.  Thus, the judge's decision-making
ability can be defined as the degree to which they can correctly
classify the as-yet unobserved arrestee's behavior upon release.
Under the proposed framework, we introduce a variety of classification
ability measures.  If AI recommendations are helpful, their provision
should improve a human decision-maker's ability to correctly classify
potential outcomes.

This reasoning demonstrates a primary methodological challenge in
evaluating the impact of various decision-making systems known as the
\textit{selective labels problem}; the decision-makers determine, by
making endogenous decisions, which potential outcomes are observed
\citep[e.g.,][]{lakkaraju2017selective}. In the aforementioned
example, since judges decide which arrestees to release on their own
recognizance or subject to cash bail, we cannot observe whether an
arrestee who received the cash bail decision would have committed a
new crime if a judge were to release them without additional
conditions.

To overcome this selective labels problem, we consider an evaluation
design where the provision of AI recommendations is assumed to be,
possibly conditional on observed covariates, randomly assigned to
human decision-makers across cases.  Such an evaluation design is
feasible even in settings where, for legal or other reasons, a human,
rather than an AI system, must make the final decisions. We consider a
single-blinded treatment assignment, which guarantees that the
provision of AI recommendations affects the outcome only through human
decisions.

Under this evaluation design, we show that, without additional
assumptions, it is possible to point-identify the {\it difference} in
misclassification rates, or more generally the difference in
classification risk, between the human-alone and human-with-AI systems
even though the risk of each decision-making system is not
identifiable.  Moreover, although the proposed evaluation design does
not include the use of an AI-alone decision-making system as one of
the treatment conditions, we derive sharp bounds on the classification
ability differences between an AI-alone system and a human-alone or
human+AI system.  This enables us to evaluate the relative merit of
any AI-alone system \textit{regardless} of whether it was used in the
study.

We demonstrate empirically that these bounds can be informative,
making it possible for researchers to compare the performance of these
decision-making systems without imposing additional assumptions.
Finally, we derive optimal decision rules for when AI recommendations
should be provided to a human decision-maker and when one should
follow such recommendations.

In our empirical illustration, we focus specifically on an
experimental setting, in which all identification assumptions are
guaranteed by the design. However, our methodology is general and also
applicable to observational studies under an additional assumption of
unconfoundedness.

\subsection{Experimental evaluation of a public risk assessment instrument}
\label{subsec:Dane}

We apply the proposed methodology to our own randomized controlled
trial (RCT) to assess how an algorithmically-generated predisposition
risk assessment instrument, called the Public Safety Assessment (PSA),
affects judges' decisions at a criminal first appearance hearing, or
``bail hearing'' \citep{greiner2020dane, imai2023experimental}.  In
Dane County, Wisconsin, where we conducted the RCT, a judge at a first
appearance hearing must decide whether to release an arrestee on their
own recognizance or to impose cash bail as a condition of release. In
this county, own recognizance release is called a ``signature
bond''---a term we will use in the remainder of this paper.

As mentioned earlier, if a judge assigns an arrestee a signature bond,
the arrestee does not need to deposit money with the court to achieve
pretrial release.  For cash bail, the arrested individual must deposit
the specified amount with the court to be released (note that there is
no bail bondsman industry in Wisconsin). The decision between
signature bond and cash bail does not conclusively determine whether
an arrestee will achieve pretrial release. For example, some arrestees
assigned cash bail in fact pay it and thus obtain their pretrial
freedom, while others assigned signature bonds remain incarcerated
because immigration or other criminal justice authorities request a
``hold'' from the relevant jail. Additionally, in Dane County, the judge
at the first appearance hearing can impose other conditions of
release, such as monitoring, drug testing, or stay-away orders. For
the purposes of this paper, however, we do not consider these other decisions.

The PSA provides information to the judge for each decision regarding
the arrested individual's risk of (1) failure to appear (FTA) at
subsequent court dates, (2) new criminal activity (NCA), and (3) new
{\it violent} criminal activity (NVCA). In the RCT, the judge receives
the PSA for a randomly selected subset of all first appearance/bail
hearings \citep[see][for details of the PSA instrument and
experiment]{greiner2020dane}.

The PSA provides three numerical scores that correspond to its
classification of FTA, NCA, and NVCA risks.  The FTA and NCA risk
scores have a total of six levels, while the NVCA risk score is
binary.  Nine factors about prior criminal history as well as age,
which is the only demographic factor, serve as inputs to construct the
PSA's scores; neither race nor gender is an input.  Finally, a
deterministic formula called the Decision-Making Framework (DMF)
combines the PSA's three risk scores with other information, such as a
jurisdiction's resources and risk misbehavior tolerance, to produce an
overall recommendation of either cash bail or signature bond. This
overall PSA-DMF recommendation is the focus of our analysis in this
paper.

To demonstrate our methodology, we analyze the interim data from this
experiment, which is a slightly updated version of the data originally
analyzed by \cite{imai2023experimental} and has been subsequently made
publicly available as \cite{imai2023data}.

\begin{table}[t!]
    \centering
    \small 
    \mbox{
    \begin{tabular}{l|p{1.5cm}|p{1.5cm}|p{1.5cm}|}
        \multicolumn{2}{c}{} & \multicolumn{2}{c}{\textbf{PSA}} \\ \cline{3-4}
        \multicolumn{2}{c|}{} & Signature & Cash \\
        \multicolumn{2}{c|}{} & bond & bail \\ \cline{2-4}
        & Signature & $54.1\%$ & $20.7\%$ \\
        & bond & $(510)$ & $(195)$ \\ \cline{2-4}
        \multirow{-2}{*}{\textbf{Human}} & Cash & $9.4\%$ & $15.8\%$ \\
        & bail & $(89)$ & $(149)$ \\ \cline{2-4}
    \end{tabular}
    \hspace{1em}
    \begin{tabular}{l|p{1.5cm}|p{1.5cm}|p{1.5cm}|}
        \multicolumn{2}{c}{} & \multicolumn{2}{c}{\textbf{PSA}} \\ \cline{3-4}
        \multicolumn{2}{c|}{} & Signature & Cash \\
        \multicolumn{2}{c|}{} & bond & bail \\ \cline{2-4}
        & Signature & $57.3\%$ & $17.1\%$ \\
        & bond & $(543)$ & $(162)$ \\ \cline{2-4}
        \multirow{-2}{*}{\textbf{Human+PSA}} & Cash & $7.4\%$ & $18.2\%$ \\
        & bail & $(70)$ & $(173)$ \\ \cline{2-4}
    \end{tabular}
    }
    \caption{Comparison between human decisions and PSA-generated
      recommendations. The left table compares the PSA recommendations
      (columns) against the judge's decisions without PSA
      recommendations (rows).  Similarly, the right table compares the
      PSA recommendations (columns) with the decisions made by a human
      judge who was provided with PSA recommendations (rows).  Each
      cell presents the proportion of corresponding cases with the
      number of such cases in parentheses.}
    \label{tbl:psai}
\end{table}

Table~\ref{tbl:psai} compares the
PSA-generated recommendations with the judge's decisions  (left table)
and with the human-with-PSA
decisions (right).  The number in each cell represents the proportion
of corresponding cases with the number of such cases in parentheses.
We find that human decisions, with or without PSA recommendations, do
not always agree with the PSA recommendations.  Indeed, the judge goes
against the PSA recommendations in slightly more than 30\% of the
cases.  In the cases of disagreement, the PSA recommendations tend to
be harsher than human decisions.  When provided with the PSA
recommendations, the judge agrees with them slightly more often (by
approximately $5.6$ percentage points with the standard error of
$2.0$; see Figure~\ref{fig:human_ai_disagreement}).  Even in these
cases, however, there exists a substantial amount of disagreement
between the human decisions and PSA recommendations.  Similar to the
human-alone vs. PSA-alone comparison, when they disagree, the PSA
recommendations tend to harsher than the human-with-PSA decisions.

A primary goal of our empirical analysis is to evaluate whether PSA
recommendations improve judges' classification ability in
decision-making. In addition, we are also interested in comparing the
classification ability of the PSA-alone decisions with that of the 
human-alone decisions.  As mentioned above, the key challenge is the presence of
selective labels: for cases where the judge issued a cash bail
decision we do not observe the counterfactual outcome (FTA, NCA, NVCA)
under a signature bond decision.  The evaluation of the PSA-alone
decision-making system in itself is even more difficult because the
experiment does not have a PSA-alone condition.  Our proposed
methodological framework shows how to overcome these challenges with
no additional assumptions beyond those guaranteed by the experimental
design.

Our empirical analysis presented in Section~\ref{sec:empirical} show
that PSA-generated recommendations do not significantly improve the
classification ability of judge's decision-making.  For example, the
misclassification rate of judge's decisions is unchanged when the PSA
recommendations are provided.  We also find that PSA-alone decisions
tend to perform worse than either human-alone or human-with-PSA
decisions.  For both FTA and NVCA, the misclassification rate of
PSA-alone decisions is significantly greater than human decisions with
or without PSA recommendations.  In particular, the PSA system tends
to have a greater proportion of false positives (i.e., imposing cash
bail on an arrestee who would not commit a crime if released on their
own recognizance).

\subsection{Related literature}

Existing literature on algorithmic decision-making has primarily
focused on three areas: (i) the performance evaluation of algorithms
in terms of their underlying classification tasks
\citep[e.g.,][]{berk2016forecasting, goel2016personalized,
  kleinberg2018human, rambachan2022counterfactual}, (ii) issues of
algorithmic fairness and the potential for biased algorithmic or human
recommendations \citep[e.g.,][]{barocas2017fairness,
  corbett2018measure, chouldechova2020snapshot,
  mitchell2021algorithmic, Arnold2021_algo, Arnold2022_bail_decisions,
  imai2023principal}, and (iii) understanding how humans incorporate
algorithmic recommendations into their decision-making
\citep[e.g.,][]{binns2018s, cai2019hello, lai2021human,
  cheng2022heterogeneity,imai2023experimental}.

We focus on a question at the intersection of
these three areas: Do AI recommendations help humans make better
decisions than those made by a human alone or an AI alone?  While other
scholars have proposed using a classification framework with selective
labels to consider the performance of algorithmic decision-making
\citep[e.g.,][]{goel2016personalized, kleinberg2018human,
  rambachan2022counterfactual}, we focus on the relative
gains of AI recommendations over human decisions and use an RCT to
identify more credibly our key quantities of interest without
additional assumptions.

Our work contributes to a growing literature that addresses the
selective labels problem when evaluating human decisions and AI
recommendations. In particular, we consider an evaluation design in
which the provision of AI recommendations is either randomized or
assumed to be unconfounded while single-blinding the treatment
assignment so that AI recommendations affect the outcome only through
human decisions.  We show that under this design, it is possible to
evaluate the classification performance of human-alone, AI-alone, and
human-with-AI decision-making systems.  In contrast, related studies
have been restricted to observational settings. For example, previous
works exploit discontinuities at algorithmic thresholds and staggered
roll-outs of algorithms \citep[e.g.][]{berk2017impact, albright2019if,
  stevenson2022algorithmic, coston2021characterizing,
  guerdan2023ground} or use survey evaluations
\citep[e.g.,][]{miller2013practitioner, skeem2020impact}.

Several studies have advocated designs that use quasi-random
assignment to decision-makers with different decision rates.  They use
the differing decision rates as an instrumental variable to estimate
various performance measures of AI recommendations relative to human
decision-makers \citep[e.g.][]{kleinberg2018human, dobbie2018effects,
  Arnold2022_bail_decisions, angelova2023algorithmic}.  Unlike these
studies, we consider an experimental setting where we are able to
guarantee the required identification assumptions \textit{by design}.
Furthermore, while related approaches have been used to evaluate
algorithmic decisions, we also evaluate the relative performance of
human decision-makers with and without AI recommendations, as well as
the AI-alone decision-making system.

Closest to our approach is \citet{angelova2023algorithmic}, who
compare AI-assisted human decisions to those of the algorithm by
studying cases where humans override the algorithmic
recommendation \citep[see also][]{hoffman2018discretion}. Our
framework is also similar to the one proposed by
\cite{imai2023experimental}, but we focus on a single potential
outcome rather than joint potential outcomes, allowing us to avoid
making additional assumptions.

When point identification is not possible, we use partial identification to bound
the quantities of interest 
\citep[e.g.,][]{manski2007}.  This methodological development is
related to partial identification approaches proposed by
\cite{Rambachan2021} and \cite{rambachan2022counterfactual}. In
particular, \cite{rambachan2022counterfactual} consider general
approaches to partial identification of the predictive performance of
classification algorithms. In contrast, we focus on comparing the
predictive performance of the aforementioned three different
decision-making systems that involve humans and/or AI, leading to
different identification results.

\section{The Evaluation Framework}
\label{sec:estimands}

In this section, we introduce a methodological framework for
evaluating statistically the relative performance of human-alone,
human-with-AI, and AI-alone decision-making systems.  We describe our
evaluation design, and then formalize a decision-maker's ability as a
classification problem.

\subsection{Design and assumptions}
\label{subsec:experiment}

Let $A_i \in \{0,1\}$ represent the binary AI-generated recommendation
for case $i$.  We assume that AI recommendations can be computed for
all cases.  In our application, $A_i = 1$ means that AI recommends
cash bail while $A_i = 0$ indicates that AI recommends a signature
bond.  We use $Z_i \in \{0,1\}$ to denote the binary treatment
variable representing the provision of such an AI recommendation.  In
our experiment, $Z_i = 1$ indicates that case $i$ is assigned to a
human judge with a PSA recommendation, whereas $Z_i = 0$ means that no
PSA recommendation is given to the judge.  We will also assume that we
observe case-level covariate information, denoted as
$X_i \in \mathcal{X}$ where $\mathcal{X}$ is the support.

The proposed methodology can be generalized to settings with more than
two treatment conditions.  For example, researchers may use different
AI systems or include an AI-alone decision as a separate treatment
arm. For the sake of simplicity, we focus on the binary case of two
treatment arms in this paper.

We use $D_i \in \{0, 1\}$ to denote the observed binary decision made
by a human.  In our application, $D_i = 1$ represents a judge's
decision to require cash bail as opposed to a signature bond.  In a
medical setting, $D_i = 1$ may signify a doctor's decision to
prescribe medication rather than not to treat.  Because the AI
recommendation can affect the human decision, we use potential
outcomes notation and denote the decision under the treatment
condition $Z_i =z$ as $D_i(z)$.  That is, $D_i(1)$ and $D_i(0)$
represent the decisions made with and without an AI recommendation,
respectively.  The observed decision, therefore, is given by
$D_i = D_i(Z_i)$.

In addition, let $Y_i \in \{0,1\}$ denote the binary outcome of
interest.  Without loss of generality, we assume that $Y_i = 1$
represents an undesired outcome relative to $Y_i = 0$.  In our
empirical application, this variable represents whether or not an
arrestee fails to appear in court (FTA), or engages in new criminal
activity (NCA) or new violent criminal activity (NVCA).

We consider the use of single-blinded treatment assignment.  In our
application, this means that an arrestee does not know whether a judge
receives an AI recommendation.  In other words, we assume that the
provision of an AI recommendation, or lack thereof, can affect the
outcome only through the human decision.  The assumption is violated
if a judge informs an arrestee about the AI recommendation, which in
turn affects the arrestee's behavior directly other than through the
judge's decision.  Formally, let $Y_i(z, d)$ denote the potential
outcome under the treatment condition $Z_i = z$ and the decision
$D_i = d$.  The single-blinded experiment assumption implies
$Y_i(0,d)=Y_i(1,d) = Y_i(d)$ for any $d$ where the observed outcome is
given by $Y_i=Y_i(D_i(Z_i))$.

In sum, our evaluation design requires three assumptions:
unconfoundedness, overlap, and single-blindedness of treatment
assignment. We formally present these identification assumptions.
\begin{assumption}[Single-blinded and unconfounded treatment
  assignment] \label{assum:single_blinded} \spacingset{1} The
  treatment assignment $Z_i$, potential decisions $D_i(z)$,
  pre-treatment covariates $X_i$, and potential outcomes
  $Y_i(z, D_i(z))$ satisfy:
  \begin{enumerate}[label=(\alph*)]
  \item Single-blinded treatment assignment: $Y_i(z,D_i(z)) = Y_i(z',D_i(z'))$ for all $z,z'$ such that $D_i(z) = D_i(z')$
  \item Unconfounded treatment assignment:
    $Z_i \ \indep \ \{A_i, \{D_i(z), Y_i(d)\}_{z,d \in \{0,1\}}\} \mid
    X_i$
  \item Overlap: $e(x) := P(Z_i = 1 \mid X_i = x) \in [\eta, 1-\eta]$
    for $\eta > 0$
  \end{enumerate}
\end{assumption}

In RCTs, this unconfoundedness assumption allows for
covariate-dependent random assignment designs based on pre-treatment
covariates $X_i$ such as blocking and matched-pair designs. We also
assume the overlap condition that the treatment probabilities are
bounded away from zero and one, i.e., $e(x) \in [\eta, 1 - \eta]$.  In
observational studies, the treatment probabilities are unknown and
hence must be estimated.  Below, we will state all results in
generality for the observational study case and note where they
simplify for the experimental case.

The notation above implicitly assumes no spillover effects across
cases.  In our application, this means that a judge's decision should
not be influenced by the treatment assignments of prior cases. To
increase the credibility of this assumption, we focus on first arrest
cases (see Section~S3 of Online Supplementary Information of
\cite{imai2023experimental} for empirical evidence consistent with
this assumption).  Finally, we assume that we have an independently
identically distributed sample of cases with size $n$ from a target
distribution $\mathcal{P}$, i.e.,
$(A_i, X_i, D_i(0), D_i(1), Y_i(0), Y_i(1)) \overset{i.i.d.}{\sim}
\mathcal{P}$. In subsequent sections, we will omit the $i$ subscript
from expressions whenever convenient.

\subsection{Measures of classification ability}
\label{subsec:ability}

We use a classification framework to formalize the `classification
ability' of a decision-maker.  We focus on the baseline potential
outcome $Y(0)$.  In our application, this corresponds to the outcome
we would observe (e.g., an NCA) if an arrestee is released on their
own recognizance ($D = 0$).  The fact that the PSA is designed to
predict the behavior of an arrestee if released on their own
recognizance justifies the focus on $Y(0)$ in our application.

\begin{table}[t]
\centering 
    \begin{tabular}{l|c|c|c|} 
  \multicolumn{2}{c}{}  & \multicolumn{2}{c}{\textbf{Decision}} \\\cline{3-4}
      \multicolumn{2}{c|}{}  & \multirow{2}{*}{Negative $(D^\ast = 0$)}  & \multirow{2}{*}{Positive $(D^\ast = 1$)}\\
      \multicolumn{2}{c|}{} & \multicolumn{1}{c|}{} & \multicolumn{1}{c|}{} \\ \cline{2-4}
 & \multirow{2}{*}{Negative ($Y(0) = 0$)} & \cellcolor[HTML]{B9EBB7}True Negative (TN) & \cellcolor[HTML]{FFCCC9}{\color[HTML]{333333} False Positive (FP)} \\
 &   & $\ell_{00}$ &  $\ell_{01}$ \\ \cline{2-4}
 \multirow{-2}{*}{\textbf{Outcome}} &  \multirow{2}{*}{Positive ($Y(0) = 1$)} & \cellcolor[HTML]{FFCCC9}{\color[HTML]{333333} False Negative (FN)} & \cellcolor[HTML]{B9EBB7}True Positive (TP)\\
 &   & $\ell_{10} = 1$ &  $\ell_{11}$ \\ \cline{2-4}
 \cline{2-4} 
\end{tabular}
\caption{Confusion matrix for each combination of baseline potential
  outcome $Y(0)$ and decision $D^*$.  Each cell is assigned a loss
  $\ell_{yd}$ for $y,d \in \{0,1\}$. The loss is standardized by
  setting $\ell_{10}=1$.} 
\label{tbl:confusion}
\end{table}

Table~\ref{tbl:confusion} shows the confusion matrix for all four
possible pairs of baseline potential outcome $Y(0)$ and a generic
decision $D^\ast$, which is different from the observed human decision
$D$ in the study.  If the baseline potential outcome is negative in
the classification sense, i.e., $Y(0)=0$ (e.g., an arrestee would not
commit a new crime) and the decision is also negative $D^\ast=0$
(e.g., a judge decides to assign a signature bond), then we call this
instance a `true negative' or TN.  In contrast, if the baseline
outcome is positive (i.e., an undesired outcome in our application)
and yet the decision is negative (e.g., a judge decides to release the
arrestee on their own recognizance), then this instance is called
`false negative' or FN. False positives (FP) and true positives (TP)
are similarly defined.

Using this confusion matrix, we can derive a range of classification
ability measures.  To do so, we first assign a loss (or negative
utility) to each cell of the confusion matrix and then aggregate
across cases.  As shown in Table~\ref{tbl:confusion}, let $\ell_{yd}$
denote a loss that is incurred when the baseline potential outcome is
$Y(0)=y$ and the decision is $D^\ast=d$ for $y,d\in\{0,1\}$.  Without
loss of generality (no pun intended), we set the loss of a false
negative to one, i.e., $\ell_{10}=1$.

This setup allows for both symmetric and asymmetric loss functions
\citep{benm:imai:jian:23}. If a false positive (e.g., unnecessary cash
bail) and a false negative (e.g., signature bond, resulting in
FTA/NCA/NVCA) incur the same loss, i.e., $\ell_{01}=1$, then the loss
function is said to be symmetric.  An asymmetric loss function arises,
for example, if avoiding false negatives is deemed more valuable than
preventing false positives, i.e., $\ell_{01} < 1$. To simplify the
exposition, we will consider loss functions where true negatives and
true positives incur zero loss (i.e., $\ell_{11} = \ell_{00} = 0$; see
Appendix~\ref{app:general_case} for a discussion of generic loss
functions).

Once the loss function is defined, we compute the {\it
  classification risk} (or expected classification loss) as the
average of the the false negative proportion (FNP) and false positive
proportion (FPP), weighted by their respective losses,
\begin{equation}
  R(\ell_{01}; D^\ast) \ := \ p_{10}(D^\ast)+ \ell_{01}
  p_{01}(D^\ast), 
  \label{eqn:expected_loss}
\end{equation}
where $p_{10}(D^\ast)$ and $p_{01}(D^\ast)$ represent the overall FNP
and FPP, respectively, under a decision making system $D^\ast$.  More
generally, we can define the following confusion matrix:
\begin{equation}
  C(D^\ast) \ := \ \left[
    \begin{array}{c c}
      p_{00}(D^\ast) & p_{01}(D^\ast) \\
      p_{10}(D^\ast) & p_{11}(D^\ast) 
    \end{array}
  \right], \label{eq:cm}
\end{equation}
where $p_{yd}(D^\ast) :=\Pr(Y(0) = y, D^\ast = d)$ for
$y, d \in \{0,1\}$. 

We use this measure of classification ability to evaluate three
decision-making systems: human-alone ($D^\ast = D(0)$), human-with-AI
($D^\ast=D(1)$), and AI-alone ($D^\ast = A$).  We are particularly
interested in contrasting the classification abilities of these three
systems.  For example, the comparison of human-alone and human-with-AI
tells us whether AI recommendations are able to improve human
decision-making.

We can also define other measures of classification ability.  Examples
include (suppressing $D^\ast$ for notational simplicity): (1) the
misclassification rate, i.e., $p_{10} + p_{01} =R(1)$, which
represents the overall proportion of incorrect decisions, (2) the
false discovery rate, i.e., $p_{01} /(p_{01}+p_{11})$, which is equal
to the proportion of incorrect decisions among positive decisions, (3)
the false negative rate, i.e., $p_{10} / (p_{10}+p_{11})$, which is
the proportion of incorrect decisions among positive baseline
outcomes, and (4) the false positive rate, i.e.,
$p_{01}/(p_{00} + p_{01})$, which equals the proportion of incorrect
decisions among negative baseline outcomes. See Appendix
\ref{app:bounds_alt_measures} for more details.

\subsection{Discussion}

Prior work has considered these classification ability measures.  For
example, in the pre-trial risk assessment setting,
\citet{angelova2023algorithmic} consider the misconduct rate among
released defendants; this corresponds to the \emph{false negative
  rate}.  \citet{dobbie2018effects} consider the proportion of
individuals who are detained erroneously, i.e., the \emph{false
  discovery rate}.  Other work develops a more general framework.  For
instance, \citet{rambachan2022counterfactual} consider a generalized
notion of performance that includes functions of the confusion matrix
as well as other measures like calibration and mean square error,
which we do not consider here.

One important limitation of our framework and related approaches,
however, is that we only consider the baseline potential outcome
rather than the joint potential outcomes.  The ``correct'' or
``wrong'' decision might be depend on both potential outcomes instead
of the baseline potential outcome alone.  Unfortunately, the
consideration of joint potential outcomes requires stronger
assumptions than those considered under our approach.

Nevertheless, \citet{imai2023experimental} introduce a principal
stratification framework that considers the joint set of potential
outcomes and three principal strata of individuals: (1)
\textit{preventable cases} $(Y(1), Y(0)) = (0,1)$) --- individuals who
would engage in misconduct only if released, (2) risky cases
$(Y(1), Y(0)) = (1,1)$) --- individuals who would engage in misconduct
regardless of the judge’s decision, and (3) safe cases
$(Y(1), Y(0) = (0,0)$) --- individuals who would not engage in
misconduct regardless of the detention decision. The authors focus on
the effect of provision of the AI recommendation on the human
decision, conditioned on these principal strata.
\cite{imai2023principal} also introduce a related fairness notion,
called principal fairness.

\section{Identification, Estimation, and Hypothesis Testing}

In this section, we present the proposed methodology based on the
evaluation framework introduced above.  We first show how to compare
the human-alone and human-with-AI decision-making systems.  Next, we
develop a partial identification approach to the evaluation of the
AI-alone system, which does not correspond to an arm in the experiment,
in comparison with the human and human+AI decision-making systems.
Finally, we investigate the question of when a human decision-maker
should follow or ignore an AI recommendation.  

\subsection{Comparing human decisions with and without AI recommendations}\label{sec:human}

To compare the performance of human decisions with and without AI
recommendations, we first derive the key identification result before
presenting our estimation strategy.  We also propose a
statistical hypothesis testing framework to compare different loss
functions.

\subsubsection{Identification}

As explained in Section~\ref{subsec:ability}, our primary
methodological challenge is the selective labels problem.
Specifically, we observe the baseline potential outcome under the
negative decision $Y(0)$ only for cases where the decision is
actually negative, i.e., $D = 0$.  Despite this problem, we show that
under the experimental design introduced in
Section~\ref{subsec:experiment}, it is possible to point-identify the
difference in the misclassification rate --- or more generally the
difference in classification risk --- between human decisions with and
without AI recommendations.

To begin, the difference in classification risk between these two
decision-making systems is given by,
\begin{equation*}
R_{\textsc{human+AI}}(\ell_{01})-R_{\textsc{human}}(\ell_{01}) \ = \
\{p_{10}(D(1)) - p_{10}(D(0))\} + \ell_{01} \{p_{01}(D(1)) - p_{01}(D(0))\}.
\end{equation*}
Under Assumption~\ref{assum:single_blinded}, we can immediately
identify the effect of providing an AI recommendation on the FNP,
i.e., $p_{10}(D(1)) - p_{10}(D(0))$.  Indeed, we can identify both the
FNP and true negative proportion (TNP) separately for the human-alone
and human-with-AI decision-making systems.

Unfortunately, the FPP, $p_{01}(D(z))$, is not identifiable for
$z=0,1$. Despite this fact, we can identify the average {\it effect}
of access to AI recommendations on the FPP. Specifically, under
Assumption~\ref{assum:single_blinded}, the distribution of the
baseline potential outcome under the negative decision is the same
across the treatment and control groups within strata, i.e.,
$\Pr(Y(0) = 0 \mid Z = 1, X = x) = \Pr(Y(0) = 0 \mid Z = 0, X =x)$.
By the law of total probability, this equality implies
${\color{blue}{p_{01}(D(1)\mid X = x)}}+ p_{00}(D(1) \mid X = x) =
{\color{blue}{p_{01}(D(0) \mid X = x)}} + p_{00}(D(0) \mid X = x)$
where
$p_{yd}(D^\ast\mid X= x) := \Pr(Y(0) = y, D^\ast = d \mid X = x)$ and
the terms in blue are not identifiable. Then, we can rearrange the
terms and obtain: \begin{equation} {\color{blue}p_{01}(D(1) \mid X =
    x) - p_{01}(D(0) \mid X = x)} \ = \ p_{00}(D(0) \mid X = x) -
  p_{00}(D(1) \mid X = x). \label{eqn:tp_id} \end{equation}
Equation~\eqref{eqn:tp_id} allows us to point-identify the difference
in classification risk between human decisions with and without an AI
recommendation.  The following theorem formally states the result.
\begin{theorem} {\sc (Identification of the difference in
    classification risk between human-alone and human-with-AI
    systems)}
  \label{thm:z_vs_z0} \spacingset{1} Under
  Assumption~\ref{assum:single_blinded}, we can identify the
  difference in risk between human decisions with
  ($Z = 1$) and without ($Z = 0$) an AI recommendation as:
  $$\begin{aligned}
    &  R_{\textsc{human+AI}}(\ell_{01})-R_{\textsc{human}}(\ell_{01}) \\
\ = \ & \E\left[\Pr(Y=1, D =0 \mid Z=1, X) - \Pr(Y = 1, D = 0 \mid Z =
  0, X) \right. \\
  & \left. \quad - \ell_{01}
  \times \left\{\Pr(Y = 0, D = 0 \mid Z = 1, X) - \Pr(Y = 0, D =0 \mid Z = 0, X)
  \right\}\right],
\end{aligned}
$$
where $R_{\textsc{human}}(\ell_{01}):=R(\ell_{01}; D(0))$ and
$R_{\textsc{human+AI}}(\ell_{01}):=R(\ell_{01}; D(1))$ as defined in
  Equation~\eqref{eqn:expected_loss}.
\end{theorem}
A special case of this result when $\ell_{01}=1$ corresponds to the
identification of the difference in the misclassification rates
between the two decision-making systems, which we present in our
empirical analysis in Section~\ref{sec:empirical}.

\subsubsection{Estimation}

To estimate the difference in classification risk from the
identification result in Theorem~\ref{thm:z_vs_z0}, we first notice
that we can write the identified form as the difference in means of a
compound outcome:
$$W_i := Y_i(1-D_i)- \ell_{01} (1-Y_i)(1-D_i).$$
We can estimate this difference in classification risk via a variety
of approaches, including the simple difference-in-means estimator.

Here, we consider a more general estimation approach based on the
Augmented Inverse Probability Weighting (AIPW) estimator that can also
be applied to observational studies \citep{Robins1994}. We begin by
defining two nuisance components: (i) the decision model
$m^D(z, x) := \Pr(D = 1 \mid Z = z, X = x)$ and
(ii) the outcome model
$m^Y(z, x) := \Pr(Y = 1 \mid D = 0, Z = z, X = x)$. For notational
simplicity, we also define the propensity score under the treatment
assignment $z$ as $e(z, x) := z e(x) + (1 - z)(1 - e(x))$ for a given
pre-treatment covariate value $x$.  We assume that we estimate these
nuisance components (and the propensity score $e(x)$) on a separate
sample and that they satisfy the following rate conditions:
\begin{assumption}{\sc Nuisance component rate conditions}
  \label{assum:rates}
  For each $z =0,1$, we have:
  $$\begin{aligned}\left(\|m^D(z,\cdot) - \hat{m}^D(z, \cdot)\|_2  +
      \|m^Y(z,\cdot) - \hat{m}^Y(z, \cdot)\|_2\right) \times \|e -
    \hat{e}\|_2 & = o_p(n^{-\frac{1}{2}}), \\
    \|m^Y(z,\cdot) - \hat{m}^Y(z,\cdot)\|_\infty \ = o_p(1), \quad
    \|m^D(z,\cdot) - \hat{m}^D(z,\cdot)\|_\infty = o_p(1), & \quad
    \|e- \hat{e}\|_\infty = o_p(1),\end{aligned}$$ where for a given function
  $f$, $\|f\|_2^2 = \E[f(X)^2]$ and
  $\|f\|_\infty = \sup_{x \in \mathcal{X}}|f(x)|$.
\end{assumption}

Assumption~\ref{assum:rates} relates to the standard product-rate
assumption for doubly-robust AIPW estimators, but it involves both the
decision and outcome models because the compound outcome $W_i$ involves
the product of the decision and the outcome.\footnote{To establish
  Theorem~\ref{thm:z_vs_z0_est}, it is sufficient to have only a rate
  requirement for a combination of the two models. For clarity,
  however, we give a somewhat stronger sufficient condition in
  Assumption~\ref{assum:rates}.}  In our main case of a randomized
experiment, the propensity scores are known, and so we only require
that we consistently estimate the outcome models, with no particular
rate requirement.

Once these nuisance components are estimated, we estimate the
difference in classification risk as
$$\hat{\beta} = \frac{1}{n}\sum_{i=1}^n \left\{ \widehat{\varphi}_1(Z_i, X_i,
D_i, Y_i; \ell_{01}) - \widehat{\varphi}_0(Z_i, X_i, D_i, Y_i;
\ell_{01})\right\},$$ where the estimates of the (uncentered) influence
function given by,
\begin{align*}
  & \widehat{\varphi}_z(Z, X, D, Y; \ell_{01}) \\
  := &  \left(1 - \hat{m}^D(z, X)\right)\left\{(1 + \ell_{01}) \hat m^Y(z, X) - \ell_{01}\right\}+ (1 + \ell_{01})\frac{\bbone\{Z = z\}(1 - D)}{\hat e(z, X)}\left(Y - \hat m^Y(z, X)\right)\\
  & \qquad  - \left\{(1 + \ell_{01})\hat m^Y(z, X) - \ell_{01}\right\}\frac{\bbone\{Z  = z\}}{\hat e(z, X)}\left(D - \hat m^D(z, X)\right),
\end{align*}
for $z=0,1$.
We similarly define the true (uncentered) influence function as $\varphi_z(Z, X, D, Y; \ell_{01})$.

When the rate and consistency conditions are satisfied, this estimator
is asymptotically normally distributed around the true classification
risk difference.  Although for simplicity we assume that the nuisance
components are fit on a separate sample, this is not necessary.  All
results readily extend to cross-fit estimators such as those we use in
our application.
\begin{theorem}[Asymptotic normality] \spacingset{1}
  \label{thm:z_vs_z0_est}
  Under Assumptions~\ref{assum:single_blinded}~and~\ref{assum:rates},
  \[\sqrt{n} \left(\hat{\beta} - (R_{\textsc{human+AI}}(\ell_{01})-R_{\textsc{human}}(\ell_{01}))\right) \stackrel{d}{\longrightarrow} N(0, V),\] where $V = \E[\{\varphi_1(Z, X, D, Y;\ell_{01}) - \varphi_0(Z, X, D, Y;\ell_{01}) - (R_{\textsc{human+AI}}(\ell_{01})-R_{\textsc{human}}(\ell_{01}))\}^2]$. 
\end{theorem}

Using this result, we can construct asymptotic confidence intervals
for the classification risk difference by first estimating the
asymptotic variance as
$$\widehat{V} = \frac{1}{n}\sum_{i=1}^n\left(\widehat{\varphi}_1(Z_i, X_i, D_i,
  Y_i;\ell_{01}) - \widehat{\varphi}_0(Z_i, X_i, D_i, Y_i;\ell_{01}) -
  \hat{\beta}\right)^2.$$ Then, we obtain Wald-type $1-\alpha$
confidence intervals as $\hat{\beta} \pm z_{1-\alpha/2} \sqrt{\widehat{V}/n}$,
where $z_{1 - \alpha/2}$ is the $1-\alpha/2$ quantile of a standard
normal distribution.

\subsubsection{Comparing different loss functions}

Whether one prefers the human decision-making system with or without
AI recommendations depends on the chosen loss function (i.e., the
value of $\ell_{01}$ in Equation~\eqref{eqn:expected_loss}).  Using
Theorem~\ref{thm:z_vs_z0}, we can ask under what loss functions we
might prefer the human-with-AI decision-making system over the
human-alone system.

Specifically, we first consider the following hypothesis test that for
a given ratio of the loss between false positives and false negatives
$\ell_{01}$, the risk is lower for the human-with-AI system:
$$
  H_0: R_{\textsc{human}}(\ell_{01}) \leq R_{\textsc{human+AI}}(\ell_{01}), \quad
  H_1:  R_{\textsc{human}}(\ell_{01}) > R_{\textsc{human+AI}}(\ell_{01}). 
$$
Inverting this hypothesis test for the parameter $\ell_{01}$ gives the
values of the false positive loss for which we cannot rule out that
the human-alone system is
better than the human-with-AI system.  Conversely, the
region of $\ell_{01}$ where we reject $H_0$ gives the loss functions
for which we can rule out the possibility that the human-alone system is
better than the human-with-AI system.

Similarly, if we flip the null and alternative hypotheses so that
$H_1$ becomes the null hypothesis, the region where we can reject it
gives the relative values of the false positive loss for which we can
rule out the scenario that the human-with-AI system is better. The
remaining cases are ambiguous.

\subsection{Comparing AI decisions with human-alone and human-with-AI
  systems}
\label{sec:partial}

We next compare the classification ability of AI-alone decisions with
human-alone and human-with-AI systems.  The proposed approach is extremely general. In particular, 
we are able to consider a
generic AI-alone decision system that need not be the same AI
recommendation system used in the experiment.  In other words, we can
analyze how a hypothetical AI-alone decision system would perform in
comparison with a human-alone or human-with-AI system.  This allows researchers to
use the data from a single experiment to evaluate different
individualized decision rules, as long as the AI decision can
be computed for any unit.  In Section~\ref{subsec:LLM}, we compare the
classification ability of a large language model with that of a human
judge.

Unlike the comparison between human-alone and
human-with-AI systems analyzed above, we cannot point-identify the
risk difference without imposing additional assumptions. This is
because the proposed evaluation design does not have a treatment arm where an AI system yields decisions without human input.  We do,
however, observe AI recommendations for all cases since they can be
readily computed.  Below, we leverage this fact and derive informative
bounds on the difference in classification risk between AI and human
(with or without AI recommendations) decisions.  Together with the
results presented above, these bounds enable analysts to compare the
ability of all three decision-making systems using our evaluation
design.

\subsubsection{Partial identification}

The fundamental problem is that we do not observe the potential
outcome under AI decisions $Y_i(A_i)$ whenever human decisions in the
experiment (with or without AI recommendations) disagree with AI
decisions, i.e., $A_i \ne D_i$.  For example, Table~\ref{tbl:psai}
shows that in our application the judge disagrees with AI
recommendations in more than 25\% of the cases. To evaluate the
AI-alone system, therefore, we must deal with this selective labels
problem.

We begin by considering the confusion matrix for the AI-alone system
that marginalizes over the different human decisions $C(A)$ (see
Equation~\eqref{eq:cm}). The classification risk of the AI-alone
system is defined as:
$$R_{\textsc{AI}}(\ell_{01}) \ := \ R(\ell_{01}; A) \ = \ \emp{p_{10}(A)} +
\ell_{01} \emp{p_{01}(A)}.$$ where the terms in blue are
unidentifiable. Because the experiment does not contain an AI-alone
treatment arm, each element of the confusion matrix $C(A)$ is a
mixture of identifiable and non-identifiable parts:
$$p_{ya}(A) = \Pr(Y(0) = y, A = a) = {\color{blue}{\Pr(Y(0) = y, A = a, D = 1)}} +  \Pr(Y(0) = y, A = a, D = 0).$$
As a result, without further assumptions, the classification risk of
an AI-alone system cannot be identified.

However, we can partially identify the differences in classification
risk between the AI-alone and human-alone/human-with-AI
decision-making systems, focusing on the cases where AI
recommendations differ from human
decisions. Theorem~\ref{thm:partial_id_risk} provides sharp bounds on
the range of possible values that the risk differences can take on
(see Theorem~\ref{thm:partial_Dstar_deterministic} in
Appendix~\ref{app:Dstar} for the sharp bounds on the classification
risk of a generic AI decision system).

\begin{theorem} {\sc (Sharp Bounds on the differences in classification risk
    between AI-alone and human systems)} \label{thm:partial_id_risk}
  \spacingset{1} Under Assumption~\ref{assum:single_blinded}, the risk differences are sharply bound by the following:
  $$\begin{aligned}
    \E[L_0(X)] & \ \le \ R_{\textsc{AI}}(\ell_{01})-R_{\textsc{Human}}(\ell_{01}) & \ \le  \ \E[U_0(X)], \\
    \E[L_1(X)] & \ \le \ R_{\textsc{AI}}(\ell_{01})-R_{\textsc{Human+AI}}(\ell_{01}) & \ \le \ \E[U_1(X)], 
  \end{aligned}$$
  where $L_z(x)$ and $U_z(x)$ are defined as
  \begin{eqnarray*}
    L_z(x)&:=& (1+\ell_{01}) \left\{  \max_{z'} \Pr(Y=1,D=0,A=0\mid Z=z',X = x) - \Pr(Y=1,D=0\mid Z=z, X = x) \right\} \\
    \nonumber&&+\ell_{01}  \left\{ \Pr(D = 0, A = 1 \mid Z= z, X = x)- \Pr(D=1,A=0\mid Z=z, X = x) \right\},\\
    U_z(x) &:=&(1+\ell_{01}) \bigg \{ \Pr(A=0 \mid X = x) -  \Pr(Y=1,D=0\mid Z=z, X = x)\\
    \nonumber&& \left. \qquad \qquad - \max_{z'} \Pr(Y=0,D=0,A=0\mid Z=z', X = x)\right\} \\
    \nonumber&&+\ell_{01}  \left\{ \Pr(D = 0, A = 1 \mid Z= z, X = x)- \Pr(D=1,A=0\mid Z=z, X = x) \right\}.
  \end{eqnarray*}
\end{theorem}
The expressions of the lower and upper bounds involve the maximum
value taken over the treatment assignment $Z$, showing that the
randomization of two treatment arms helps narrow the bounds.  The
width of the bounds is equal to,
  \begin{eqnarray*}
\E\{U_z(X)-L_z(X)\} &=&(1+\ell_{01}) \E \left\{ \Pr(A=0 \mid X )- \max_{z'} \Pr(Y=1,D=0,A=0\mid Z=z',X ) \right.\\
&&\hspace{2cm} \left. - \max_{z'} \Pr(Y=0,D=0,A=0\mid Z=z', X ) \right\}.
\end{eqnarray*}
Thus, the bounds tend to be narrow when the judge's decisions (with or
without AI recommendations) align with the AI recommendations.  Note
that the only cases with $D=0$ matter because we focus on the
potential outcome $Y(0)$.

\subsubsection{Estimation}

We now turn to estimation. Again, we consider a general approach that
is applicable to observational studies.  To do so, we define
additional nuisance components corresponding to the decision and outcome
models additionally conditioned on the AI recommendation $A$:
$m^D(z, x, a) := \Pr(D = 1 \mid Z = z, X = x, A = a)$ and
$m^Y(z, x, a) := \Pr(Y = 1 \mid D = 0, Z = z, X = x, A = a)$.
Estimating the sharp bounds derived in
Theorem~\ref{thm:partial_id_risk} is complex, requiring (i) the
determination of which choice of $z'$ achieves the tighter bound and
(ii) the estimation of the bound given the optimal value of $z'$.

Tackling the first component, for each value of the covariates
$x \in \mathcal{X}$, we can characterize whether using $z' = z$ or
$z' = 1-z$ results in a greater lower bound with a \emph{nuisance
  classifier}
\[
g_{L_z}(x) = \bbone\{(1 - m^D(1-z, x, 0))m^Y(1-z, x, 0) \geq (1 - m^D(z, x, 0))m^Y(z, x, 0)\},
\]
where $g_{L_z}(x) = 0$ denotes that the optimal choice is $z'=z$ and
$g_{L_z}(x) = 1$ denotes that it is $z' = 1-z$.  Similarly, we can
characterize the choice of $z'$ for the least upper bound with another
nuisance classifier,
\[
  g_{U_z}(x) = \bbone\{(1 - m^D(1-z, x, 0))(1 - m^Y(1-z, x, 0)) \geq (1 - m^D(z, x, 0))(1 - m^Y(z, x, 0))\}.
\]
We estimate these nuisance classifiers using a simple plugin approach:
first we estimate $\hat{m}^D(z, x, a)$ and $\hat{m}^Y(z, x, a)$, then plug in
these estimates into the formulas for the nuisance classifiers.

For the second step, we estimate the bound corresponding to the choice
of $z'$ using an efficient AIPW estimator.  We do this by noting that
we can write the conditional probabilities in
Theorem~\ref{thm:partial_id_risk} as conditional expectations of
compound outcomes: $Y(1-D)(1-A)$, $(1-Y)(1-D)(1-A)$, $(1-A)D$, and $A
(1-D)$.  The final estimator uses the following two sets of influence
function estimates.  The first is for the models of $Y(1-D)(1-A)$ and
$(1-Y)(1-D)(1-A)$,
\begin{align*}
  \widehat{\varphi}_{z1}(Z, X, D, A, Y) & = (1 - A) (1 - \hat{m}^D(z, X, 0))\hat{m}^Y(z, X, 0) + \frac{\bbone\{Z = z\}(1 - A)(1-D)}{\hat{e}(z,X)}(Y - \hat{m}^Y(z, X, 0))\\
  & \qquad - \hat{m}^Y(z, X, 0)\frac{\bbone\{Z = z\}(1 - A)}{\hat{e}(z,X)}(D - \hat{m}^D(z, X, 0)),\\
  \widehat{\varphi}_{z0}(Z, X, D, A, Y) & = (1 - A) (1 - \hat{m}^D(z, X, 0))(1 - \hat{m}^Y(z, X, 0)) - \frac{\bbone\{Z = z\}(1 - A)(1-D)}{\hat{e}(z,X)}(Y - \hat{m}^Y(z, X, 0))\\
  & \qquad - (1-\hat{m}^Y(z, X, 0)) \frac{\bbone\{Z = z\}(1 - A)}{\hat{e}(z,X)}(D -
    \hat{m}^D(z, X, 0)).
\end{align*}
The second set is for the models of $A (1-D)$, and $(1-A)D$,
\begin{align*}
  \widehat{\varphi}^D_{z1}(Z, X, D, A) & = A(1 - m^D(z, X, 1)) - \frac{\bbone\{Z = z\} }{\hat{e}(Z,X)} A (D - m^D(z, X, 1))),\\
  \widehat{\varphi}^D_{z0}(Z, X, D, A) & = (1 - A)m^D(z, X, 0) + \frac{\bbone\{Z = z\} }{\hat{e}(Z,X)} (1 -A) (D - m^D(z, X, 0)).
\end{align*}
As before, we remove the circumflexes to refer to the true uncentered influence functions. Finally, we estimate the upper and lower bound as
\begin{align*}
  \widehat{L}_z & = \frac{1}{n}\sum_{i=1}^n (1 + \ell_{01})\left(\widehat{\varphi}_{z1}(Z_i, X_i, D_i, A_i, Y_i) - \widehat{\varphi}_z(Z_i, X_i, D_i, Y_i; 0)\right)\\
  & \qquad \qquad +  \ell_{01}\left(\widehat{\varphi}^D_{z1} (Z_i, X_i, D_i, A_i, Y_i) - \widehat{\varphi}^D_{z0} (Z_i, X_i, D_i, A_i, Y_i)\right) \\
  & \qquad \qquad + (1 + \ell_{01})\hat{g}_{L_z}(X_i)\left(\widehat{\varphi}_{1-z,1}(Z_i, X_i, D_i, A_i, Y_i) - \widehat{\varphi}_{z1}(Z_i, X_i, D_i, A_i, Y_i) \right),\\
  \widehat{U}_z & = \frac{1}{n}\sum_{i=1}^n (1 + \ell_{01})\left(\widehat{\varphi}_{z1}(Z_i, X_i, D_i, A_i, Y_i) - \widehat{\varphi}_z(Z_i, X_i, D_i, Y_i; 0)\right)\\
  & \qquad \qquad +  \ell_{01} \widehat{\varphi}^D_{z1} (Z_i, X_i, D_i, A_i, Y_i) + \widehat{\varphi}^D_{z0} (Z_i, X_i, D_i, A_i, Y_i) \\
  & \qquad \qquad - (1 + \ell_{01})\hat{g}_{U_z}(X_i)\left(\widehat{\varphi}_{1-z,0}(Z_i, X_i, D_i, A_i, Y_i) - \widehat{\varphi}_{z0}(Z_i, X_i, D_i, A_i, Y_i) \right).
\end{align*}

To estimate the bounds well, we need the plugin nuisance classifier to
classify correctly which bound to use.  Systematic mis-classifications
will lead to bias.  One way to characterize the complexity of the
classification problem is via a margin condition \citep{Audibert2007}
that quantifies how often the difference between the two bounds is
small.  We formally state this margin condition here.
\begin{assumption}{\sc Margin condition.}
  \label{assum:margin}
  There exist constants $C > 0$ and $\alpha > 0$ such that:
$$\begin{aligned}
  \Pr\left(\left|(1 - m^D(1-z, X, 0))m^Y(1-z, x, 0) - (1 - m^D(z, x,
      0))m^Y(z, x, 0)\right | \leq t \right) & \leq C t^\alpha, \\
  \Pr\left(\left|(1 - m^D(1-z, X, 0))(1 - m^Y(1-z, x, 0)) - (1 - m^D(z,
      x, 0))(1 - m^Y(z, x, 0)) \right |  \leq t \right) & \leq C t^\alpha.
\end{aligned}
$$
\end{assumption}

Larger values of the margin parameter $\alpha$ imply that the
difference in the bounds is often large, and so it is easy to classify
which is tighter. Conversely, smaller values of $\alpha$ mean that the
classification problem is harder because the difference between the
bounds is often small. In the continuous case, if the covariates have a bounded
density then $\alpha >= 1$ \citep{Audibert2007}.  Margin conditions
such as this have been used when estimating partially identified
parameters \citep[e.g.][]{kennedy_sharp_iv_2020, kallus_harm_2022,
  levis_iv_2024} and for policy learning \citep[e.g.][]{Qian2011,
  Luedtke2016, DAdamo2023, benm:imai:jian:23}.  This margin condition,
along with the following additional rate conditions, yield
asymptotically normal estimates of the upper and lower bounds.
\begin{assumption} {\sc Additional rate conditions.}
  \label{assum:margin_error_rates}
  For $z=0,1$ and $a=0,1$,
  \begin{enumerate}
    \item $\left(\|m^Y(z, \cdot, 0) - \hat{m}^Y(z, \cdot, 0)\|_2  + \|m^D(z, \cdot, a) - \hat{m}^D(z,\cdot, a)\|_2 \right)\times \|\hat{e}(z,\cdot) - e(z,\cdot)\|_2 = o_p\left(n^{-1/2}\right)$, $\|m^Y(z, \cdot, 0) - \hat{m}^Y(z, \cdot, 0)\|_\infty = o_p(1)$, and $\|m^D(z, \cdot, a) - \hat{m}^D(z,\cdot, a)\|_\infty = o_p(1)$
    \item $(\|\hat{m}^D(z, \cdot, 0) - m^D(z, \cdot, 0)\|_\infty + \|\hat{m}^Y(z, \cdot, 0) - m^Y(z, \cdot, 0)\|_\infty )^{1 + \alpha} = o_p\left(n^{-\frac{1}{2}}\right)$
  \end{enumerate}
\end{assumption}

\begin{theorem}[Asymptotic Normality of Estimated Bounds]
  \label{thm:partial_id_risk_est}
  Under
  Assumptions~\ref{assum:single_blinded}--\ref{assum:margin_error_rates},
  the estimated bounds are asymptotically normal,
  $$\sqrt{n}(\widehat{L}_z - L_z) \ \stackrel{d}{\longrightarrow} N(0, V_{L_z}), \quad
  \sqrt{n}(\widehat{U}_z - U_z) \ \stackrel{d}{\longrightarrow} N(0, V_{U_z}),$$
where 
\begin{align*}
  V_{L_z} = \E[\{& (1 + \ell_{01}) (\varphi_z(Z, X, D, A, Y) - \varphi_z(Z, X, D, Y; 0)) +  \ell_{01}(\varphi^D_{z1} (Z, X, D, A, Y) - \varphi^D_{z0} (Z, X, D, A, Y))\\
   & + (1 + \ell_{01})g_{L_z}(X)\left(\varphi_{1-z,1}(Z, X, D, A, Y) - \varphi_{z}(Z, X, D, A, Y) \right) - L_z\}^2],\\
   V_{U_z} = \E[\{ & (1 + \ell_{01})\varphi_{z1}(Z, X, D, A, Y) - \varphi_z(Z, X, D, Y; 0) + \ell_{01} \varphi^D_{z1} (Z, X, D, A, Y) + \varphi^D_{z0} (Z, X, D, A, Y) \\
   & \qquad \qquad + (1 + \ell_{01})g_{U_z}(X)\left(\varphi_{1-z,0}(Z, X, D, A, Y) - \varphi_{z0}(Z, X, D, A, Y) \right)\}^2].
\end{align*}
\end{theorem}

The first rate condition in Assumption~\ref{assum:margin_error_rates}
is analogous to the rate condition in Assumption~\ref{assum:rates}, but
for the outcome and decision model conditional on the AI
recommendation $A$.  As before, in a randomized experiment the rate
condition will be satisfied because the true propensity score is
known, and we only require consistency of the outcome and decision
model estimates.

In contrast, the second condition in
Assumption~\ref{assum:margin_error_rates} requires that we can
estimate the outcome and decision models at a sufficiently fast rate
to estimate the nuisance classifiers well enough, where the margin
parameter $\alpha$ determines how fast the rate needs to be.  If the
classification task is more difficult and $\alpha$ is small, then we
will need to estimate the nuisance components at closer to the
parametric $n^{-1/2}$ rate; if the task is easier and $\alpha$ is
large, then the rate can be slower. However, note that the required
rate is always strictly slower than the parametric rate because
$\alpha > 0$.  Knowing the propensity score by design in a randomized
experiment does not remove this requirement.  In a randomized
experiment, however, we can choose what covariates to include.
Including more covariates can lead to more informative bounds, though
estimation may become more challenging.

Finally, to obtain confidence intervals via
Theorem~\ref{thm:partial_id_risk_est}, we first estimate the
asymptotic variances by taking the requisite sample averages of
estimated nuisance functions to get $\hat{V}_{L_z}$ and
$\hat{V}_{U_z}$.  We then follow \citet{imbens_confidence_2004} and
compute lower and upper $1-\alpha$ confidence intervals for the lower
and upper bounds, respectively, and create a confidence interval for
the partially identified set as
$\left[\widehat{L}_z - z_{1-\alpha}\sqrt{\widehat{V}_{L_z}/n},\
  \widehat{U}_z + z_{1-\alpha}\sqrt{\widehat{V}_{U_z}/n}\right]$.

\subsubsection{Comparing different loss functions}
\label{subsubsec:compare}

Similarly to Section~\ref{sec:human}, we can conduct a statistical
hypothesis test to examine how the preference of the AI-alone system
over human-alone (or human-with-AI) system depends on the magnitude of
loss $\ell_{01}$ assigned to false positives relative to false
negatives.  For example, to test whether the human-alone system is
preferable to the AI-alone system, the null and alternative hypotheses
are given by:
\begin{equation} 
H_0:  R_\textsc{AI} (\ell_{01}) \le R_{\textsc{Human}}(\ell_{01}),
\quad H_1: R_\textsc{AI} (\ell_{01}) > R_{\textsc{Human}}(\ell_{01}).
\label{eqn:hypothesis}
\end{equation} 
If we reject $H_0$ for a given value of $\ell_{01}$, then we would
know that the human-alone system has a lower risk than the AI-alone
system.

Since the classification risk difference is only partially identified,
however, we instead test the null hypothesis that its lower bound is
less than or equal to zero, $H_0: L_0 \le 0$ versus the alternative
hypothesis $H_1: L_0 > 0$.  If we reject this null hypothesis, then we
know that
$R_{\textsc{AI}}(\ell_{01})-R_{\textsc{Human}}(\ell_{01}) \geq L_0 >
0$, implying that the risk of the AI-alone system is likely to be
greater than that of the human-alone system and hence the latter is
preferable.  Similarly, if we reject the null hypothesis of
$H_0: U_0 \ge 0$ in favor of the alternative hypothesis
$H_1: U_0 < 0$, then we prefer the AI-alone system over the
human-alone system.  As explained above, inverting these hypothesis
tests will give us a range of loss functions under which the data
either support preferring the human-alone or AI-alone systems (there
will also be a region where the preference is ambiguous).

\section{Policy Learning}
\label{sec:optimal}

The discussion so far has focused on evaluating and comparing the
classification ability of the human-alone, AI-alone, and human-with-AI
decision-making systems.  We now consider whether these decision
making systems perform better in some cases than others, and how to
derive rules for choosing which one to use.  We first discuss learning
when to provide AI recommendations and then analyze when human
decision-makers should follow AI recommendations.  

\subsection{Learning when to provide AI recommendations}

Let $\pi: \mathcal{X} \to \{0,1\}$ be a covariate-dependent policy
that determines whether to provide the AI recommendation ($\pi(x)=1$)
or not ($\pi(x)=0$). Here, the covariate space $\mathcal{X}$ may
include the AI recommendation $A$.  We consider a class of policies
$\Pi$, each of which combines the human-alone and human-with-AI
systems.  Then, the classification risk of policy $\pi \in \Pi$ is
given by,
\begin{align*}
   R_\textsc{rec}(\ell_{01}; \pi) \ := \ &   p_{10}(D(\pi(X)))+\ell_{01}p_{01}(D(\pi(X)))\\
  = \ & R_\textsc{human}(\ell_{01}) + \E\left[\pi(X) \left
        \{p_{10}(D(1) \mid X) - p_{10}(D(0) \mid X) \right. \right.\\
  & \left.\left. \hspace{1.25in} -\ell_{01} \times \left(p_{00}(D(1) \mid X) - p_{00}(D(0) \mid X) \right)\right\}\right],
\end{align*}
where we have used Theorem~\ref{thm:z_vs_z0} to write the
classification risk in terms of observable components.

Our goal is to find an optimal policy in $\Pi$ that minimizes the
classification risk,
$$\pi^\ast_\textsc{rec} \in \argmin_{\pi \in \Pi} R_\textsc{rec}(\ell_{01}; \pi).$$
We estimate this policy by solving the following empirical risk
minimization problem with doubly-robust estimators:
\begin{equation}
  \label{eq:optimal_recommendation_rule}
  \hat{\pi}_\textsc{rec} \in \underset{\pi \in \Pi}{\argmin} \;  \frac{1}{n}\sum_{i=1}^n \pi(X_i)\left(\widehat{\varphi}_1(Z_i, X_i, D_i, Y_i; \ell_{01}) -  \widehat{\varphi}_0(Z_i, X_i, D_i, Y_i; \ell_{01})\right).
\end{equation}
The following theorem bounds the \emph{excess risk} of this learned
policy, i.e., the difference in classification risk between the best
combined decision rule $\pi^\ast_\textsc{rec}$ and the empirical rule
$\hat{\pi}_\textsc{rec}$.
\begin{theorem}[Bounding the excess risk]
  \label{thm:policy_learn_rec} \spacingset{1}
  Under
  Assumptions~\ref{assum:single_blinded}--\ref{assum:margin_error_rates},
  we have:
  \begin{align*}
    & R_\textsc{rec}(\hat{\pi}_\textsc{rec}; \ell_{01}) -
    R_\textsc{rec}(\pi^\ast_\textsc{rec}; \ell_{01})\\
    \leq &  \ C\left(\sum_{z=0}^1 \|m^Y(z, \cdot) - \hat{m}^Y(z, \cdot)\|_2 + \|m^D(z, \cdot) - \hat{m}^D(z,\cdot)\|_2)\right) \times \|\hat{e} - e\|_2\\
    & \quad  +
    \left(1 + \frac{4}{\eta}\right)(1 + \ell_{01}) \mathcal{R}_n(\Pi) + \frac{t}{\sqrt{n}},
  \end{align*}
  with probability at least $1 - 2\exp(-t^2/2)$, where
  $\mathcal{R}_n(\Pi) := \E_{X, \varepsilon}\left[\sup_{\pi \in
      \Pi}\left|\frac{1}{n}\sum_{i=1}^n \varepsilon_i
      \pi(X_i)\right|\right]$ is the population Rademacher complexity
  of the policy class $\Pi$.
\end{theorem}
Theorem~\ref{thm:policy_learn_rec} shows that the estimated policy
will have low excess risk if the nuisance components are estimated
well and the policy class is not too complex.  The first component of
the bound is related to the product-error rate seen in doubly-robust
policy learning \citep{Athey2021}.  As in Section~\ref{sec:human}, due
to the compound nature of the outcome, we can bound the excess risk in
terms of the error rates for the outcome and decision models. In a
randomized experiment, if we use the known propensity score rather than an
estimate, this term will not appear.

The analyst chooses the complexity of the policy class; flexible
decision rules will be harder to estimate than simple ones, but
simple, transparent rules are often preferred at the cost of
potentially larger risk.  As an example, if the policy class $\Pi$ has
a finite VC dimension $v$, the Rademacher complexity is
$\mathcal{R}_n(\Pi) = O(\sqrt{v/n})$ \citep[ \S5]{wainwright_2019}.

\subsection{Learning when human decision-makers should follow AI recommendations}

We next turn to learning when a human decision-maker should follow AI
recommendations.  We consider a policy $\pi$ that determines when a
human decision maker should make the decision on their own
($\pi(x) = 0$) or simply follow the AI recommendation ($\pi(x) = 1$).
The classification risk for a given policy $\pi$ is a combination of
the risk under the human-alone and the AI-alone system:
\begin{align*}
  R_\textsc{dec}(\ell_{01}; \pi) \ := \  & p_{10}(\widetilde{D})+\ell_{01}p_{01}(\widetilde{D}) \\
  = \ & R_\textsc{Human}(\ell_{01}) + \E\left[\pi(X) \left\{\emp{p_{10}(A \mid X )}  -  p_{10}(D(0) \mid X) +
  \ell_{01} (\emp{p_{01}(A \mid X )}  -   \emp{p_{01}(D(0) \mid X)}) \right\}\right].
\end{align*}
where $\widetilde{D}=A \pi(X)+D(0)(1-\pi(X))$. As in
Section~\ref{sec:partial}, the expected risk has several
unidentifiable terms highlighted in blue.  Therefore, we take a
conservative approach and consider finding the decision rule that
minimizes the worst case excess risk relative to the human-alone
system:
\[
\pi^\ast_\textsc{dec} \in \underset{\pi \in \Pi}{\argmin} \; \E[\pi(X) U_0(X)],
\]
where $U_0(x)$ is the upper bound on the conditional risk difference
derived in Theorem~\ref{thm:partial_id_risk}.  This worst-case
criterion requires strong evidence that the AI-alone decision is
better before (hypothetically) overriding human decisions. It takes
the human-alone decision as the baseline, and only follows the AI
recommendation if it will lead to a lower loss even in the worst case.
Alternative criteria such as minimax regret are also possible, but we
leave a thorough exploration of them to future work.

We again take an empirical risk minimization approach and find a
policy $\hat{\pi}_\textsc{dec}$ that minimizes our estimate of the
worst-case excess risk by solving: 
\begin{equation}
  \label{eq:optimal_ai_rule}
  \begin{aligned}
    \hat{\pi}_\textsc{dec} \in \underset{\pi \in \Pi}{\argmin} &
    \frac{1}{n}\sum_{i=1}^n \pi(X_i) \left [ (1 +
      \ell_{01})\left(\widehat{\varphi}_{01}(Z_i, X_i, D_i, A_i, Y_i) -
      \widehat{\varphi}_0(Z_i, X_i, D_i, Y_i; 0)\right) \right.\\
  & \qquad \qquad \qquad +  \ell_{01} \widehat{\varphi}^D_{01} (Z_i, X_i, D_i, A_i, Y_i) + \widehat{\varphi}^D_{00} (Z_i, X_i, D_i, A_i, Y_i) \\
  & \qquad \qquad \qquad \left. - (1 + \ell_{01})\hat{g}_{U_0}(X_i)\left(\widehat{\varphi}_{10}(Z_i, X_i, D_i, A_i, Y_i) - \widehat{\varphi}_{00}(Z_i, X_i, D_i, A_i, Y_i) \right) \right].
  \end{aligned}
\end{equation}
Due to the lack of point identification, we bound the excess
\emph{worst-case} risk of the estimated policy
$\hat{\pi}_\textsc{dec}$ versus the population policy
$\pi^\ast_\textsc{dec}$.
\begin{theorem}[Bounding the excess worst-case
  risk]\label{thm:policy_learn_dec} \spacingset{1}
  Under
  Assumptions~\ref{assum:single_blinded}--\ref{assum:margin_error_rates},
  we have:
  \begin{align*}
    & \E[(\hat{\pi}_\textsc{dec}(X) - \pi^\ast_\textsc{dec})U_{z}(X)]
    \\ \leq \ &   C\left ( \sum_{z' = 0}^1\|m^Y(z', \cdot, 0) - \hat{m}^Y(z', \cdot, 0)\|_2 + \|m^D(z', \cdot, 0) - \hat{m}^D(z',\cdot, 0)\|_2 \right. \\
    & + \quad  \|m^Y(z, \cdot) - \hat{m}^Y(z, \cdot)\|_2 + \|m^D(z, \cdot) - \hat{m}^D(z,\cdot)\|_2 + \|m^D(z, \cdot, 1) - \hat{m}^D(z,\cdot, 1)\|_2 \bigg)\\
    & \qquad \times \|\hat{e} - e\|_2 + 2C (\|\hat{m}^D(\cdot, \cdot, 0) - m^D(\cdot, \cdot, 0)\|_\infty + \|\hat{m}^Y(\cdot, \cdot, 0) - m^Y(\cdot, \cdot, 0)\|_\infty )^{1 + \alpha}\\
    &  +
    \left(1 + \frac{2}{\eta}\right)(4 + 6 \ell_{01}) \mathcal{R}_n(\Pi) + \frac{t}{\sqrt{n}},
  \end{align*}
  with probability at least $1 - 2\exp(-t^2/2)$.
\end{theorem}
Theorem~\ref{thm:policy_learn_dec} shows that the error in the
nuisance components and the complexity of the policy class control the
excess worst-case risk.  In contrast to the case of the estimated
AI-recommendation provision rule $\hat{\pi}_\textsc{rec}$, however,
the knowledge of the propensity score is not sufficient for the
estimated AI-alone decision rule $\hat{\pi}_\textsc{rec}$ to have low
excess risk. As with Theorem~\ref{thm:partial_id_risk_est}, because
optimizing for the worst-case excess risk involves estimating sharp
bounds (and the nuisance classifier $g_{U_z}(x)$), the estimation
error of the outcome and decision models enter into the bound alone.

\section{Empirical Analysis}
\label{sec:empirical}

We now use the proposed methodology to analyze the experiment
described in Section~\ref{subsec:Dane}, focusing on evaluating three
different decision-making systems --- human-alone, PSA-alone, and
human-with-PSA systems.  Since we analyze the interim data, the
results reported below should be interpreted as an illustration of the
proposed methodology rather than the final analysis results from our
RCT.  

\subsection{Setup}

The dataset comprises a total of $1,891$ first arrest cases, in which
judges made decisions on whether to impose a signature bond or cash
bail: $D_i = 1$ if the judge set a cash bail and $D_i = 0$ if the
judge set a signature bond. We dichotomize the PSA-DMF recommendation:
$A_i = 1$ if it recommends a cash bail and $A_i = 0$ if the
recommendation is a signature bond.  We use the following case-level
covariates $X$: gender (male or female), race (white or non-white),
the interaction between gender and race, age, inputs for the PSA
recommendation including variables for current/past charges and prior
convictions, three PSA risk scores, and the overall PSA-DMF
recommendation (abbreviated as the PSA recommendation). 

The provision of the PSA recommendation is randomized, with $Z_i = 1$
denoting the randomized treatment indicator. In other words, the
decision-maker in the treatment group is a human judge who is given the PSA
recommendation, whereas in the control group, the same human judge
makes decisions without the PSA recommendation.  We use the true propensity
score, $e(z,x) = 0.5$, for the estimation throughout the analysis.
The results remain essentially identical when we use the estimated
propensity score.

We analyze three binary outcomes: FTA, NCA, and NVCA, where $Y_i = 1$
indicates an incidence of misconduct, and $Y_i = 0$ indicates the
absence of such an outcome. Among the $1,891$ cases, $40\%$ are white
male arrestees, $39\%$ are non-white males, $13\%$ are white females,
and $8\%$ are non-white females.  The proportions of negative outcomes
are $18\%$, $25\%$, and $5\%$ for FTA, NCA, and NVCA, respectively.

\subsection{PSA recommendations do not improve human decisions}

We begin by estimating the impact of providing PSA recommendations on
human decisions. Specifically, we use the method described in
Section~\ref{sec:human} to estimate the difference in
misclassification rates between decisions made by the human judge
alone and those made with the PSA recommendation. Recall that the
misclassification rate is equivalent to the symmetric loss function,
i.e., $R(1)$.

\begin{figure}[p]
    \centering
    \includegraphics[width = \textwidth]{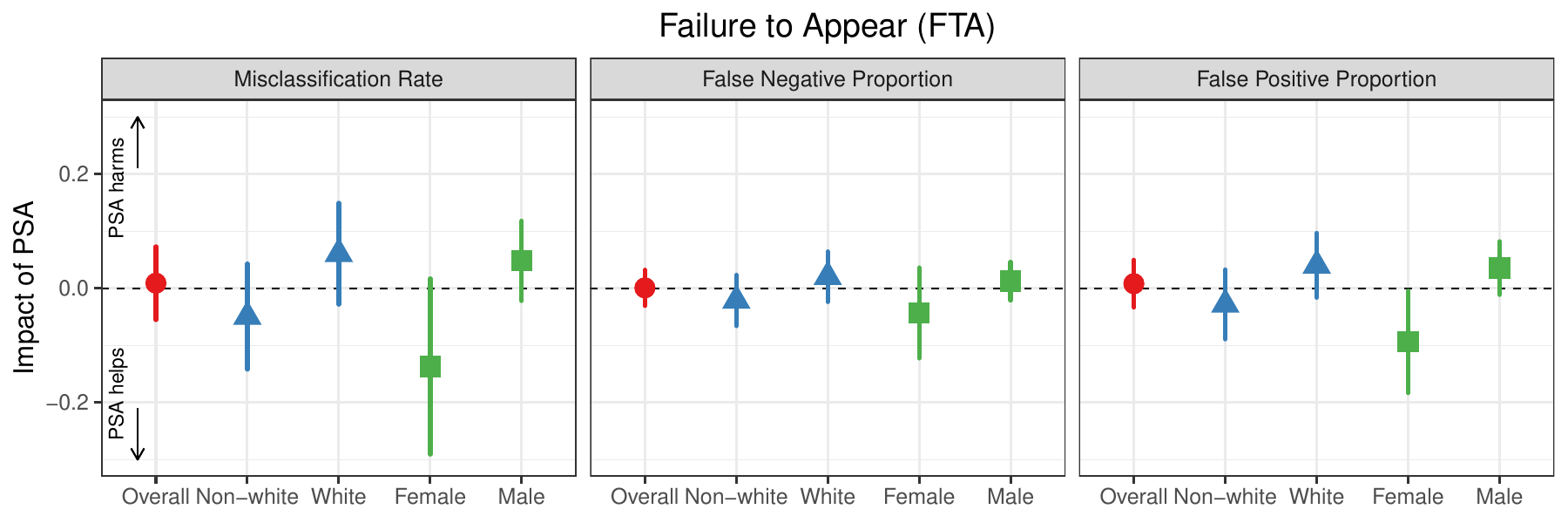}
    \includegraphics[width = \textwidth]{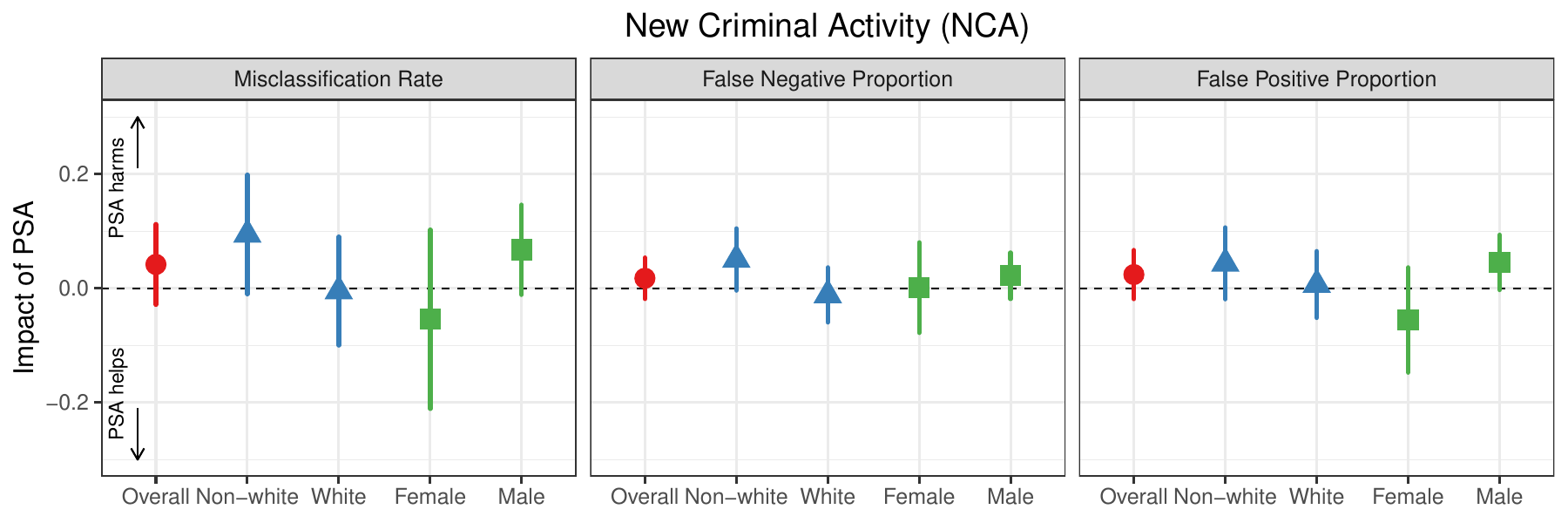}
    \includegraphics[width = \textwidth]{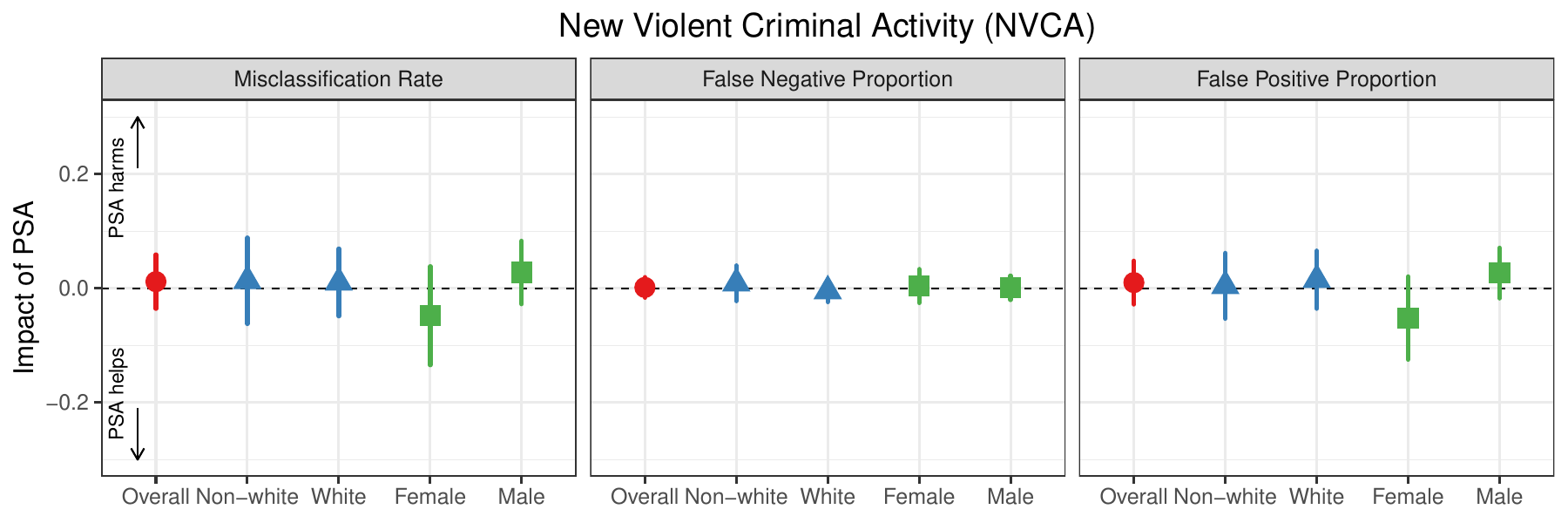}
    \caption{Estimated Impact of PSA Recommendations on Human
      Decisions. The figure shows how PSA recommendations change a
      human judge's cash bail decisions in terms of misclassification
      rate, false negative proportion, and false positive
      proportion. Each panel presents the overall and
      subgroup-specific results for a different outcome variable. For
      each quantity of interest, we report a point estimate and its
      corresponding 95\% confidence interval for the overall sample
      (red circle), non-white and white subgroups (blue triangle), and
      female and male subgroups (green square). The results show that
      the PSA recommendations do not significantly improve the judge's
      decisions.}
    \label{fig:human_point}
\end{figure}

Figure~\ref{fig:human_point} presents the estimated impact of PSA
recommendations on human decisions in terms of the misclassification
rate, proportion of false negatives, and proportion of false
positives.  We find that the PSA recommendations do not significantly
improve the judge's decisions. Indeed, none of the classification risk
differences between the judge's decisions with and without the PSA
recommendations are statistically significant though the estimates are
relatively precise.

In Appendix~\ref{app:additional}
(Figures~\ref{fig:human_override_A1}~and~\ref{fig:human_override_A0}),
we examine how often the judge correctly overrides the PSA
recommendations by conducting a subgroup analysis based on whether or
not the PSA recommends cash bail.  For example, for the cases with $A_i=1$, a
true negative implies that the judge issues a signature bond decision
against the PSA recommendation of cash bail to an arrestee who would
not commit misconduct if released on their own recognizance.  By
comparing the true negative proportions between the human-alone and
human-with-PSA system, we can adjust for the baseline disagreement
between the human and PSA decisions and examine the judge's ability to
correctly override the PSA recommendations.  The estimates are
qualitatively similar to those presented in
Figure~\ref{fig:human_point}, implying that the judge does not
necessarily override the PSA recommendations correctly.

\subsection{PSA-alone decisions are less accurate than human decisions}

Next, we compare the classification performance of PSA-alone decisions
with that of human decisions, using the proposed methodology described
in Section~\ref{sec:partial}.  Specifically, we estimate the upper and
lower bounds of the differences in the misclassification rate, false
negative proportion, and false positive proportion between PSA-alone
and human decisions (with and without PSA recommendations).

\begin{figure}[p]
    \centering
    \includegraphics[width = \textwidth]{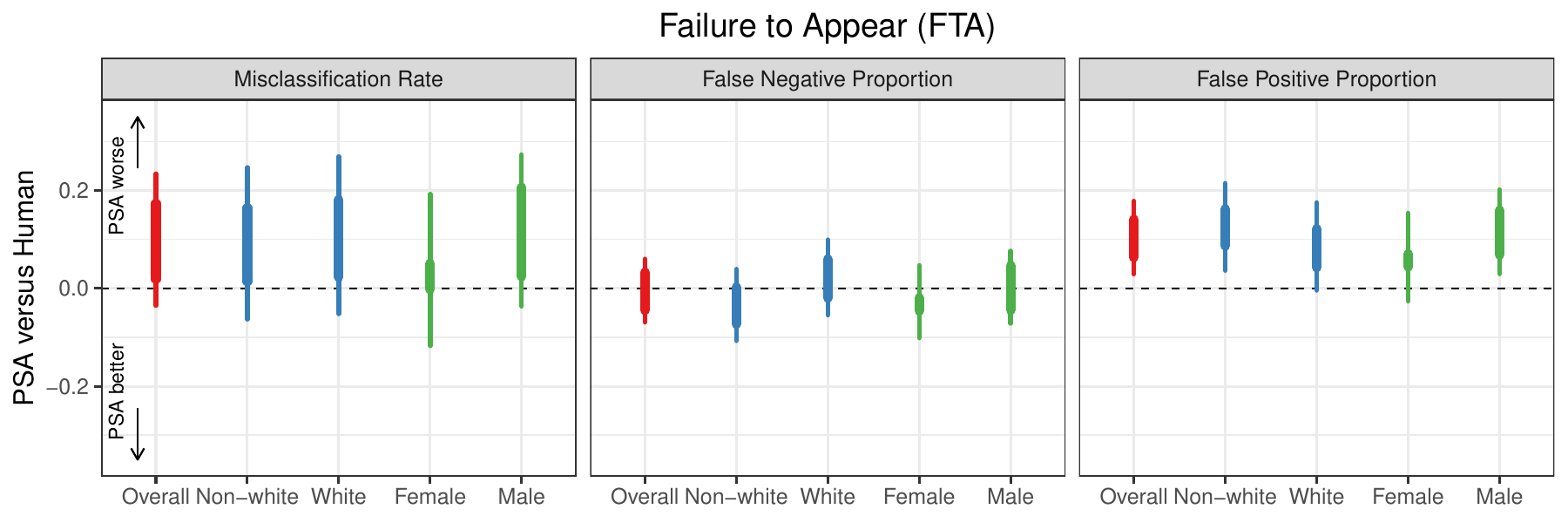}
    \includegraphics[width = \textwidth]{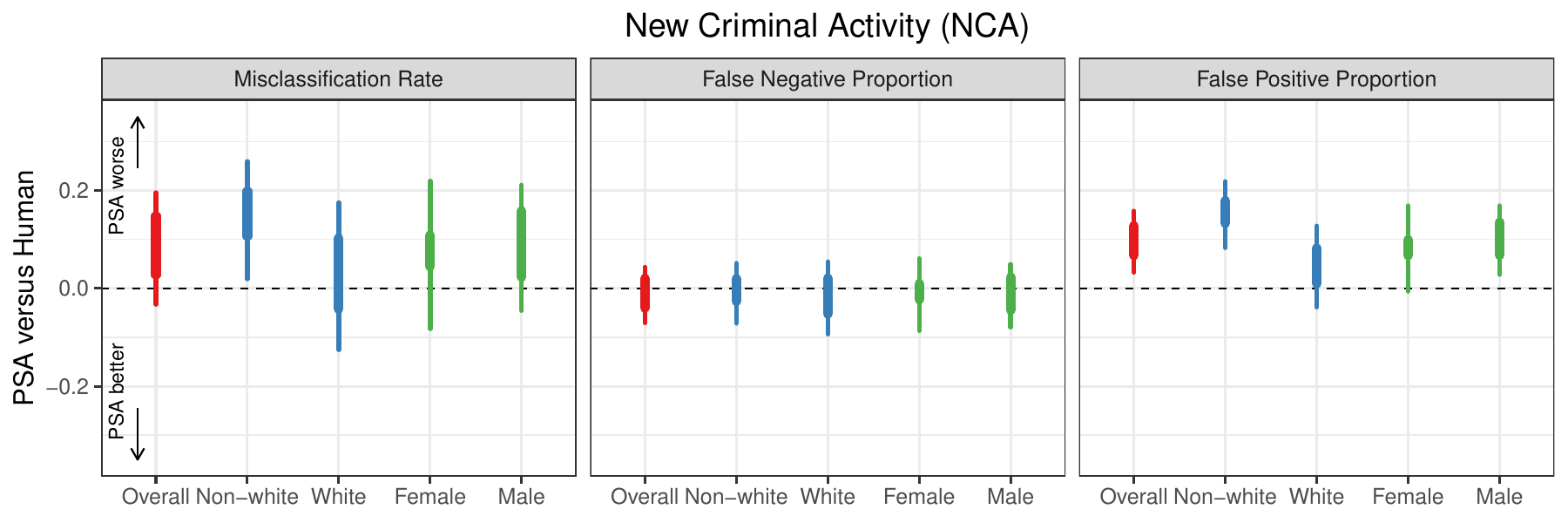}
    \includegraphics[width = \textwidth]{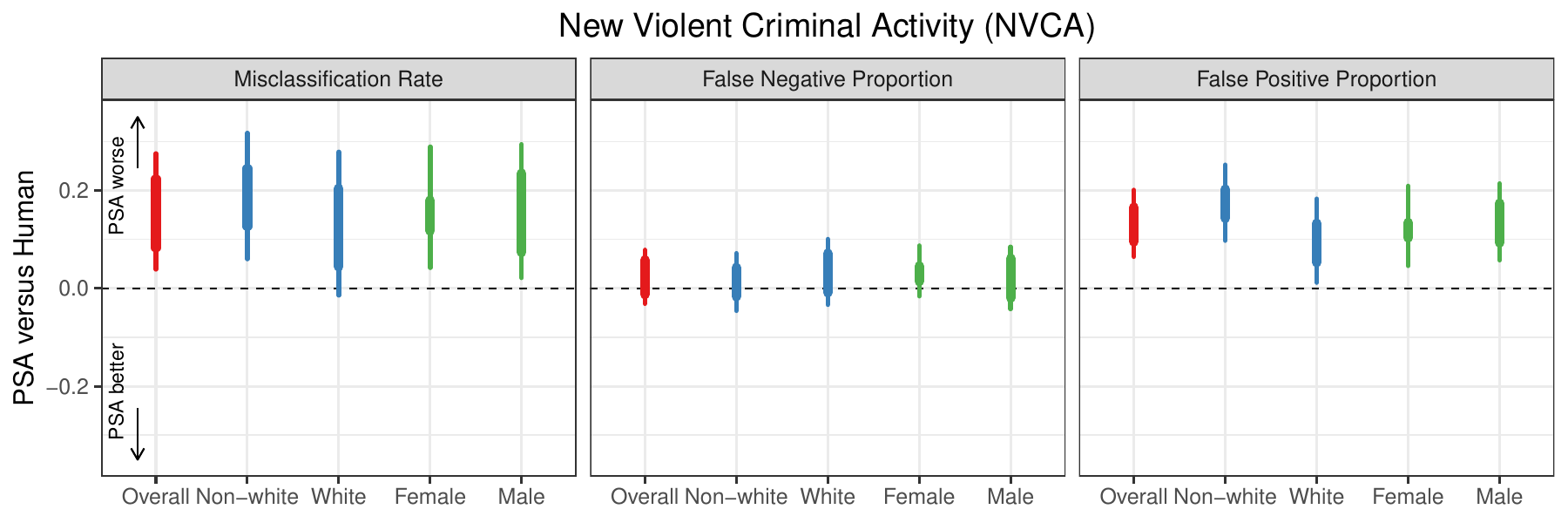}
    \caption{Estimated Bounds on Difference in Classification Ability
      between PSA-alone and Human-alone Decisions. The figure shows the
      misclassification rate, false negative proportion, and false
      positive proportion. Each panel presents the overall and
      subgroup-specific results for a different outcome variable. For
      each quantity of interest, we report estimated bounds (thick
      lines) and their corresponding 95\% confidence interval (thin
      lines) for the overall sample (red), non-white and white
      subgroups (blue), and female and male subgroups (green). The
      results indicate that PSA-alone decisions are less accurate than
      human judge's decisions in terms of the false positive
      proportion.}
    \label{fig:ai_human_bounds}
\end{figure}

Figure~\ref{fig:ai_human_bounds} shows that the PSA-alone system
results in substantially higher false positive proportions, in
comparison to the judge's decisions.  For NVCA, the misclassification
rates are also significantly higher for the PSA-alone system than the human-alone system.
Qualitatively similar
results are obtained when comparing the PSA-alone and human-with-PSA
systems (see Figure~\ref{fig:ai_humanAI_bounds} in
Appendix~\ref{app:additional}).
This finding holds across the
overall sample and within every subgroup for each outcome variable,
although the results are not statistically significant for the FTA and
NCA outcomes among white/female arrestees.  

We find that the PSA system is generally harsher than the human judge,
resulting in a greater number of unnecessary cash bail decisions
across subgroups and different outcome variables.  This pattern is
particularly strong for non-white arrestees, where the PSA-alone
system exhibits a large false positive proportion. For this group of
arrestees, the NCA/NVCA misclassification rate under the PSA-alone
system is statistically significantly greater than the human-alone
system.

For the false negative proportion, the differences between the
PSA-alone and human-alone systems are generally not statistically
significant. In sum, our analysis shows that when compared to human
decisions, the PSA-alone decision-making system is likely to result in
a higher number of false positives, corresponding to a higher number of
unnecessary bail decisions.

\subsection{Human decisions are preferred over PSA-alone system when
  the cost of false positives is high}

Next, we analyze how one's loss function determines their
preference over different decision-making systems.  Specifically, we
follow the discussion at the end of Section~\ref{sec:partial} and
invert the hypothesis test using the bounds on the difference in
classification risk derived in Theorem~\ref{thm:partial_id_risk}.
This analysis allows us to estimate the range of the loss of false
positives ($\ell_{01}$), relative to the loss of false negatives, that
would lead us to prefer human decisions over the PSA-alone system.

We invert the hypothesis test shown in Equation~\eqref{eqn:hypothesis}
over the range of values, $\ell_{01} \in [0.01, 100]$ using the $0.05$
significance level. Specifically, for each candidate value of
$\ell_{01}$, we conduct two one-sided hypothesis tests; one
right-tailed and the other left-tailed, using the $z$-score of
the lower and upper bounds of the difference in misclassification
rates, respectively.  If the left-tailed test null hypothesis,
$H_0: U_0 \ge 0$, is rejected (and thus the right-tailed test is not),
we can conclude that the classification risk of the human-alone system
is likely to be greater than that of PSA-alone system, suggesting a
preference for the PSA-alone system over human decisions. Conversely,
if the right-tailed test of null hypothesis $H_0: L_0 \le 0$ is
rejected, it indicates a preference for the human-alone system. If
neither test is rejected, we conclude that the preference between the
decision-making systems is ambiguous.

\begin{figure}[p]
    \centering
    \includegraphics[width = \textwidth]{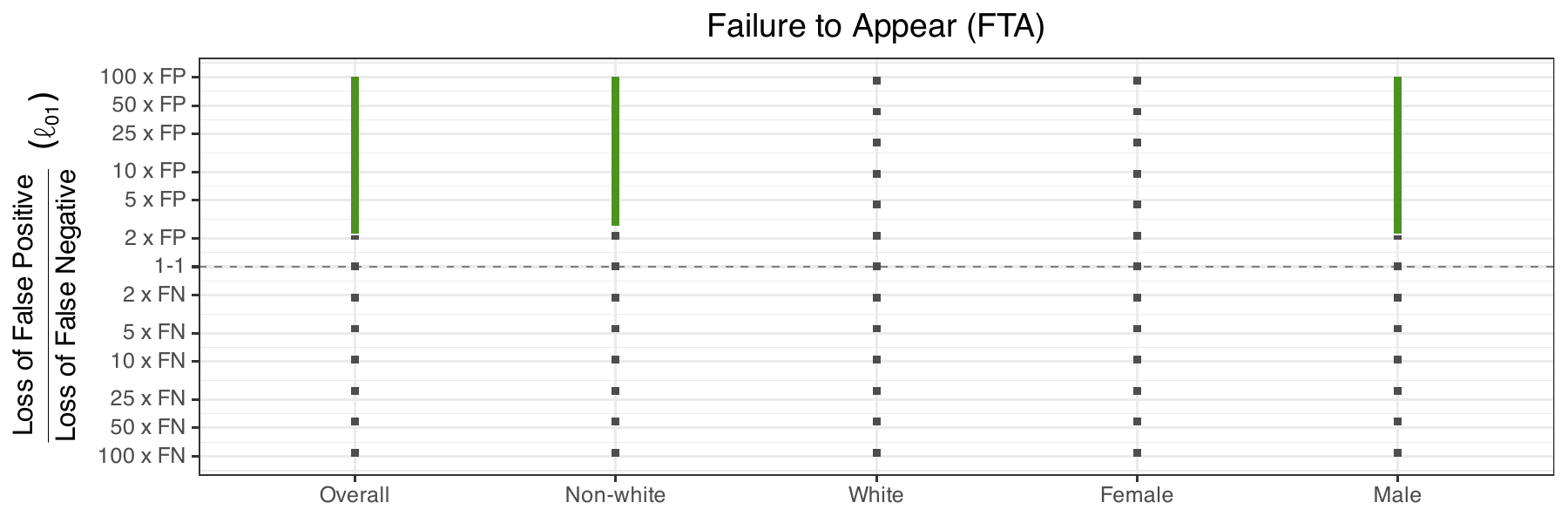}
    \includegraphics[width = \textwidth]{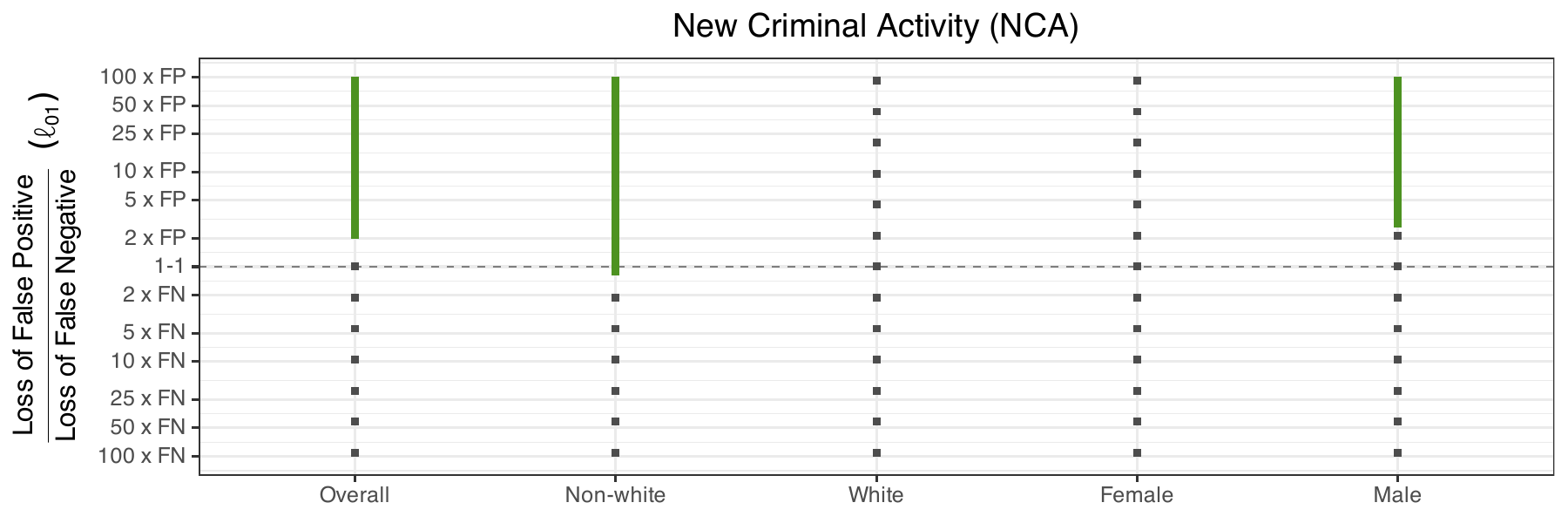}
    \includegraphics[width = \textwidth]{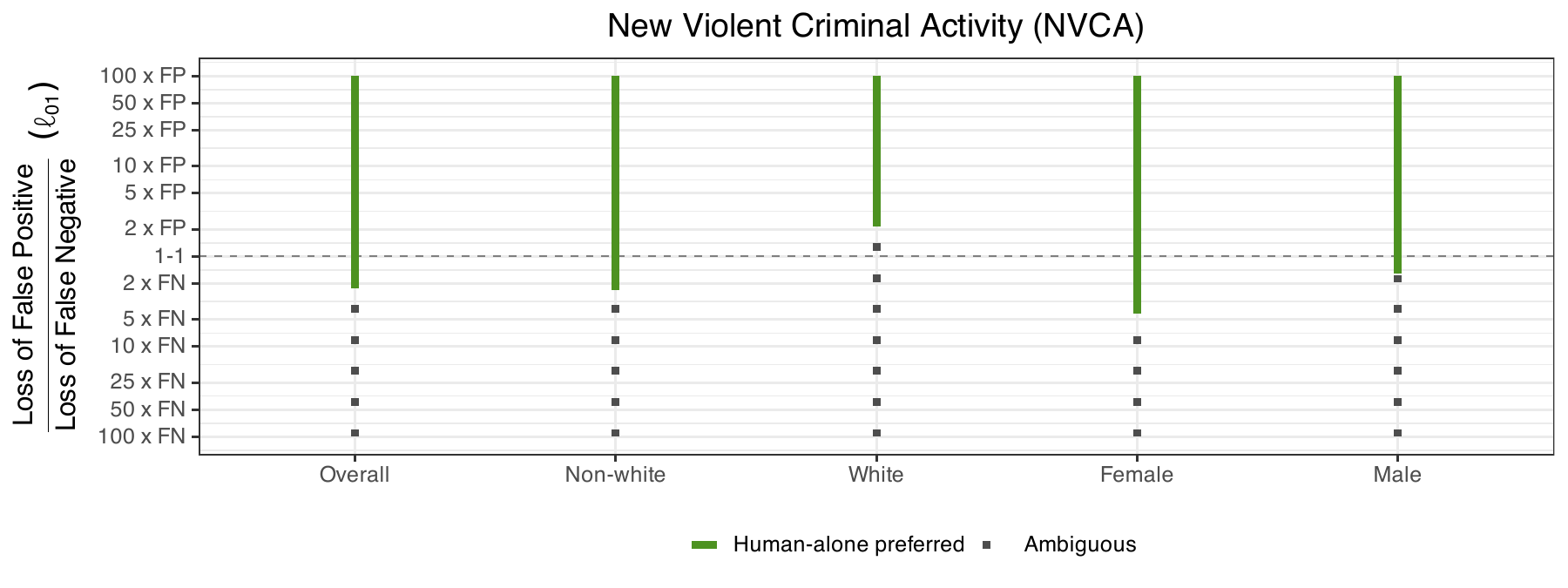}
    \caption{Estimated Preference for Human-alone Decisions over
      PSA-alone Decision-Making System. The figure illustrates the
      range of the ratio of the loss between false positives and false
      negatives, $\ell_{01}$, for which one decision-making system is
      preferable over the other. A greater value of the ratio
      $\ell_{01}$ implies a greater loss of false positive relative to
      that of false negative. Each panel displays the overall and
      subgroup-specific results for different outcome variables. For
      each quantity of interest, we show the range of $\ell_{01}$ that
      corresponds to the preferred decision-making system; human-alone (green lines), and ambiguous (dotted
      lines). The results suggest that the human-alone system is
      preferred over the PSA-alone system when the loss of false
      positive is about the same as or greater than that of false
      negative.
      The PSA-alone system is never preferred within the specified range of $\ell_{01}$.}
    \label{fig:pref_human}
\end{figure}

Figure~\ref{fig:pref_human} shows that the human-alone system is
preferred over the PSA-alone system when the loss of false positive is
about the same as or greater than that of false negatives. For
instance, with FTA as an outcome, the human-alone system is preferred
over the PSA-alone system when $\ell_{01} \ge 1.63$.  Similar results
are observed across various outcomes and subgroups, particularly for
NVCAs, where we find that the human-alone system is preferred when
$\ell_{01} \ge 0.35$.  Exceptions are white and female arrestees with
either FTA or NCA as an outcome, where we observe ambiguous results.
Qualitatively similar results are also obtained when comparing the
PSA-alone and human-with-PSA systems (see Figure~\ref{fig:pref_humanAI}
in Appendix~\ref{app:additional}), though we find that the results are
ambiguous for white and male arrestees with either FTA and NCA as an
outcome.

\subsection{Optimally combining PSA recommendations with human decisions}

Next, we investigate how to integrate optimally PSA recommendations
into human decisions.  We apply the methods developed in
Section~\ref{sec:optimal}, and estimate (1) an optimal policy for
determining when to provide the PSA recommendations, and (2) an
optimal decision rule regarding when a human decision-maker should
follow the PSA recommendations. Specifically, we solve the empirical
risk minimization problems outlined in
Equations~\eqref{eq:optimal_recommendation_rule}~and~\eqref{eq:optimal_ai_rule},
respectively.  Here, we consider a policy class $\Pi$ that maps FTA,
NCA, and NVCA risk scores to a binary decision, subject to a
monotonicity constraint (either increasing or decreasing).  For
example, under an increasing monotonicity constraint, if any of the
three risk scores increases, the resulting binary rule should not
decrease (i.e., a decision of $D=1$ cannot then be altered to
$D = 0$). We consider the NCA as the outcome.

\begin{figure}[t]
    \centering
    \includegraphics[width = 0.45\textwidth]{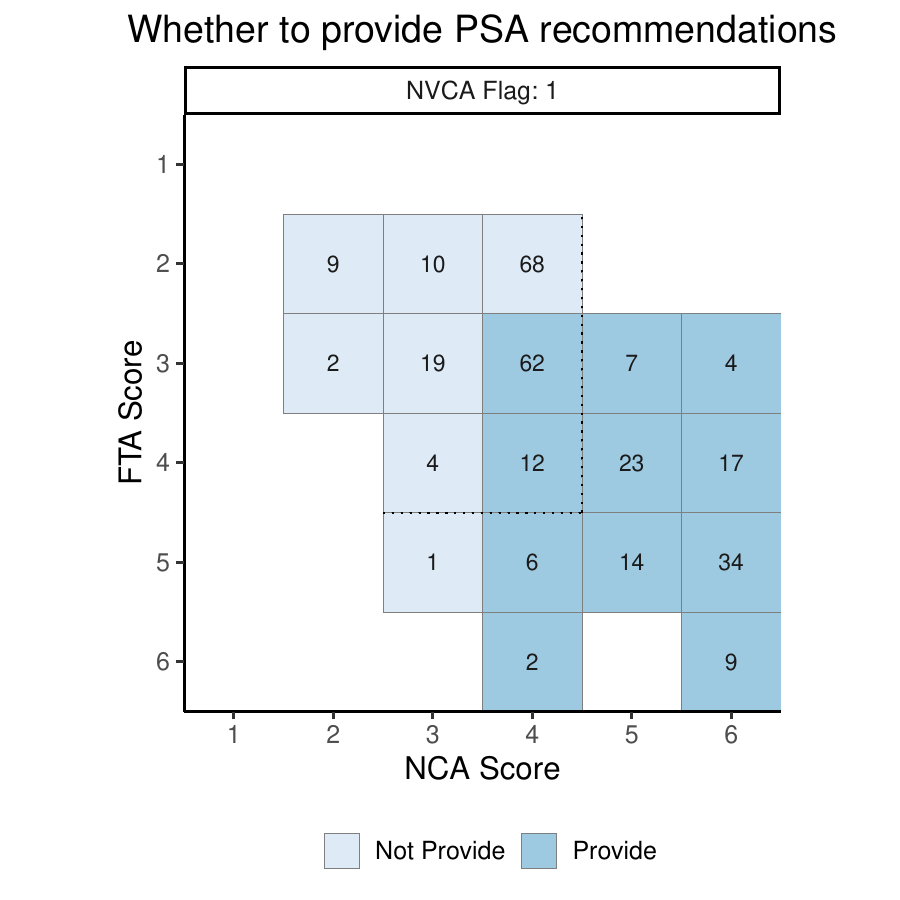}
    \includegraphics[width = 0.45\textwidth]{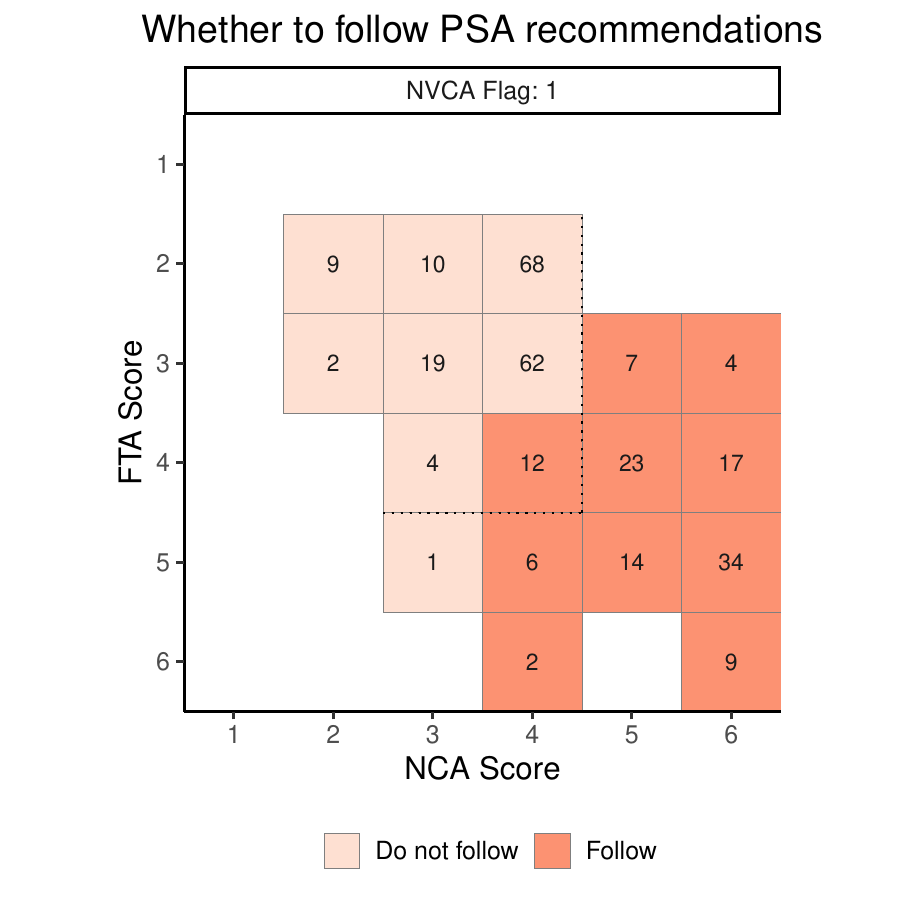}
    \caption{Optimally Combining Human Decisions with PSA
      Recommendations when NCA is the outcome.  The left plot shows an
      estimated optimal policy for determining when to provide PSA
      recommendations to a human judge.  The right plot shows an
      estimated optimal policy regarding when a human decision-maker
      should follow PSA recommendations.  Each shaded area represents
      the optimal policy for specific combinations of risk scores:
      light shading indicates a decision rule of ``not provide''
      (left) or ``do not follow'' (right), while dark shading
      indicates a decision rule of ``provide'' (left) or ``follow''
      (right).  Unshaded areas represent combinations of risk scores
      that are not possible.  The number of observations for each
      combination is also shown.}
      \label{fig:optimal_policy}
\end{figure}

The left plot of Figure~\ref{fig:optimal_policy} shows when one should
provide PSA recommendations to a human judge (indicated by dark blue)
under this monotonicity constraint.  We find that providing an PSA
recommendation is advisable only when the FTA and NCA risk scores are
relatively high and the NVCA flag is 1.  The right plot of the figure
shows when a human judge should follow the PSA recommendation
(indicated by dark orange), again under the monotonicity constraint.
Our finding suggests that unless the FTA and NCA risk scores are
relatively high and the NVCA flag is 1, a human judge should not
follow the PSA recommendations and should instead use their judgement.

Taken together, both the provision and decision rules suggest not to provide
PSA input for the vast majority of cases. Rather, the PSA should only be
provided (or followed) in extreme cases where arrestees have many
risk factors present.
We emphasize that the magnitude of improvement due to these optimal
policies is relatively small.  Our analysis shows that under the monotonicity constraint, for the NCA as
an outcome, the optimal provision of PSA recommendations results in a decrease of $0.01$ in the
misclassification rate when compared to not providing PSA
recommendations for any case.  Similarly, the optimal policy regarding when
to follow PSA recommendations, under the monotonicity constraint,
results in a decrease of $0.004$ in the worst-case difference in the
misclassification rate relative to not following PSA recommendations
at all.  See Table~\ref{tbl:optimal} in Appendix~\ref{app:additional}
for the results of the estimated worst-case value of the empirical
risk minimization problem under the optimal policies. 

\subsection{AI-alone decisions are less accurate than human decisions}
\label{subsec:LLM}
A natural question that arises is whether or not an alternative algorithmically generated risk score would perform better than the PSA. To evaluate this, we compare the classification ability of AI-alone decisions with
that of human decisions, using an open-source large language model, Llama3 \cite{touvron2023llama} to generate the AI decisions. We
prompt the model to provide binary recommendations--whether to impose
cash bail or release--based on the same set of PSA inputs for each
arrestee.  Here, we use deterministic decoding so that Llama3 always
returns the same output for a given prompt
\citep{imai2024causalrepresentationlearninggenerative}.  The prompt we
use is as follows:
\begin{quote} \singlespacing
    You are a judge in Dane County, Madison, Wisconsin and are asked to decide whether or not an arrestee should be released on their own recognizance or be required to post a cash bail.
    If you think the risk of unnecessary incarceration is too high, then the arrestee should receive own recognizance release. On the other hand, you should assign cash bail if the following risks are too high: the risk of failure to appear at subsequent court dates, the risk of engaging in new criminal activity, and the risk of engaging in new violent criminal activity.
    You are provided with the following 12 characteristics about an arrestee (label - description): \textbf{[description of PSA inputs]}.
    This arrestee has the following characteristics (label - arrestee's value): 
    \textbf{[arrestee's PSA inputs]}.
    Should this arrestee be released on their own recognizance or given cash bail?
    Please provide your answer in binary form (0 for released on their own recognizance and 1 for cash bail), followed by a detailed explanation of your decision. Example: {binary decision} - {reason}.
\end{quote}

\begin{figure}[p]
    \centering
    \includegraphics[width = \textwidth]{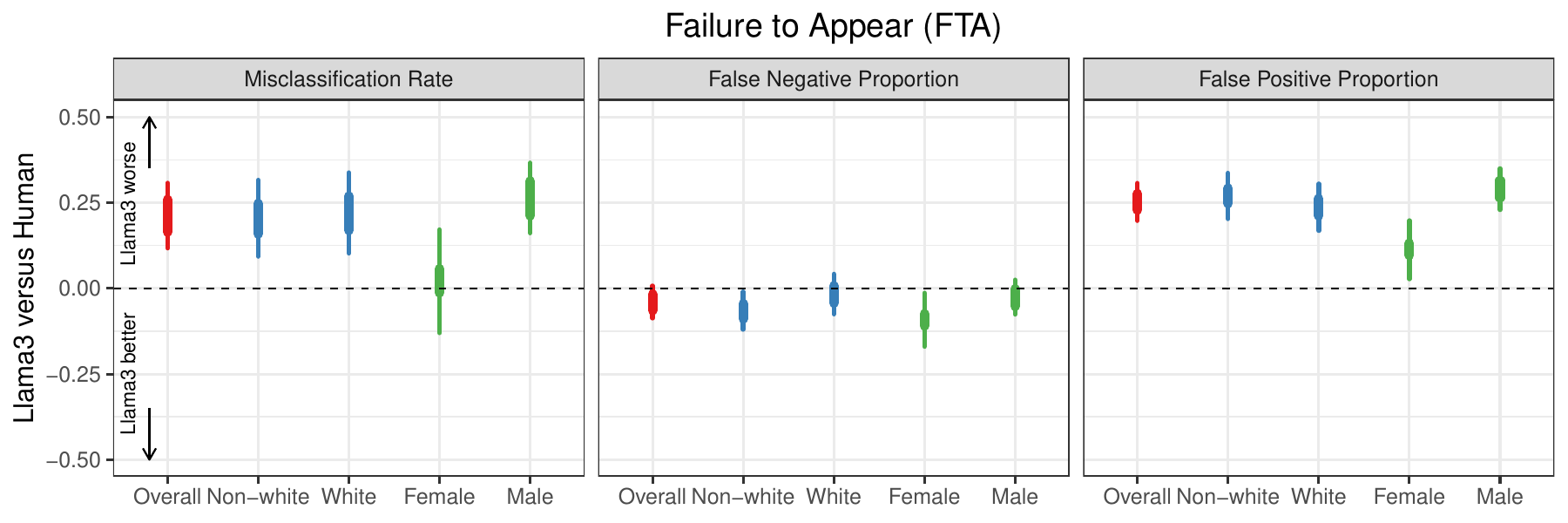}
    \includegraphics[width = \textwidth]{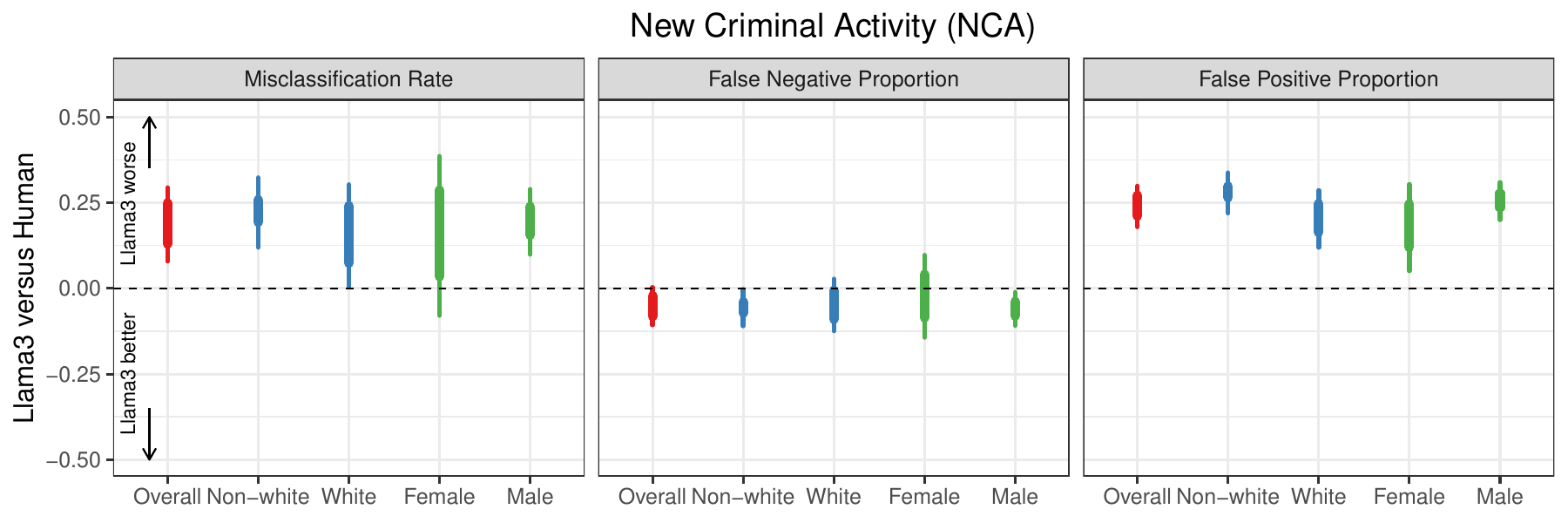}
    \includegraphics[width = \textwidth]{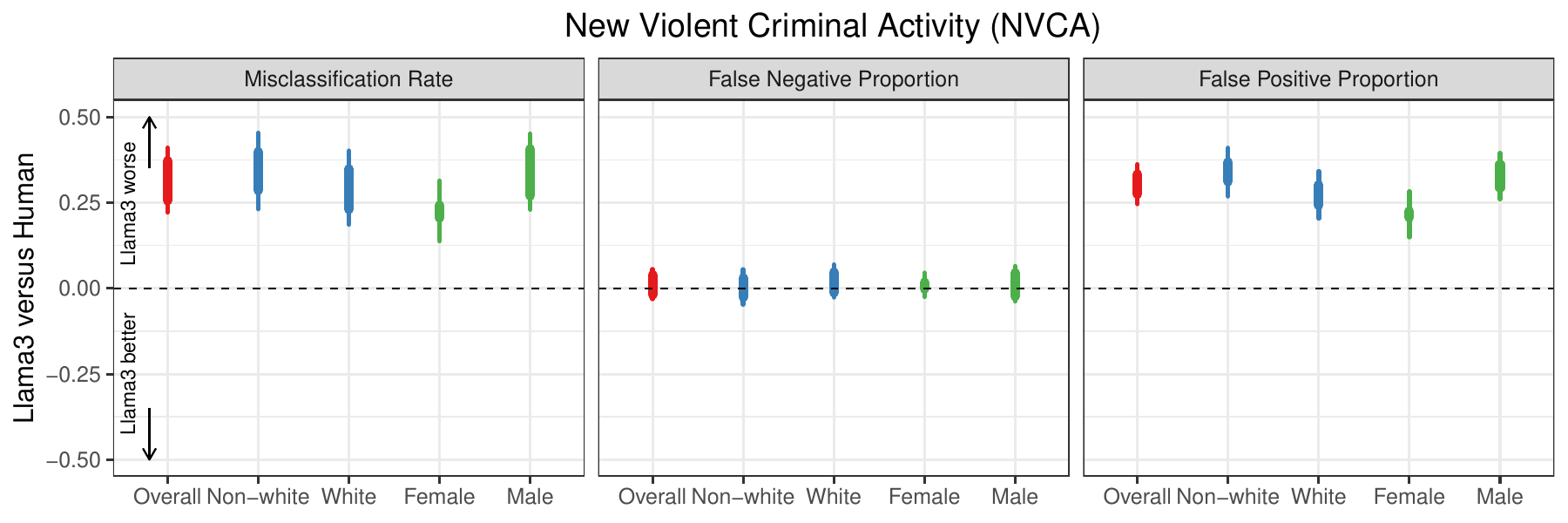}
    \caption{Estimated Bounds on the Difference in Classification
      Ability between Llama3 and Human-alone Decisions. The figure
      shows the differences in terms of misclassification rate, false
      negative proportion, and false positive proportion. Each panel
      presents the overall and subgroup-specific results for one of
      the three outcome variables. For each quantity of interest, we
      report estimated bounds (thick lines) and their corresponding
      95\% confidence interval (thin lines) for the overall sample
      (red), non-white and white subgroups (blue), and female and male
      subgroups (green). The results indicate that Llama3 decisions
      are less accurate than human judge's decisions in terms of the
      false positive proportion and the overall misclassification rate.}
    \label{fig:llama_human_bounds}
\end{figure}

Following the proposed methodology described in
Section~\ref{sec:partial}, we estimate the upper and lower bounds of
the differences in the misclassification rate, false negative
proportion, and false positive proportion between Llama3 and human
decisions.  Figure~\ref{fig:llama_human_bounds} shows that the
Llama3 decisions result in substantially higher false positive
proportions, in comparison to the judge's decisions.  This finding
holds across the overall sample and within every subgroup for each
outcome variable.  The results suggest that the recommendations by
Llama3 are generally harsher than the human judge, yielding a greater
number of unnecessary cash bail decisions across subgroups and
different outcome variables.  For the false negative proportion, the
differences between the Llama3 and human-alone decisions are generally
not statistically significant.

\section{Concluding Remarks}

We have introduced a new methodological framework that provides a way
to evaluate empirically the performance of three different
decision-making systems: human-alone, AI-alone, and human-with-AI
systems.  We formalized the classification ability of each
decision-making system using standard confusion matrices based on
potential outcomes.  We then showed that under single-blinded and
unconfounded treatment assignment, we can directly identify the
differences in classification ability between human decision-makers
with and without AI recommendations. Furthermore, we derived partial
identification bounds to compare the differences in classification
ability between an AI-alone and human decision-making systems and
separately evaluate the performance of each system.

To illustrate the power of the proposed methodological framework, we
apply our framework to the data from our own RCT whose goal is to
evaluate the impact of the PSA risk assessment scores on a judge's
decision to impose a cash bail or release arrestees on their own
recognizance. We compare the human-alone and human-with-PSA decisions
and find little to no impact of providing PSA recommendations.  Our
comparison of the human decision-maker with the PSA-alone system
suggests, based on the baseline potential outcome and around 40\% of
the RCT's enrolled cases, that PSA-alone decisions may underperform as
compared to human decisions, resulting in a greater proportion of
unnecessarily harsh decisions.  All together, these empirical findings
suggest that integrating algorithmic recommendations into judicial
decision-making warrants careful consideration and rigorous empirical
evaluation.

There are several exciting future methodological research directions
that can build on the current work. The proposed methodological
framework can be extended to common settings where decisions and
outcomes are non-binary.  Another possible extension is the
consideration of joint potential outcomes as done in
\cite{imai2023experimental} and the dynamic settings where multiple
decisions and outcomes are observed over time.  Finally, the proposed
methodology and its extensions can be applied to a variety of real
world settings where AI decision-making systems have been integrated
or considered for future use.

\bigskip
\addcontentsline{toc}{section}{\refname}

\bibliographystyle{chicago}
\bibliography{bibliography}

\begin{thebibliography}{}

\bibitem[\protect\citeauthoryear{Albright}{Albright}{2019}]{albright2019if}
Albright, A. (2019).
\newblock If you give a judge a risk score: evidence from kentucky bail
  decisions.
\newblock {\em Law, Economics, and Business Fellows’ Discussion Paper
  Series\/}~{\em 85}.

\bibitem[\protect\citeauthoryear{Angelova, Dobbie, and Yang}{Angelova
  et~al.}{2023}]{angelova2023algorithmic}
Angelova, V., W.~S. Dobbie, and C.~Yang (2023).
\newblock Algorithmic recommendations and human discretion.
\newblock Technical report, National Bureau of Economic Research.

\bibitem[\protect\citeauthoryear{Arnold, Dobbie, and Hull}{Arnold
  et~al.}{2021}]{Arnold2021_algo}
Arnold, D., W.~Dobbie, and P.~Hull (2021).
\newblock {Measuring Racial Discrimination in Algorithms}.
\newblock {\em AEA Papers and Proceedings\/}~{\em 111}, 49--54.

\bibitem[\protect\citeauthoryear{Arnold, Dobbie, and Hull}{Arnold
  et~al.}{2022}]{Arnold2022_bail_decisions}
Arnold, D., W.~Dobbie, and P.~Hull (2022).
\newblock {Measuring Racial Discrimination in Bail Decisions}.
\newblock {\em American Economic Review\/}~{\em 112\/}(9), 2992--3038.

\bibitem[\protect\citeauthoryear{Athey and Wager}{Athey and
  Wager}{2021}]{Athey2021}
Athey, S. and S.~Wager (2021).
\newblock Policy learning with observational data.
\newblock {\em Econometrica\/}~{\em 89\/}(1), 133--161.

\bibitem[\protect\citeauthoryear{Audibert and Tsybakov}{Audibert and
  Tsybakov}{2007}]{Audibert2007}
Audibert, J.~Y. and A.~B. Tsybakov (2007).
\newblock Fast learning rates for plug-in classifiers.
\newblock {\em Annals of Statistics\/}~{\em 35\/}(2), 608--633.

\bibitem[\protect\citeauthoryear{Barocas, Hardt, and Narayanan}{Barocas
  et~al.}{2017}]{barocas2017fairness}
Barocas, S., M.~Hardt, and A.~Narayanan (2017).
\newblock Fairness in machine learning.
\newblock {\em Nips tutorial\/}~{\em 1}, 2017.

\bibitem[\protect\citeauthoryear{Ben-Michael, Imai, and Jiang}{Ben-Michael
  et~al.}{2023}]{benm:imai:jian:23}
Ben-Michael, E., K.~Imai, and Z.~Jiang (2023).
\newblock Policy learning with asymmetric counterfactual utilities.
\newblock {\em Journal of the American Statistical Association\/}, Forthcoming.

\bibitem[\protect\citeauthoryear{Berk}{Berk}{2017}]{berk2017impact}
Berk, R. (2017).
\newblock An impact assessment of machine learning risk forecasts on parole
  board decisions and recidivism.
\newblock {\em Journal of Experimental Criminology\/}~{\em 13}, 193--216.

\bibitem[\protect\citeauthoryear{Berk, Sorenson, and Barnes}{Berk
  et~al.}{2016}]{berk2016forecasting}
Berk, R.~A., S.~B. Sorenson, and G.~Barnes (2016).
\newblock Forecasting domestic violence: A machine learning approach to help
  inform arraignment decisions.
\newblock {\em Journal of empirical legal studies\/}~{\em 13\/}(1), 94--115.

\bibitem[\protect\citeauthoryear{Binns, Van~Kleek, Veale, Lyngs, Zhao, and
  Shadbolt}{Binns et~al.}{2018}]{binns2018s}
Binns, R., M.~Van~Kleek, M.~Veale, U.~Lyngs, J.~Zhao, and N.~Shadbolt (2018).
\newblock 'it's reducing a human being to a percentage' perceptions of justice
  in algorithmic decisions.
\newblock In {\em Proceedings of the 2018 Chi conference on human factors in
  computing systems}, pp.\  1--14.

\bibitem[\protect\citeauthoryear{Cai, Winter, Steiner, Wilcox, and Terry}{Cai
  et~al.}{2019}]{cai2019hello}
Cai, C.~J., S.~Winter, D.~Steiner, L.~Wilcox, and M.~Terry (2019).
\newblock " hello ai": uncovering the onboarding needs of medical practitioners
  for human-ai collaborative decision-making.
\newblock {\em Proceedings of the ACM on Human-computer Interaction\/}~{\em
  3\/}(CSCW), 1--24.

\bibitem[\protect\citeauthoryear{Cheng and Chouldechova}{Cheng and
  Chouldechova}{2022}]{cheng2022heterogeneity}
Cheng, L. and A.~Chouldechova (2022).
\newblock Heterogeneity in algorithm-assisted decision-making: A case study in
  child abuse hotline screening.
\newblock {\em Proceedings of the ACM on Human-Computer Interaction\/}~{\em
  6\/}(CSCW2), 1--33.

\bibitem[\protect\citeauthoryear{Chouldechova and Roth}{Chouldechova and
  Roth}{2020}]{chouldechova2020snapshot}
Chouldechova, A. and A.~Roth (2020).
\newblock A snapshot of the frontiers of fairness in machine learning.
\newblock {\em Communications of the ACM\/}~{\em 63\/}(5), 82--89.

\bibitem[\protect\citeauthoryear{Corbett-Davies and Goel}{Corbett-Davies and
  Goel}{2018}]{corbett2018measure}
Corbett-Davies, S. and S.~Goel (2018).
\newblock The measure and mismeasure of fairness: A critical review of fair
  machine learning.
\newblock {\em arXiv preprint arXiv:1808.00023\/}.

\bibitem[\protect\citeauthoryear{Coston, Rambachan, and Chouldechova}{Coston
  et~al.}{2021}]{coston2021characterizing}
Coston, A., A.~Rambachan, and A.~Chouldechova (2021).
\newblock Characterizing fairness over the set of good models under selective
  labels.
\newblock In {\em International Conference on Machine Learning}, pp.\
  2144--2155. PMLR.

\bibitem[\protect\citeauthoryear{D'Adamo}{D'Adamo}{2023}]{DAdamo2023}
D'Adamo, R. (2023).
\newblock {Orthogonal Policy Learning Under Ambiguity}.

\bibitem[\protect\citeauthoryear{Dobbie, Goldin, and Yang}{Dobbie
  et~al.}{2018}]{dobbie2018effects}
Dobbie, W., J.~Goldin, and C.~S. Yang (2018).
\newblock The effects of pre-trial detention on conviction, future crime, and
  employment: Evidence from randomly assigned judges.
\newblock {\em American Economic Review\/}~{\em 108\/}(2), 201--240.

\bibitem[\protect\citeauthoryear{Goel, Rao, and Shroff}{Goel
  et~al.}{2016}]{goel2016personalized}
Goel, S., J.~M. Rao, and R.~Shroff (2016).
\newblock Personalized risk assessments in the criminal justice system.
\newblock {\em American Economic Review\/}~{\em 106\/}(5), 119--123.

\bibitem[\protect\citeauthoryear{Greiner, Halen, Stubenberg, and Chistopher
  L.~Griffen}{Greiner et~al.}{2020}]{greiner2020dane}
Greiner, D.~J., R.~Halen, M.~Stubenberg, and J.~Chistopher L.~Griffen (2020).
\newblock Randomized control trial evaluation of the implementation of the
  psa-dmf system in dane county.
\newblock Technical report, Access to Justice Lab, Harvard Law School.

\bibitem[\protect\citeauthoryear{Guerdan, Coston, Wu, and Holstein}{Guerdan
  et~al.}{2023}]{guerdan2023ground}
Guerdan, L., A.~Coston, Z.~S. Wu, and K.~Holstein (2023).
\newblock Ground (less) truth: A causal framework for proxy labels in
  human-algorithm decision-making.
\newblock In {\em Proceedings of the 2023 ACM Conference on Fairness,
  Accountability, and Transparency}, pp.\  688--704.

\bibitem[\protect\citeauthoryear{Hoffman, Kahn, and Li}{Hoffman
  et~al.}{2018}]{hoffman2018discretion}
Hoffman, M., L.~B. Kahn, and D.~Li (2018).
\newblock Discretion in hiring.
\newblock {\em The Quarterly Journal of Economics\/}~{\em 133\/}(2), 765--800.

\bibitem[\protect\citeauthoryear{Imai and Jiang}{Imai and
  Jiang}{2023}]{imai2023principal}
Imai, K. and Z.~Jiang (2023).
\newblock Principal fairness for human and algorithmic decision-making.
\newblock {\em Statistical Science\/}~{\em 38\/}(2), 317--328.

\bibitem[\protect\citeauthoryear{Imai, Jiang, Greiner, Halen, and Shin}{Imai
  et~al.}{2023a}]{imai2023experimental}
Imai, K., Z.~Jiang, D.~J. Greiner, R.~Halen, and S.~Shin (2023a).
\newblock Experimental evaluation of algorithm-assisted human decision-making:
  Application to pretrial public safety assessment.
\newblock {\em Journal of the Royal Statistical Society Series A: Statistics in
  Society\/}~{\em 186\/}(2), 167--189.

\bibitem[\protect\citeauthoryear{Imai, Jiang, Greiner, Halen, and Shin}{Imai
  et~al.}{2023b}]{imai2023data}
Imai, K., Z.~Jiang, D.~J. Greiner, R.~Halen, and S.~Shin (2023b).
\newblock Replication data for: Experimental evaluation of algorithm-assisted
  human decision-making: Application to pretrial public safety assessment.
\newblock {\em Harvard Dataverse, DOI: 10.7910/DVN/L0NHQU\/}.

\bibitem[\protect\citeauthoryear{Imbens and Manski}{Imbens and
  Manski}{2004}]{imbens_confidence_2004}
Imbens, G.~W. and C.~F. Manski (2004).
\newblock Confidence {Intervals} for {Partially} {Identified} {Parameters}.
\newblock {\em Econometrica\/}~{\em 72\/}(6), 1845--1857.

\bibitem[\protect\citeauthoryear{Kallus}{Kallus}{2022}]{kallus_harm_2022}
Kallus, N. (2022).
\newblock What's the {Harm}? {Sharp} {Bounds} on the {Fraction} {Negatively}
  {Affected} by {Treatment}.
\newblock In {\em 36th {Conference} on {Neural} {Information} {Processing}
  {Systems}}.

\bibitem[\protect\citeauthoryear{Kennedy}{Kennedy}{2023}]{kennedy_semiparametric_2023}
Kennedy, E.~H. (2023, January).
\newblock Semiparametric doubly robust targeted double machine learning: a
  review.
\newblock arXiv:2203.06469 [stat].

\bibitem[\protect\citeauthoryear{Kennedy, Balakrishnan, and G'Sell}{Kennedy
  et~al.}{2020}]{kennedy_sharp_iv_2020}
Kennedy, E.~H., S.~Balakrishnan, and M.~G'Sell (2020, August).
\newblock Sharp instruments for classifying compliers and generalizing causal
  effects.
\newblock {\em The Annals of Statistics\/}~{\em 48\/}(4), 2008--2030.
\newblock Publisher: Institute of Mathematical Statistics.

\bibitem[\protect\citeauthoryear{Kleinberg, Lakkaraju, Leskovec, Ludwig, and
  Mullainathan}{Kleinberg et~al.}{2018}]{kleinberg2018human}
Kleinberg, J., H.~Lakkaraju, J.~Leskovec, J.~Ludwig, and S.~Mullainathan
  (2018).
\newblock Human decisions and machine predictions.
\newblock {\em The quarterly journal of economics\/}~{\em 133\/}(1), 237--293.

\bibitem[\protect\citeauthoryear{Lai, Kankanhalli, and Ong}{Lai
  et~al.}{2021}]{lai2021human}
Lai, Y., A.~Kankanhalli, and D.~Ong (2021).
\newblock Human-ai collaboration in healthcare: A review and research agenda.

\bibitem[\protect\citeauthoryear{Lakkaraju, Kleinberg, Leskovec, Ludwig, and
  Mullainathan}{Lakkaraju et~al.}{2017}]{lakkaraju2017selective}
Lakkaraju, H., J.~Kleinberg, J.~Leskovec, J.~Ludwig, and S.~Mullainathan
  (2017).
\newblock The selective labels problem: Evaluating algorithmic predictions in
  the presence of unobservables.
\newblock In {\em Proceedings of the 23rd ACM SIGKDD International Conference
  on Knowledge Discovery and Data Mining}, pp.\  275--284.

\bibitem[\protect\citeauthoryear{Levis, Kennedy, and Keele}{Levis
  et~al.}{2024}]{levis_iv_2024}
Levis, A.~W., E.~H. Kennedy, and L.~Keele (2024, February).
\newblock Nonparametric identification and efficient estimation of causal
  effects with instrumental variables.
\newblock arXiv:2402.09332 [stat].

\bibitem[\protect\citeauthoryear{Luedtke and {van der Laan}}{Luedtke and {van
  der Laan}}{2016}]{Luedtke2016}
Luedtke, A.~R. and M.~J. {van der Laan} (2016).
\newblock {Statistical inference for the mean outcome under a possibly
  non-unique optimal treatment strategy}.
\newblock {\em Annals of Statistics\/}~{\em 44\/}(2), 713--742.

\bibitem[\protect\citeauthoryear{Manski}{Manski}{2007}]{manski2007}
Manski, C.~F. (2007).
\newblock {\em Identification for Prediction and Decision}.
\newblock Cambridge, MA: Harvard University Press.

\bibitem[\protect\citeauthoryear{Miller and Maloney}{Miller and
  Maloney}{2013}]{miller2013practitioner}
Miller, J. and C.~Maloney (2013).
\newblock Practitioner compliance with risk/needs assessment tools: A
  theoretical and empirical assessment.
\newblock {\em Criminal Justice and Behavior\/}~{\em 40\/}(7), 716--736.

\bibitem[\protect\citeauthoryear{Mitchell, Potash, Barocas, D'Amour, and
  Lum}{Mitchell et~al.}{2021}]{mitchell2021algorithmic}
Mitchell, S., E.~Potash, S.~Barocas, A.~D'Amour, and K.~Lum (2021).
\newblock Algorithmic fairness: Choices, assumptions, and definitions.
\newblock {\em Annual Review of Statistics and Its Application\/}~{\em 8},
  141--163.

\bibitem[\protect\citeauthoryear{Neyman}{Neyman}{1923}]{neyman1923application}
Neyman, J. (1923).
\newblock On the application of probability theory to agricultural experiments.
  essay on principles.
\newblock {\em Ann. Agricultural Sciences\/}, 1--51.

\bibitem[\protect\citeauthoryear{Qian and Murphy}{Qian and
  Murphy}{2011}]{Qian2011}
Qian, M. and S.~A. Murphy (2011).
\newblock {Performance guarantees for individualized treatment rules}.
\newblock {\em The Annals of Statistics\/}~{\em 39\/}(2), 1180--1210.

\bibitem[\protect\citeauthoryear{Rambachan}{Rambachan}{2021}]{Rambachan2021}
Rambachan, A. (2021).
\newblock {Identifying prediction mistakes in observational data}.

\bibitem[\protect\citeauthoryear{Rambachan, Coston, and Kennedy}{Rambachan
  et~al.}{2022}]{rambachan2022counterfactual}
Rambachan, A., A.~Coston, and E.~Kennedy (2022).
\newblock Counterfactual risk assessments under unmeasured confounding.
\newblock {\em arXiv preprint arXiv:2212.09844\/}.

\bibitem[\protect\citeauthoryear{Robins, Rotnitzky, and Ping~Zhao}{Robins
  et~al.}{1994}]{Robins1994}
Robins, J.~M., A.~Rotnitzky, and L.~Ping~Zhao (1994).
\newblock Estimation of {Regression} {Coefficients} {When} {Some} {Regressors}
  are not {Always} {Observed}.
\newblock {\em Journal of the American Statistical Association\/}~{\em 89427},
  846--866.

\bibitem[\protect\citeauthoryear{Rubin}{Rubin}{1974}]{rubin1974estimating}
Rubin, D.~B. (1974).
\newblock Estimating causal effects of treatments in randomized and
  nonrandomized studies.
\newblock {\em Journal of Educational Psychology\/}~{\em 66\/}(5), 688.

\bibitem[\protect\citeauthoryear{Skeem, Scurich, and Monahan}{Skeem
  et~al.}{2020}]{skeem2020impact}
Skeem, J., N.~Scurich, and J.~Monahan (2020).
\newblock Impact of risk assessment on judges’ fairness in sentencing
  relatively poor defendants.
\newblock {\em Law and human behavior\/}~{\em 44\/}(1), 51.

\bibitem[\protect\citeauthoryear{Stevenson and Doleac}{Stevenson and
  Doleac}{2022}]{stevenson2022algorithmic}
Stevenson, M.~T. and J.~L. Doleac (2022).
\newblock Algorithmic risk assessment in the hands of humans.
\newblock {\em Available at SSRN 3489440\/}.

\bibitem[\protect\citeauthoryear{Wainwright}{Wainwright}{2019}]{wainwright_2019}
Wainwright, M.~J. (2019).
\newblock {\em High-Dimensional Statistics: A Non-Asymptotic Viewpoint}.
\newblock Cambridge Series in Statistical and Probabilistic Mathematics.
  Cambridge University Press.

\end{thebibliography}

\clearpage 
\appendix
\onehalfspacing

\setcounter{page}{1}
\setcounter{equation}{0}
\setcounter{figure}{0}
\setcounter{assumption}{0}
\setcounter{theorem}{0}
\setcounter{lemma}{0}
\setcounter{section}{0}
\renewcommand {\theequation} {S\arabic{equation}}
\renewcommand {\thefigure} {S\arabic{figure}}
\renewcommand {\thetheorem} {S\arabic{assumption}}
\renewcommand {\thetheorem} {S\arabic{theorem}}
\renewcommand {\theproposition} {S\arabic{proposition}}
\renewcommand {\thelemma} {S\arabic{lemma}}
\renewcommand {\thesection} {S\arabic{section}}

\begin{center}
  \huge {\bf Supplementary Appendix}
\end{center}

\section{Generic Loss Functions} \label{app:general_case}

As a generic loss function, we can define separate weights for true
positives $\ell_{11}$, true negatives $\ell_{00}$ and false positive
$\ell_{01}$ so that the expected loss is given by
$R(\ell_{00}, \ell_{01}, \ell_{11}) = \ell_{10} q_{10} + \ell_{01}
q_{01} + \ell_{11} q_{11} + \ell_{00} q_{00}$ with a proper
normalization constraint such as $\ell_{10}=1$.  All of the quantities
we have considered in the main text are special cases of this generic
loss function. For example, the difference in risk between the
human-with-AI and human-alone systems is given by,
$$\begin{aligned}
& R_{\textsc{human+AI}}(\ell_{00}, \ell_{01}, \ell_{11})-R_{\textsc{human}}(\ell_{00}, \ell_{01}, \ell_{11}) \\
= \ & p_{10}(D(1)) - p_{10}(D(0)) + \ell_{01} (p_{01}(D(1)) -
p_{01}(D(0))) + \ell_{11} \left(p_{11}(D(1)) - p_{11}(D(0)) \right) \\
& + \ell_{00} \left(p_{00}(D(1)) - p_{00}(D(0)) \right).
\end{aligned}$$ 

Now, recall that although the false positive proportion under each
system $p_{01}(D(z))$ is not identified, the difference between
$p_{01}(D(1))$ and $p_{01}(D(0))$ is identifiable as
$p_{01}(D(1)) - p_{01}(D(0)) = p_{00}(D(0)) - p_{00}(D(1))$.  We can
similarly point identify the difference in true positive proportions
as $p_{11}(D(1)) - p_{11}(D(0)) = p_{10}(D(0)) - p_{10}(D(1))$. So,
following the argument for Theorem~\ref{thm:z_vs_z0}, we can point
identify the difference in risk with the generic loss function as:
$$\begin{aligned} & R_{\textsc{human+AI}}(\ell_{00}, \ell_{01}, \ell_{11})-R_{\textsc{human}}(\ell_{00}, \ell_{01}, \ell_{11})\\ = \ & (1 - \ell_{11}) \left(p_{10}(D(1)) - p_{10}(D(0))\right) + (\ell_{00} - \ell_{01}) \left(p_{00}(D(1)) - p_{00}(D(0)) \right).\end{aligned}$$ 

We can similarly evaluate the human-with-AI vs AI-alone and
human-alone vs AI-alone systems by following the partial
identification arguments in Section~\ref{sec:partial}, and evaluate
the decision-making systems independently directly using the results
shown in Section~\ref{sec:separate_eval}.

\section{Classification Risk of a Generic Decision-making System}
\label{app:Dstar}

In Section~\ref{sec:partial}, we focus on the evaluation of an
AI-alone decision-making system based on the specific AI
recommendation $A$ used in the experiment.  For completeness, here we
show how to evaluate the classification ability of any alternative
decision-making system $D^\ast$ in itself that satisfies the following
conditional independence $D^\ast \indep Y(0)\mid A, X$.

First, we consider a deterministic decision rule
$D^*= f(a,x) \in \{0,1\}$.  Then, the classification risk of this
decision-making system $D^\ast$ can be written as,
\begin{eqnarray*}
R(\ell_{01}; D^\ast)&=& \E\left[(1-f(A,X)) \Pr(Y(0)= 1\mid A,X) + \ell_{01} f(A,X) \Pr(Y(0)= 0 \mid A,X)\right].
\end{eqnarray*}
While this classification risk is not identifiable, we can obtain an
optimal policy by minimizing the upper bound (i.e., the worst case) of
the classification risk over the class of policies
$f \in \mathcal{F}$.  The sharp bounds are derived below (see
Theorem~\ref{thm:partial_Dstar} for more general results with a
stochastic policy).
\begin{theorem} {\sc (Sharp Bounds on the classification risk of a
    generic deterministic decision-making system)}
  \label{thm:partial_Dstar_deterministic} \spacingset{1} Consider a deterministic decision
making system $D^*= f(a,x) \in \{0,1\}$.
Then, the classification
risk of this decision-making system $D^\ast$ can be written as,
\begin{eqnarray*}
R(\ell_{01}; D^\ast)&=& \E\left[(1-f(A,X)) \Pr(Y(0)= 1\mid A,X) + \ell_{01} f(A,X) \Pr(Y(0)= 0 \mid A,X)\right].
\end{eqnarray*}
The sharp bounds on its
  classification risk $R(\ell_{01}; D^\ast)$ are given by:
\begin{eqnarray*}
  R(\ell_{01};D^\ast)&\in & 
  \Bigg[ \E\Big[(1 - f(A,X))\max_{z^\prime} \Pr(Y=1,D=0\mid A, X,Z=z^\prime)\\
  &&\qquad+ \ell_{01}f(A,X)\max_{z^\prime} \Pr(Y=0,D=0\mid A, X,Z=z^\prime)\Big],\\
  &&\quad \E\Big[\left(1 - f(A,X)\right)\big\{1-\max_{z^\prime}\Pr(Y=0,D=0\mid A, X,Z=z^\prime)\big\} \\
  &&\qquad + \ell_{01}f(A,X) \big\{
  1 - \max_{z^\prime} \Pr(Y=1,D=0\mid A, X,Z=z^\prime)
  \big\}\Big]\Bigg].
\end{eqnarray*}
\end{theorem}

We can generalize this result to include a stochastic decision-making
system $D^\ast$ that satisfies the following conditional independence
$D^\ast \indep Y(0)\mid A, X$.  Define such a decision-making system
as $f(a,x) := \Pr(D^*= 1 \mid A = a, X =x)$. The classification risk
of this decision-making system $D^\ast$ can be written as,
\begin{eqnarray*}
R(\ell_{01}; D^\ast)&=& \E\left[(1-f(A,X)) \Pr(Y(0)= 1\mid A,X) + \ell_{01} f(A,X) \Pr(Y(0)= 0 \mid A,X)\right].
\end{eqnarray*}
The sharp bounds are given by the following theorem.
\begin{theorem} {\sc (Sharp Bounds on the classification risk of a
    generic stochastic decision-making system)}
  \label{thm:partial_Dstar} \spacingset{1} Consider the decision
  making system $f(a,x) := \Pr(D^*= 1 \mid A = a, X =x)$ that
  satisfies the conditional independence relation,
  $D^\ast \indep Y(0)\mid X,A$.  The sharp bounds on its
  classification risk $R(\ell_{01}; D^\ast)$ are given by:
\begin{eqnarray*}
  R(\ell_{01};D^\ast)&\in & 
  \Bigg[ \E\Big[\ell_{01}\cdot f(A,X) 
  + \left\{1-(1+\ell_{01})f(A,X)\right\} \big[g_{f}(A,X)
  \max_{z^\prime} \Pr(Y=1,D=0\mid A, X,Z=z^\prime) \\
  & &\qquad +(1-g_{f}(A,X))
  \{1-\max_{z^\prime} \Pr(Y=0,D=0\mid A, X,Z=z^\prime)\} \big]\Big],\\
  & &\quad\E\Big[\ell_{01}\cdot f(A,X) 
  + \left\{1-(1+\ell_{01})f(A,X)\right\} \big[g_{f}(A,X)
  \{1-\max_{z^\prime} \Pr(Y=0,D=0\mid A, X,Z=z^\prime)\} \\
  & &\qquad +(1-g_{f}(A,X))
  \max_{z^\prime} \Pr(Y=1,D=0\mid A, X,Z=z^\prime) \big]\Big] \Bigg]
\end{eqnarray*}
where $g_{f}(a,x) = \1\{1-(1+\ell_{01})f(a,x)\geq0\}$.
\end{theorem}

\section{Useful Lemmas}
\label{app:lemmas}

We present two useful lemmas used to derive sharp bounds on
classification risks and their differences.  These lemmas provide
sharp bounds on the two key unidentifiable quantities
$\theta_a:=\Pr(Y(0)=1,D=1,A=a)$ and
$\xi_{az} :=\Pr(Y(0) = 1, D(z) = 1, A=a)$.
\begin{lemma} \label{cor:sharp_bounds} Define
  $\theta_a:=\Pr(Y(0)=1,D=1,A=a)$ for $a=0,1$.  Then, its sharp bounds
  are given by $\theta_{a} \in [\ubar{\theta}_a, \bar{\theta}_a]$
  where
  $$\begin{aligned}
  \ubar{\theta}_a \ = \  & \max_z \Pr(Y=1,D=0,A=a\mid Z=z)-
                  \Pr(Y=1,D=0,A=a), \\
   \bar{\theta}_a  \ = \ &  \Pr(A=a) -  \Pr(Y=1,D=0,A=a)  - \max_z \Pr(Y=0,D=0,A=a\mid Z=z).
\end{aligned}
$$
for $a=0,1$.  These sharp bounds can be achieved simultaneously for
$a=0,1$.

\end{lemma}

\begin{lemma} \label{cor:sharp_bounds-p} Define
  $\xi_{az} := \Pr\{Y(0) = 1, D(z) = 1, A=a\}$ for $a,z=0,1$.  Then,
  its sharp bounds are given by $\xi_{az}  \in [\ubar{\xi}_{az}, \bar{\xi}_{az}]$
  where
  $$
  \begin{aligned}
    \ubar{\xi}_{az} \ = \ & \max_{z^\prime}
      \Pr(Y=1,D=0,A=a\mid Z=z^\prime) - \Pr(Y=1,D=0,A=a\mid Z=z), \\
    \bar{\xi}_{az} \ = \ & \Pr(A=a) -  \Pr(Y=1,D=0,A=a\mid Z=z) - \max_{z^\prime} \Pr(Y=0,D=0,A=a\mid Z=z^\prime)
  \end{aligned}
  $$
  for $a,z=0,1$.  These sharp bounds can be achieved simultaneously for
  $a=0,1$ given $z=0,1$.

\end{lemma}

\section{Separate Evaluation of Each Decision-making System}
\label{sec:separate_eval}

While we have focused on identifying and bounding the {\it
  differences} in classification risks of the three decision-making
systems --- human-alone, AI-alone, and human-with-AI, it is also
possible to partially identify the classification risk of each
decision-making system separately.
\begin{theorem}[Sharp bounds on the classification risk of each
  decision-making system] \label{thm:bounds_each} \spacingset{1}
  The sharp bounds on the risk of each
  decision-making system are given by:
  \begin{enumerate}[label=(\alph*)]
\item Human-alone system:
  $$
  R_{\textsc{Human}}(\ell_{01}) \in [p_{10}(D(0)) + \ell_{01}  \cdot \ubar{p}_{01}(D(0)), \
  p_{10}(D(0)) + \ell_{01} \cdot \bar{p}_{01}(D(0))],
  $$
\item Human-with-AI system
  $$
  R_{\textsc{Human+AI}}(\ell_{01})\in [ p_{10}(D(1)) + \ell_{01}
  \cdot \ubar{p}_{01}(D(1)), \ p_{10}(D(1)) + \ell_{01}  \cdot
  \bar{p}_{01}(D(1))],
  $$
\item AI-alone system:
\begin{eqnarray*}
  R_{\textsc{AI}}(\ell_{01}) &\in  &\left[  \max_{z^\prime} \Pr(Y=1,D=0,A=0\mid Z=z^\prime) + \ell_{01} \cdot  \max_{z^\prime} \Pr(Y=0,D=0,A=1\mid Z=z^\prime) \right.,\\
    && \quad \Pr(A=0)- \max_{z^\prime} \Pr(Y=0,D=0,A=0\mid Z=z^\prime) \\
    && \quad \left. + \ell_{01} \cdot \left\{ \Pr(A=1)- \max_{z^\prime} \Pr(Y=1,D=0,A=1\mid Z=z^\prime)\right\}\right],
\end{eqnarray*}
\end{enumerate}
where
$\ubar{p}_{01}(D(z)):=\Pr(D = 1 \mid Z = z) - \bar{\xi}_{0z} -
\bar{\xi}_{1z}$ and
$\bar{p}_{01}(D(z)) :=\Pr(D = 1 \mid Z = z) - \ubar{\xi}_{0z} -
\ubar{\xi}_{1z}$ with the following upper and lower bounds of
$\xi_{az}:=\Pr\{Y(0)=1, D(z)=1, A=a\}$:
$$
\begin{aligned}
  \ubar{\xi}_{az} \ = \ & \max_{z^\prime}
  \Pr(Y=1,D=0,A=a\mid Z=z^\prime) - \Pr(Y=1,D=0,A=a\mid Z=z), \\
  \bar{\xi}_{az} \ = \ & \Pr(A=a) -  \Pr(Y=1,D=0,A=a\mid Z=z) - \max_{z^\prime} \Pr(Y=0,D=0,A=a\mid Z=z^\prime)
\end{aligned}
$$
for $a,z=0,1$. 
\end{theorem}
Similarly, we can derive the partial identification bounds for
non-linear classification measures, such as FNR, FPR, and FDR (see
Appendix \ref{app:bounds_alt_measures}).

\section{Sharp Bounds on Other Classification Ability Measures} \label{app:bounds_alt_measures}

For completeness, we present the sharp bounds for other classification
ability measures.
\begin{proposition} {\sc (Sharp bounds on false negative rate (FNR), false
    positive rate (FPR), and false discovery rate (FDR))}
  \spacingset{1}
\begin{enumerate}[label=(\alph*)] 
\item Human-alone system $(z=0)$ and Human-with-AI system ($z=1$):
  $$\begin{aligned}
     \frac{\Pr(Y = 1, D = 0 \mid Z = z)}{\Pr(Y = 1, D = 0 \mid Z = z) + \bar{p}_{11}(D(z))} \  \leq \text{FNR}(z) & \leq \frac{\Pr(Y = 1, D = 0 \mid Z = z)}{\Pr(Y = 1, D = 0 \mid Z = z) + \ubar{p}_{11}(D(z))} \\
     \frac{\ubar{p}_{01}(D(z))}{\Pr(Y = 0, D = 0 \mid Z = z) + \bar{p}_{01}(D(z))} \ \le \
     \text{FPR}(z) & \le \ \frac{\bar{p}_{01}(D(z))}{\Pr(Y = 0, D = 0 \mid Z = z) + \ubar{p}_{01}(D(z))}, \\
    \frac{\ubar{p}_{01}(D(z))}{\Pr(D = 1 \mid Z = z)} \ \le \ 
    \text{FDR}(z) & \le \ \frac{\bar{p}_{01}(D(z))}{\Pr(D = 1 \mid
      Z = z)}.
  \end{aligned}
  $$
\item AI-alone system:
  $$\begin{aligned}
    \frac{\ubar{p}_{10}(A)}{\max_{z \in \{0,1\}} \left\{p_{10}(D(z)) + \bar{p}_{11}(D(z)) \right\}} \ \le \
    \text{FNR}_{\textsc{AI}} & \le \  \frac{\bar{p}_{10}(A)}{\min_{z \in
        \{0,1\}} \left\{p_{10}(D(z)) + \ubar{p}_{11}(D(z)) \right\}},\\
    \frac{\ubar{p}_{01}(A) }{\max_{z \in \{0,1\}} \left\{p_{10}(D(z)) + \bar{p}_{11}(D(z)) \right\}}
    \ \le \ \text{FPR}_{\textsc{AI}} & \le \ \frac{\bar{p}_{01}(A)}{\min_{z \in \{0,1\}}\left\{p_{00}(D(z)) + \ubar{p}_{01}(D(z)) \right\}}, \\
    \frac{\ubar{p}_{01}(A)}{\Pr(A=1)} \
    \le \ \text{FDR}_{\textsc{AI}} &  \le \ \frac{\bar{p}_{01}(A)}{\Pr(A=1)}.
  \end{aligned}
  $$
\end{enumerate}
\end{proposition}

\section{Proofs}  \label{app:proofs}

\subsection{Proof of Theorem~\ref{thm:z_vs_z0}}

The law of total probability implies:
$$\Pr(Y(0) = y \mid Z = z) \ = \ p_{y1}(D(z)) + p_{y0}(D(z)), \text{ for } z \in \{0,1\}.$$
Then, since $\Pr(Y(0) = y \mid Z = 1) =\Pr(Y(0) = y \mid Z = 0)$ under
the random assignment of $Z$, we have:
\begin{equation*} 
 p_{y1}(D(1)) + p_{y0}(D(1)) =  p_{y1}(D(0)) + p_{y0}(D(0)) \implies
 p_{y1}(D(1)) - p_{y0}(D(0))=  p_{y0}(D(0)) - p_{y1}(D(1)),
\end{equation*}
for $y \in \{0,1\}$. Using this result, we obtain:
\begin{align*} 
R_{\textsc{human+AI}}(\ell_{01})-R_{\textsc{human}}(\ell_{01})
&= p_{10}(D(1)) + \ell_{01} p_{01}(D(1)) - \left\{p_{10}(D(0)) + \ell_{01} p_{01}(D(0))\right\}\\
&= p_{10}(D(1)) - p_{10}(D(0)) + \ell_{01}  \left\{p_{01}(D(1)) - p_{01}(D(0)) \right\}\\
&= p_{10}(D(1)) - p_{10}(D(0))  - \ell_{01} \left\{p_{00} (D(1)) - p_{00}(D(0))\right\}.
\end{align*} 
\qed

\subsection{Proof of Theorem~\ref{thm:z_vs_z0_est}}

Define $\beta_z$ and then rewrite it as
\begin{align*}
  \beta_z & = \E[\Pr(Y = 1, D = 0 \mid Z = z, X) - \ell_{01} \times \Pr(Y = 0, D = 0 \mid Z = z, X)]\\
           & = \E[(1 - m^D(z, X))m^Y(z, X) - \ell_{01} \times (1 - m^D(z, X))(1 - m^Y(z, X))]\\
           & = \E[(1 - m^D(z, X))\{(1 + \ell_{01})m^Y(z, X) - \ell_{01}\}]\\
           & = \E[(1-D(z))\{(1 + \ell_{01})Y(z) - \ell_{01}\}]\\
           & = \E[\E[W \mid X, Z = z]].
\end{align*}
Using this, we can write the classification risk difference as
$R_{\textsc{human+AI}}(\ell_{01})-R_{\textsc{human}}(\ell_{01}) =
\beta_1 - \beta_0$. 

Next, we define:
$$m(z, x ;\ell_{01}) := \E[W \mid X = x, Z = z] = \E[(1-D)\{(1 +
\ell_{01})Y - \ell_{01}\} \mid X = x, Z = z].$$ Then, the (uncentered)
efficient influence function is given by,
\[
  \varphi_z(Z, X, W; \ell_{01}) =  m(z, x; \ell_{01}) + \frac{\bbone\{Z = z\}}{e(z,X)}(W - m(z, X_i; \ell_{01})).
\]
Now we write the compound outcome model as 
\[
  m(z, x; \ell_{01}) = (1 - m^D(z, x))\{(1 + \ell_{01}) m^Y(z, x) - \ell_{01}\}.
\]
Plugging this expression along with
$W = (1-D)\{(1 + \ell_{01})Y - \ell_{01}\}$ into the efficient
influence function yields
\begin{align*}
  \varphi_z(Z, X, D, Y; \ell_{01}) & = (1 - m^D(z, X))\{(1 + \ell_{01}) m^Y(z, X) - \ell_{01}\}\\
                                   & \qquad + (1 + \ell_{01})\frac{\bbone\{Z = z\}(1 - D)}{e(z, X)}(Y - m^Y(z, X))\\
                                   & \qquad  - \{(1 + \ell_{01})m^Y(z, X) - \ell_{01}\}\frac{\bbone\{Z  = z\}}{e(z, X)}(D - m^D(z, X)).
\end{align*}

Following the rubric for one-step estimators outlined in
\citet{kennedy_semiparametric_2023}, it remains to show that the
remainder bias is controlled under the rate conditions in
Assumption~\ref{assum:rates}.  Adding and subtracting terms, the bias
of the one-step estimator
$\hat\beta_z: = \mathbb{P}_n\{\hat\varphi_z(Z, X, D, Y; \ell_{01})\}$
is
\begin{align*}
   & \E[\hat{\beta}_z - \beta_z] \\
  = &  \E\left[ (1 - \hat{m}^D(z, X))\{(1 + \ell_{01}) \hat{m}^Y(z, X) - \ell_{01}\} - (1 - m^D(z,X))\{(1 + \ell_{01})m^Y(z,X) - \ell_{01})\}\right.\\
    & \qquad + (1 + \ell_{01})\left(\frac{e(z,X)}{\hat{e}(z,X)} - 1\right) (1 - m^D(z, X)) (m^Y(z, X) - \hat{m}^Y(z,X))\\
    & \qquad - \{(1 + \ell_{01})\hat{m}^Y(z, X) - \ell_{01}\}\left(\frac{e(z,X)}{\hat{e}(z,X)} - 1\right)(m^D(z,X) - \hat{m}^D(z, X))\\
    & \qquad + (1 + \ell_{01})(1-m^D(z,X))(m^Y(z, X) - \hat{m}^Y(z, X))\\
    & \qquad \left.- \{(1 + \ell_{01})\hat{m}^Y(z,X) - \ell_{01}\}(m^D(z,X) - \hat{m}^D(z,X))\right]\\
  =  &  \E\left[ (1 + \ell_{01})\left(\frac{e(z,X) - \hat{e}(z,X)}{\hat{e}(z,X)} \right) (1 - m^D(z, X)) (m^Y(z, X) - \hat{m}^Y(z,X))\right]\\
    & \qquad - \E\left[\{(1 + \ell_{01})\hat{m}^Y(z, X) - \ell_{01} \}\left(\frac{e(z,X) - \hat{e}(z,X)}{\hat{e}(z,X)}\right)(m^D(z,X) - \hat{m}^D(z, X))\right]\\
   =  &  \E\left[\frac{e(z,X) - \hat{e}(z,X)}{\hat{e}(z,X)} \left[\{(1 + \ell_{01}) m^Y(z, X) -
      \ell_{01}\}(1 - m^D(z, X)) \right.\right.\\
  & \qquad \qquad \qquad \qquad \qquad \left. - \{(1 + \ell_{01}) \hat{m}^Y(z, X) -
      \ell_{01}\}(1 - \hat{m}^D(z, X))\right]\bigg] \\
    =  &  \E\left[\frac{e(z,X) - \hat{e}(z,X)}{\hat{e}(z,X)} \left[(1 + \ell_{01})(m^Y(z, X) -
         \hat{m}^Y(z,X)) + \ell_{01}(m^D(z, X) - \hat{m}^D(z,X)) \right.\right. \\
  & \qquad \qquad \qquad \qquad \qquad \left. - (1 +
      \ell_{01}) (m^Y(z,X)m^D(z,X) - \hat{m}^Y(z, X) \hat{m}^D(z, X)) \right]\bigg]
\end{align*}
Therefore, the absolute bias is bounded by
\[
  |\E[\hat{\beta}_z - \beta_z]| \leq C(\|m^Y(z, \cdot) - \hat{m}^Y(z, \cdot)\|_2 + \|m^D(z, \cdot) - \hat{m}^D(z,\cdot)\|_2) \times \|\hat{e}(z,\cdot) - e(z,\cdot)\|_2,
\]
By Assumption~\ref{assum:rates} and \citet{kennedy_semiparametric_2023} (Proposition 2), we can then write 
\begin{align*}
  \hat{\beta}_1 - \hat{\beta}_0 - (\beta_1 - \beta_0) & = \frac{1}{n}\sum_{i=1}^n \varphi_1(Z_i, X_i, D_i, Y_i) - \varphi_0(Z_i, X_i, D_i, Y_i) -(\beta_1 - \beta_0) + o_p\left(n^{-1/2}\right).  
\end{align*}
This leads to the desired result,
\[
  \sqrt{n}\left(\hat{\beta}_1 - \hat{\beta}_0 - (\beta_1 - \beta_0)\right) \stackrel{d}{\longrightarrow} N(0, V),
\]
where $V = \E[(\varphi_1(Z, X, D, Y) - \varphi_0(Z, X, D, Y) -(\beta_1 - \beta_0))^2]$.

\qed

\subsection{Proof of Lemma~\ref{cor:sharp_bounds}}

We first show that the joint distribution of $(Y(0), D,Z,A)$ in terms of $\theta_a$ and the observed data distribution.  Since  $(D,Z,A)$ are observed, it suffices to show that $\Pr(Y(0)=1\mid D=1,Z,A)$ can be expressed in terms of $\theta_a$ and the observed data distribution.
We can write
\begin{eqnarray*}
\Pr(Y(0)=1,A=a\mid Z=z)&=&\Pr(Y(0)=1,A=a)\\
&=&\Pr(Y(0)=1,D=1,A=a)+\Pr(Y(0)=1,D=0,A=a)\\
&=&\theta_{a} + \Pr(Y=1,D=0,A=a).
\end{eqnarray*}
Therefore,
\begin{align}
\nonumber &\Pr(Y(0)=1,D=1,A=a\mid Z=z)\\
\nonumber = \ &\Pr(Y(0)=1,A=a\mid Z=z) -\Pr(Y(0)=1,D=0,A=a\mid Z=z)\\
\label{eqn::bounds-p}=\ &\theta_a+ \Pr(Y=1,D=0,A=a) - \Pr(Y=1,D=0,A=a\mid Z=z),
\end{align}
which yields
\begin{eqnarray*}
&&\Pr(Y(0)=1\mid D=1,A=a, Z=z)\\
&=&\frac{\theta_{a}+ \Pr(Y=1,D=0,A=a) - \Pr(Y=1,D=0,A=a\mid Z=z)}{\Pr(D=1,A=a\mid Z=z)}.
\end{eqnarray*}

Next, we derive the sharp bounds on $\theta_a$.  We only need to solve
the inequalities that $\Pr(Y(0)=1\mid D=1,Z,A)$ lie within $[0,1]$
since there is no additional restriction on this conditional
probability:
\begin{eqnarray*}
0\leq \frac{\theta_{a}+ \Pr(Y=1,D=0,A=a) - \Pr(Y=1,D=0,A=a\mid Z=z)}{\Pr(D=1,A=a\mid Z=z)}  \leq 1.
\end{eqnarray*}
This yields the following sharp lower and upper bounds:
\begin{eqnarray*}
\theta_{a} &\geq& \max_z \Pr(Y=1,D=0,A=a\mid Z=z)- \Pr(Y=1,D=0,A=a) ,\\
\theta_{a} &\leq&  \min_z \left\{ \Pr(Y=1,D=0,A=a\mid Z=z)+ \Pr(D=1,A=a\mid Z=z)\right\} -  \Pr(Y=1,D=0,A=a)\\
&=&  \Pr(A=a) -  \Pr(Y=1,D=0,A=a)  - \max_z \Pr(Y=0,D=0,A=a\mid Z=z).
\end{eqnarray*}
The above derivation implies that for any observed data distribution
of $(Y,D,Z,A)$, there exists a complete data distribution of
$(Y(d),D(z),Z,A)$ such that $\theta_0$ and $\theta_1$ equal their
upper or lower bounds simultaneously.
\qed

\subsection{Proof of Lemma~\ref{cor:sharp_bounds-p}}

By combining Lemma~\ref{cor:sharp_bounds} with
Equation~\eqref{eqn::bounds-p}, the desired sharp bounds on
$\Pr(Y(0)=1,D(z)=1,A=a)=\Pr(Y(0)=1,D=1,A=a\mid Z = z)$ follow
immediately.  These bounds can also be attained simultaneously for
$z=0,1$, because the bounds on $\theta_0$ and $\theta_1$ can be
attained simultaneously (Lemma~\ref{cor:sharp_bounds}). \qed

\subsection{Proof of Theorem~\ref{thm:partial_id_risk}}
To simplify the notation, we will focus on bounding the quantities without
conditioning on the covariates $X$ (and assuming that provision $Z$ is
independent of $(A, \{D_(z), Y_(d)\}_{z,d \in \{0,1\}})$). The proof
conditional on $X$ is analogous.
We express the risks under the three decision-making systems in terms
of the observed data distribution and $\Pr(Y(0)=1,D(z)=1,A=a)$.
For the AI-alone system, we have 
\begin{eqnarray}
\nonumber R_{\textsc{AI}}(\ell_{01}) 
\nonumber&=& \Pr(Y(0) = 1, A = 0) + \ell_{01} \times \Pr(Y(0) = 0, A = 1) \\
\nonumber&=&\Pr(Y= 1, D = 0,A=0 \mid Z = z) +\Pr(Y(0) = 1,  D(z) = 1, A=0) +  \\
\label{eqn::ai-p} &&+ \ell_{01} \times \left\{\Pr(Y= 0, D = 0,A=1 \mid Z= z) + \Pr(Y(0) = 0, D(z) = 1,A=1) \right\}.
\end{eqnarray}
The second equality holds for both $z=0,1$ due to independence between
$(Y(0),A)$ and $Z$.  Similarly, for the human-alone system, we have
\begin{equation}
\begin{aligned}
 R_{\textsc{Human}}(\ell_{01}) 
 \ = \ &\Pr(Y= 1, D = 0 \mid Z = 0)\\
&+ \ell_{01} \times \left\{  \Pr(Y(0) = 0, D(0)
  = 1,A=1)  + \Pr(Y(0) = 0, D(0) = 1,A=0)
\right\}. 
\end{aligned} \label{eqn::human-p}
\end{equation}
Finally, for the human-with-AI system, we have 
\begin{equation}
  \begin{aligned}
R_{\textsc{Human+AI}}(\ell_{01}) 
\ = \ &\Pr(Y= 1, D = 0 \mid Z = 1) \\ 
&+ \ell_{01} \times \left\{  \Pr(Y(0) = 0, D(1) = 1,A=1)  + \Pr(Y(0) =
  0, D(1) = 1,A=0) \right\}.
\end{aligned}\label{eqn::human+ai-p} 
\end{equation}
From Equation~\eqref{eqn::ai-p} with $z=0$ and Equation~\eqref{eqn::human-p}, we have 
\begin{eqnarray*}
\nonumber&&R_{\textsc{AI}}(\ell_{01}) -R_{\textsc{Human}}(\ell_{01}) \\
\nonumber&=&\Pr(Y(0) = 1,  D(0) = 1, A=0) +\Pr(Y= 1,  D = 0, A=0 \mid Z = 0) - \Pr(Y= 1, D = 0 \mid Z = 0)\\
\nonumber&& +\ell_{01} \times  \{\Pr(Y= 0, D = 0, A = 1 \mid Z= 0)-  \Pr(Y(0) = 0, D(0) = 1,A=0) \}\\
\nonumber&=&\Pr(Y(0) = 1,  D(0) = 1, A=0) -\Pr(Y= 1,  D = 0, A=1 \mid Z = 0)\\
\nonumber&&+ \ell_{01} \times   \left\{  \Pr(Y= 0, D = 0, A = 1 \mid
            Z= 0) - \Pr(D(0)=1,A=0) + \Pr(Y(0) = 1, D(0) = 1,A=0)\right\}\\
\nonumber&=&(1+\ell_{01})\times \Pr(Y(0) = 1,  D(0) = 1, A=0) -\Pr(Y= 1,  D = 0, A=1 \mid Z = 0)\\
\nonumber&&+\ell_{01} \times \left\{  \Pr(Y= 0, D = 0, A = 1 \mid Z= 0)- \Pr(D=1,A=0\mid Z=0) \right\}\\
\nonumber&=&(1+\ell_{01})\times \left\{ \Pr(Y(0) = 1,  D(0) = 1, A=0)-\Pr(Y= 1,  D = 0, A=1 \mid Z = 0) \right\} \\
\nonumber&&+\ell_{01} \times \left\{ \Pr(D = 0, A = 1 \mid Z= 0)- \Pr(D=1,A=0\mid Z=0) \right\}.
\end{eqnarray*}
Using Lemma~\ref{cor:sharp_bounds-p}, we obtain 
\begin{eqnarray*}
&& R_{\textsc{AI}}(\ell_{01}) -R_{\textsc{Human}}(\ell_{01}) \\
&\geq & (1+\ell_{01})\times \left\{  \max_z \Pr(Y=1,D=0,A=0\mid Z=z) - \Pr(Y=1,D=0\mid Z=0) \right\} \\
\nonumber&&+\ell_{01} \times \left\{ \Pr(D = 0, A = 1 \mid Z= 0)- \Pr(D=1,A=0\mid Z=0) \right\}
\end{eqnarray*}
and 
\begin{eqnarray*}
&& R_{\textsc{AI}}(\ell_{01}) -R_{\textsc{Human}}(\ell_{01}) \\
&\leq &(1+\ell_{01})\times \left\{ \Pr(A=0) -  \Pr(Y=1,D=0\mid Z=0) - \max_z \Pr(Y=0,D=0,A=0\mid Z=z)\right\} \\
\nonumber&&+\ell_{01} \times \left\{ \Pr(D = 0, A = 1 \mid Z= 0)- \Pr(D=1,A=0\mid Z=0) \right\}.
\end{eqnarray*}

Similarly, from Equation~\eqref{eqn::ai-p} with $z=1$ and Equation~\eqref{eqn::human+ai-p}, we have 
\begin{eqnarray*}
\nonumber&&R_{\textsc{AI}}(\ell_{01}) -R_{\textsc{Human+AI}}(\ell_{01}) \\
\nonumber&=&(1+\ell_{01})\times \left\{ \Pr(Y(0) = 1,  D(1) = 1, A=0)-\Pr(Y= 1,  D = 0, A=1 \mid Z = 1) \right\} \\
\nonumber&&+\ell_{01} \times \left\{ \Pr(D = 0, A = 1 \mid Z= 1)- \Pr(D=1,A=0\mid Z=1) \right\}.
\end{eqnarray*}
Again, using Lemma~\ref{cor:sharp_bounds-p}, we have the desired result:
\begin{eqnarray*}
&& R_{\textsc{AI}}(\ell_{01}) -R_{\textsc{Human+AI}}(\ell_{01}) \\
&\geq & (1+\ell_{01})\times \left\{  \max_z \Pr(Y=1,D=0,A=0\mid Z=z) - \Pr(Y=1,D=0\mid Z=1) \right\} \\
\nonumber&&+\ell_{01} \times \left\{ \Pr(D = 0, A = 1 \mid Z= 1)- \Pr(D=1,A=0\mid Z=1) \right\}
\end{eqnarray*}
and 
\begin{eqnarray*}
&& R_{\textsc{AI}}(\ell_{01}) -R_{\textsc{Human+AI}}(\ell_{01}) \\
&\leq &(1+\ell_{01})\times \left\{ \Pr(A=0) -  \Pr(Y=1,D=0\mid Z=1) - \max_z \Pr(Y=0,D=0,A=0\mid Z=z)\right\} \\
\nonumber&&+\ell_{01} \times \left\{ \Pr(D = 0, A = 1 \mid Z= 1)- \Pr(D=1,A=0\mid Z=1) \right\}.
\end{eqnarray*}
\qed

\subsection{Proof of Theorem~\ref{thm:partial_id_risk_est}}

Beginning with the lower bound, we write
\begin{align*}
  & \E[L_z(X)] \\
  = \ &  (1 + \ell_{01}) \E[\Pr(Y = 1, D = 0, A = 0 \mid Z = z, X) - \Pr(Y = 1, D = 0 \mid Z = z, X)]\\
  & \quad + \ell_{01} \E\left[\Pr(D = 0, A = 1 \mid Z = z, X) - \Pr(D = 1, A = 0 \mid Z = z, X)\right]\\
  & \quad + (1 + \ell_{01})\E[g_L(X)(\Pr(Y = 1, D = 0, A = 0 \mid Z = 1-z, X) - \Pr(Y = 1, D = 0, A = 0 \mid Z = z, X))]\\
  = \ & (1 + \ell_{01}) \vartheta^L_{1z} + \ell_{01} \vartheta^L_{2z} + (1 + \ell_{01})\vartheta^L_{3z},
\end{align*}
where
\begin{align*}
  \vartheta^L_{1z} & = \E\left[(1 - m^A(X)) \left\{(1 - m^D(z, X, 0)) m^Y(z, X, 0) - (1 - m^D(z, X))m^Y(z, X)\right\}\right],\\
  \vartheta^L_{2z} & = \E\left[m^A(X) \left\{1 - m^D(z, X, 1) - (1 - m^A(X))m^D(z, X, 0)\right\}\right],\\
  \vartheta^L_{3z} &= \E\left[g_{L_z}(X)(1 - m^A(X)) \left\{(1 - m^D(1-z, X, 0))m^Y(1-z, X, 0) - (1 - m^D(z, X, 0))m^Y(z, X, 0)\right\}\right].
\end{align*}

We show how to estimate each in turn. First, we estimate $\vartheta^L_{1z}$ as
\[
  \hat{\vartheta}^L_{1z} = \frac{1}{n}\sum_{i=1}^n \left(\widehat{\varphi}_{z1}(Z_i, X_i, D_i, A_i, Y_i) - \widehat{\varphi}_z(Z_i, X_i, D_i, Y_i; 0)\right)
\]
In the proof of Theorem~\ref{thm:z_vs_z0_est}, we have controlled the
second term, so it suffices to consider the first term. Note that it
is equivalent to the AIPW estimate with the compound outcome
$\widetilde{W} = (1-D)Y$ restricted to where $A = 0$, because
$\E[\widetilde{W} \mid X = x, Z = z] = (1 - m^D(z, x, 0))m^Y(z, x,
0)$, $A \indep Z \mid X$, and the (uncentered) efficient influence
function is
\begin{align*}
  \varphi_{z1}(Z, X, \widetilde{W})
  & = (1 - A)(1 - m^D(z, X, 0))m^Y(z, X, 0)\\
  & \qquad  + \frac{\bbone\{Z = z\}(1-A)}{e(z, X)}\left\{(1-D)Y - (1 - m^D(z, X, 0))m^Y(z, X, 0)\right\}\\
  & = (1-A)(1 - m^D(z, X, 0))m^Y(z, X, 0) + \frac{\bbone\{Z = z\}(1-A)(1-D)}{e(z, X)}(Y - m^Y(z, X, 0))\\
  & \qquad  - \frac{\bbone\{Z = z\}(1-A)m^Y(z, X, 0)}{e(z, X)}(D - m^D(z, X, 0)).
\end{align*}
Now, we control the remainder bias term,
\begin{align*}
  & \E[\widehat{\varphi}_{z1}(Z, X, D, A, Y) - (1 - m^A(X))(1 - m^D(z, X,
  0))m^Y(z, X, 0)]\\
   = \ & \E\left[(1 - m^A(X))\left\{(1 - \hat{m}^D(z, X, 0))\hat{m}^Y(z, X, 0) - (1 - m^D(z, X, 0))m^Y(z, X, 0)\right\}\right]\\
    & +  \E\left[(1 - m^A(X)) \frac{e(z,X) (1 - m^D(z, X, 0))}{\hat{e}(z,X)}(m^Y(z, X, 0) - \hat{m}^Y(z, X, 0))\right]\\
    & -  \E\left[(1 - m^A(X))m^Y(z, X, 0)\frac{e_z(X)}{\hat{e}(z,X)}(m^D(z,X, 0) - \hat{m}^D(z, X, 0))\right]
\end{align*}
where $m^A(x):=\Pr(A = 1 \mid D = 0, X = x)$.
Following the proof of Theorem~\ref{thm:z_vs_z0_est}, this is equal to
\begin{align*}
  & \E\left[ (1 - m^A(X))\left(\frac{e(z,X) - \hat{e}(z,X)}{\hat{e}(z,X)} \right) (1 - m^D(z, X, 0)) (m^Y(z, X) - \hat{m}^Y(z,X))\right]\\
  & \qquad + \E\left[(1 - m^A(X))(1 - \hat{m}^Y(z, X, 0))\left(\frac{e(z,X) - \hat{e}(z,X)}{\hat{e}(z,X)}\right)(m^D(z,X, 0) - \hat{m}^D(z, X, 0)).\right]\\
  \leq & C(\|m^Y(z, \cdot, 0) - \hat{m}^Y(z, \cdot, 0)\|_2 + \|m^D(z, \cdot, 0) - \hat{m}^D(z,\cdot, 0)\|_2) \times \|\hat{e}(z,\cdot) - e(z,\cdot)\|_2
\end{align*}

Next, we estimate $\vartheta^L_{2z}$ with
\[
  \hat{\vartheta}^L_{2z} = \frac{1}{n}\sum_{i=1}^n\widehat{\varphi}^D_{z1}(Z_i, X_i, D_i, A_i, Y_i) - \widehat{\varphi}^D_{z0}(Z_i, X_i, D_i, A_i, Y_i).
\]
This is the standard AIPW estimator for the mean of $(1 - D(z))$ restricted to $A = 1$ ($\widehat{\varphi}^D_{z1}$) minus the mean of $D(z)$ restricted to $A = 0$ ($\widehat{\varphi}^D_{z0}$). From the standard product term decomposition for the doubly robust estimator of a mean with missing outcomes, we can see that the bias is
\begin{align*}
  \E[\hat{\vartheta}^L_{2z} - \vartheta^L_{2z}] & = \E\left[(1 - m^A(X)) \left(\frac{e(z,X) - \hat{e}(z,X)}{\hat{e}(z,X)}\right)(m^D(z,X, 0) - \hat{m}^D(z, X, 0))\right]\\
                                                & \qquad + \E\left[m^A(X) \left(\frac{e(z,X) - \hat{e}(z,X)}{\hat{e}(z,X)}\right)(m^D(z,X, 1) - \hat{m}^D(z, X, 1))\right]\\
                                                & \leq C \left(\|m^D(z, \cdot, 0) - \hat{m}^D(z,\cdot, 0)\|_2 + \|m^D(z, \cdot, 1) - \hat{m}^D(z,\cdot, 1)\|_2\right) \times \|\hat{e}(z,\cdot) - e(z,\cdot)\|_2
\end{align*}

Finally, we estimate $\vartheta^L_{3z}$ with 
\[
  \hat{\vartheta}^L_{3z} = \frac{1}{n}\sum_{i=1}^n \hat{g}_{L_z}(X_i)\left(\widehat{\varphi}_{1-z,1}(Z_i, X_i, D_i, A_i, Y_i) - \widehat{\varphi}_{z1}(Z_i, X_i, D_i, A_i, Y_i) \right).
\]
To make the results more compact, let 
\begin{align*}
  f(x) & = (1 - m^A(x)) \left\{(1 - m^D(1-z, x, 0))m^Y(1-z, x, 0) - (1 - m^D(z, x, 0))m^Y(z, x, 0)\right\},  \\
  \hat{f}(x) & = (1 - m^A(x)) \left\{(1 - \hat{m}^D(1-z, x, 0))\hat{m}^Y(1-z, x, 0) - (1 - \hat{m}^D(z, x, 0))\hat{m}^Y(z, x, 0)\right\}.
\end{align*}
To compute the bias, notice that
\begin{align*}
  \E[\hat{\vartheta}^L_{3z} - \vartheta_{3z}]
  & = \E\left[\hat{g}_{L_z}(X)\left(\widehat{\varphi}_{1-z,1}(Z, X, D, A, Y) - \widehat{\varphi}_{z1}(Z, X, D, A, Y)\right)\right] - \E[g_{L_z}(X) f(X)]\\
  & = \E\left[\left(\hat{ g}_{Lz}(X) -  g_{L_z}(X)\right) f(X)\right] + \E\left[\hat{g}_{L_z}(X)\left(\widehat{\varphi}_{1-z,1}(Z, X, D, A, Y) - \widehat{\varphi}_{z1}(Z, X, D, A, Y)- f(X) \right)\right]\\
  & \leq \E\left[\left(\hat{ g}_{Lz}(X) -  g_{L_z}(X)\right) f(X)\right]\\
  & \qquad + C\left(\sum_{z' = 0}^1\|m^Y(z', \cdot, 0) - \hat{m}^Y(z', \cdot, 0)\|_2 + \|m^D(z', \cdot, 0) - \hat{m}^D(z',\cdot, 0)\|_2\right) \times \|\hat{e}(z,\cdot) - e(z,\cdot)\|_2
\end{align*}
where the final inequality follows from the same arguments about $\hat{\vartheta}^L_{1z}$ above.
Next, notice that $g_{L_z}(x) = \bbone\{f(x) \geq 0\}$, and if $\hat{g}_{L_z}(x) \neq g_{L_z}(x)$, then 
\begin{align*}
    |f(x)| & \leq |f(x) - \hat{f}(x)|\\
           & \leq |1 - m^D(1-z, x, 0)m^Y(1-z, x, 0) - (1 - \hat{m}^D(1-z, x, 0))\hat{m}^Y(1-z, x, 0)|\\
           & \qquad + |1 - m^D(z, x, 0)m^Y(z, x, 0) - (1 - \hat{m}^D(z, x, 0)\hat{m}^Y(z, x, 0)|\\
           & \leq 2\max_{z'} |\hat{m}^D(z', x, 0) - m^D(z', x, 0)| + 2\max_{z'} |\hat{m}^Y(z', x, 0) - m^Y(z', x, 0)|.
\end{align*}
Following the argument in \citet{Audibert2007}, this implies that
\begin{align*}
  & \E\left[\left(\hat{ g}_{Lz}(X) -  g_{L_z}(X)\right) f(X)\right] \\
  \leq \ & \E\left[\bbone\left\{\hat{ g}_{Lz}(X) \neq  g_{L_z}(X)\right\} |f(X)|\right]\\
  \leq \ & \E\left[\bbone\left\{|f(X)| \leq |f(X) - \hat{f}(X)|\right\} |f(X)|\right]\\
  \leq \ & \E\left[\bbone\left\{|f(X)| \leq |f(X) - \hat{f}(X)|\right\} |f(X) - \hat{f}(X)|\right]\\
  \leq \ & 2(\|\hat{m}^D(\cdot, \cdot, 0) - m^D(\cdot, \cdot, 0)\|_\infty + \|\hat{m}^Y(\cdot, \cdot, 0) - m^Y(\cdot, \cdot, 0)\|_\infty )\Pr(|f(X)| \leq |f(X) - \hat{f}(X)|)\\
  \leq \ & 2C (\|\hat{m}^D(\cdot, \cdot, 0) - m^D(\cdot, \cdot, 0)\|_\infty + \|\hat{m}^Y(\cdot, \cdot, 0) - m^Y(\cdot, \cdot, 0)\|_\infty )^{1 + \alpha}
\end{align*}
  
Putting together the pieces, under Assumptions~\ref{assum:rates},
\ref{assum:margin}, and \ref{assum:margin_error_rates}, we have
\begin{align*}
  \widehat{L}_z  & = \frac{1}{n}\sum_{i=1}^n (1 + \ell_{01})(\varphi_{z1}(Z_i, X_i, D_i, A_i, Y_i) - \varphi_z(Z_i, X_i, D_i, Y_i; 0) )\\
                & \qquad \qquad +  \ell_{01}(\varphi^D_{z1} (Z_i, X_i, D_i, A_i, Y_i) - \varphi^D_{z0} (Z_i, X_i, D_i, A_i, Y_i)) \\
                & \qquad \qquad \qquad + (1 + \ell_{01})
                  g_{L_z}(X_i)\left(\varphi_{1-z,1}(Z_i,
                  X_i, D_i, A_i, Y_i) - \varphi_{z1}(Z_i, X_i,
                  D_i, A_i, Y_i) \right) + o_p(n^{-1/2}),
\end{align*}
Thus, we have the desired result,
\[
  \sqrt{n}(\widehat{L}_z - L_z) \stackrel{d}{\longrightarrow} N(0, V_{L_z}),
\]
where 
\begin{align*}
  V_{L_z} = & \E\left[\{ (1 + \ell_{01}) (\varphi_z(Z, X, D, A, Y) - \varphi_z(Z, X, D, Y; 0)) +  \ell_{01}(\varphi^D_{z1} (Z, X, D, A, Y) - \varphi^D_{z0} (Z, X, D, A, Y))\right.\\
            &\qquad \left. \left. + (1 + \ell_{01})g_{L_z}(X)\left(\varphi_{1-z,1}(Z, X, D, A, Y) - \varphi_{z}(Z, X, D, A, Y) \right) - L_z\right\}^2\right].
\end{align*}

Turning to the upper bound, notice that
\begin{align*}
  & \E[U_z(X)] \\
  = &  (1 + \ell_{01}) \E\left[(1 - m^A(X)) - m^D(z, X)m^Y(z,X) - (1 - m^A(X))(1 - m^D(z,X,0))(1 - m^Y(z, X, 0))\right]\\
  & \qquad + \ell_{01} \E\left[m^A(X)(1 - m^D(z, X, 1)) - (1 - m^A(X))m^D(z, X, 0)\right]\\
  & \qquad + (1 + \ell_{01})\E[g_{U_z}(X)(1 - m^A(X))\\
  & \qquad \qquad \qquad \times \{(1 - m^D(1-z, X, 0))(1 - m^Y(1-z, X, 0)) - (1 - m^D(z, X, 0))(1 - m^Y(z, X, 0))\}]\\
  =  &  (1 - \ell_{01})\E\left[(1 - m^A(X))(1 - m^D(z, X, 0)m^Y(z, X, 0) - (1 - m^D(z, X))m^Y(z, X)\right]\\
  & \qquad +  \ell_{01}\E[m^A(X)(1 - m^D(z, X, 1))] + \E[(1 - m^A(X)m^D(z, X, 0))]\\
  & \qquad - (1 + \ell_{01})\E[g_{U_z}(X)(1 - m^A(X))\\
  & \qquad \qquad \qquad \times \{(1 - m^D(1-z, X, 0))(1 - m^Y(1-z, X, 0)) - (1 - m^D(z, X, 0))(1 - m^Y(z, X, 0))\}]\\
  = &  (1 + \ell_{01})\vartheta^U_{1z} +\vartheta^U_{2z} - (1 + \ell_{01})\vartheta^U_{3z},
\end{align*}
where
\begin{align*}
  \vartheta^U_{1z} & = \E[(1 - m^A(X))(1 - m^D(z, X, 0)m^Y(z, X, 0) - (1 - m^D(z, X))m^Y(z, X)]\\
  \vartheta^U_{2z} & =  \ell_{01}\E[m^A(X)(1 - m^D(z, X, 1))] + \E[(1 - m^A(X)m^D(z, X, 0))]\\
  \vartheta^U_{3z} &= \E[g_{U_z}(X)(1 - m^A(X))((1 - m^D(1-z, X, 0))(1 - m^Y(1-z, X, 0)) - (1 - m^D(z, X, 0)(1 - m^Y(z, X, 0))))]
\end{align*}

Notice that $\vartheta^U_{1z}  = \vartheta^L_{1z}$ above, which we have already analyzed.
  Next, we estimate $\vartheta^U_{2z}$ with
  \[
  \hat{\vartheta}^U_{2z} = \frac{1}{n}\sum_{i=1}^n \ell_{01}\widehat{\varphi}^D_{z1}(Z_i, X_i, D_i, A_i, Y_i) + \widehat{\varphi}^D_{z0}(Z_i, X_i, D_i, A_i, Y_i).
  \]
  Following the decomposition for $\hat{\vartheta}_{2z}^L - \vartheta_{2z}^L$ above, we can see that
  \[\E[\hat{\vartheta}^U_{2z} - \vartheta^U_{2z}] \leq C \left(\|m^D(z, \cdot, 0) - \hat{m}^D(z,\cdot, 0)\|_2 + \|m^D(z, \cdot, 1) - \hat{m}^D(z,\cdot, 1)\|_2\right) \times \|\hat{e}(z,\cdot) - e(z,\cdot)\|_2,\]
  for some $C > 0$.
  Finally, we estimate $\vartheta^U_{3z}$ with 
  \[
  \hat{\vartheta}^U_{3z} = \frac{1}{n}\sum_{i=1}^n \hat{g}_{U_z}(X_i)\left(\widehat{\varphi}_{1-z,0}(Z_i, X_i, D_i, A_i, Y_i) - \widehat{\varphi}_{z0}(Z_i, X_i, D_i, A_i, Y_i) \right).
  \]
  The analysis of $\E[\hat{\vartheta}^U_{3z}  - \vartheta^U_{3z}]$
  follows that of $\E[\hat{\vartheta}^L_{3z}  - \vartheta^L_{3z}]$,
  with $1 - m^Y(z, X, 0)$ and  $1 - \hat{m}^Y(z, X, 0)$ replacing
  $m^Y(z, X, 0)$  and  $\hat{m}^Y(z, X, 0)$ throughout. Putting
  together the pieces as with $\widehat{L}_z$ above gives the desired result.

  \qed

\subsection{Proof of Theorem~\ref{thm:policy_learn_rec}}

Denote the objective in Equation~\eqref{eq:optimal_recommendation_rule} as $\hat{R}_\textsc{rec}(\pi; \ell_{01})$. Notice that 
\begin{align*}
  & R_\textsc{rec}(\hat{\pi}_\textsc{rec}; \ell_{01}) - R_\textsc{rec}(\pi^\ast_\textsc{rec}; \ell_{01}) \\
  = \ &  R_\textsc{rec}(\hat{\pi}_\textsc{rec}; \ell_{01})  -
        \hat{R}_\textsc{rec}(\hat{\pi}_\textsc{rec}; \ell_{01})  +
        \underbrace{\hat{R}_\textsc{rec}(\hat{\pi}_\textsc{rec};
        \ell_{01}) - \hat{R}_\textsc{rec}(\pi^\ast_\textsc{rec};
        \ell_{01})}_{\leq 0} \\
  & \qquad + \hat{R}_\textsc{rec}(\pi^\ast_\textsc{rec}; \ell_{01}) -  R_\textsc{rec}(\pi^\ast_\textsc{rec}; \ell_{01})\\
  \leq \ &  2 \sup_{\pi \in \Pi}| \hat{R}_\textsc{rec}(\pi; \ell_{01}) -  R_\textsc{rec}(\pi; \ell_{01})|\\
  \leq \ & 2 \sup_{\pi \in \Pi} |  \hat{R}_\textsc{rec}(\pi; \ell_{01}) -  \E[\hat{R}_\textsc{rec}(\pi; \ell_{01})] | + 2 \sup_{\pi \in \Pi} | \E[\hat{R}_\textsc{rec}(\pi; \ell_{01})]  - R_\textsc{rec}(\pi; \ell_{01})|,
\end{align*}
where the first inequality uses the fact that $\hat{\pi}$ minimizes
$\hat{R}_\textsc{rec}(\pi; \ell_{01})$.  Now,
$\hat{R}_\textsc{rec}(\pi; \ell_{01}) - \E[\hat{R}_\textsc{rec}(\pi;
\ell_{01})]$ is a mean-zero empirical process.  In addition, note that
since $A, D, Y$ are all binary, the elements of $\hat{R}_\textsc{rec}$
are bounded by $\left(1 + \frac{4}{\eta}\right)(1 + \ell_{01})$.
Therefore, by \citet{wainwright_2019}, Theorem 4.2,
\[
  2 \sup_{\pi \in \Pi} | \hat{V}(\pi) - \E[\hat{V}(\pi)] \leq \left(1 + \frac{4}{\eta}\right)(1 + \ell_{01}) \mathcal{R}_n(\Pi) + \frac{t}{\sqrt{n}},
\]
with probability at least $1 - \exp\left(-\frac{t^2}{2}\right)$.

It remains to control
$\sup_{\pi \in \Pi} | \E[\hat{R}_\textsc{rec}(\pi; \ell_{01})] -
R_\textsc{rec}(\pi; \ell_{01})|$. Recall that we have bounded each of
the components of
$\E[\hat{R}_\textsc{rec}(\pi; \ell_{01})] - R_\textsc{rec}(\pi;
\ell_{01})$ in the proof of Theorem~\ref{thm:z_vs_z0_est}.  Combining
those bounds, along with the fact that $\pi(x) \in [0,1]$ for all
$x \in \mathcal{X}$, gives the result.

\qed

\subsection{Proof of Theorem~\ref{thm:policy_learn_dec}}

Define $\hat{V}(\pi)$ as the objective in
Equation~\eqref{eq:optimal_ai_rule} and $V(\pi) =
\E[\pi(X)U_z(X)]$. Following the proof of
Theorem~\ref{thm:policy_learn_rec}, notice that
\begin{align*}
  V(\hat{\pi}) - V(\pi^\ast)  & \leq 2 \sup_{\pi \in \Pi}| \hat{V}(\pi) - V(\pi)|\\
                              & \leq 2 \sup_{\pi \in \Pi} | \hat{V}(\pi) - \E[\hat{V}(\pi)]| + 2 \sup_{\pi \in \Pi} | \E[\hat{V}(\pi)] - V(\pi)|,
\end{align*}
because $\widehat{\pi}$ minimizes $\hat{V}(\pi)$. As in the proof of
Theorem~\ref{thm:policy_learn_rec}, $\hat{V}(\pi) - \E[\hat{V}(\pi)]$
is a mean-zero empirical process and since $A, D, Y$ are all binary,
the elements of $\hat{V}$ are bounded by
$\left(1 + \frac{2}{\eta}\right)(4 + 6 \ell_{01})$.\footnote{To see
  this, note that
  $\left|\widehat{\varphi}_{z1}(Z_i, X_i, D_i, A_i, Y_i)\right| \leq 1
  + \frac{2}{\eta}$, and similarly for the other components of
  $\hat{V}(\pi)$.}  Therefore, by \citet{wainwright_2019}, Theorem
4.2,
\[
  2 \sup_{\pi \in \Pi} | \hat{V}(\pi) - \E[\hat{V}(\pi)] \leq \left(1 + \frac{2}{\eta}\right)(4 + 6 \ell_{01}) \mathcal{R}_n(\Pi) + \frac{t}{\sqrt{n}},
\]
with probability at least $1 - \exp\left(-\frac{t^2}{2}\right)$.

It remains to control
$\sup_{\pi \in \Pi} | \E[\hat{V}(\pi)] - V(\pi)|$. To do so, notice
that we have bounded each of the components of
$\E[\hat{V}(\pi)] - V(\pi)$ in the proof of
Theorem~\ref{thm:partial_id_risk_est}.  Combining those bounds, along
with the fact that $\pi(x) \in [0,1]$ for all $x \in \mathcal{X}$,
gives the desired result.

\qed

\subsection{Proof of Theorem~\ref{thm:partial_Dstar}}

We first derive the sharp bounds on $\Pr(Y(0)=1\mid A)$.  We can
express this quantity in terms of $\theta_1$ and $\theta_0$:
\begin{eqnarray*}
\Pr(Y(0)=1\mid A=a) &=& \frac{\Pr(Y(0)=1,D=1, A=a)+\Pr(Y(0)=1,D=0, A=a)}{\Pr(A=a)}\\
&=&\frac{\theta_a+\Pr(Y=1,D=0, A=a)}{\Pr(A=a)}.
\end{eqnarray*}
From Lemma~\ref{cor:sharp_bounds}, we have the sharp bounds on
$\Pr(Y(0)=1\mid A=a)$:
\begin{eqnarray*}
  \Pr(Y(0)=1\mid A=a) &\geq&  \max_{z^\prime} \Pr(Y=1,D=0\mid A=a, Z=z^\prime)\\
  \Pr(Y(0)=1\mid A=a) &\leq&1-  \max_{z^\prime} \Pr(Y=0,D=0\mid A=a, Z=z^\prime).
\end{eqnarray*}
Following the similar procedure with $X$ in the conditioning set, we
can obtain the bounds on $\Pr(Y(0)=1\mid A=a,X)$: 
\begin{eqnarray*}
  \Pr(Y(0)=1\mid A=a,X) &\geq&  \max_{z^\prime} \Pr(Y=1,D=0\mid A=a, X,Z=z^\prime)\\
  \Pr(Y(0)=1\mid A=a,X)  &\leq&1-  \max_{z^\prime} \Pr(Y=0,D=0\mid A=a, X,Z=z^\prime).
\end{eqnarray*}
Observe that we can write the expression
of $R(\ell_{01};D^\ast)$ as follows:
\begin{eqnarray*}
R(\ell_{01}; D^\ast)&=& \E\left[\{1-f(A,X)\} \Pr(Y(0)= 1\mid A,X) + \ell_{01} f(A,X) \Pr(Y(0)= 0 \mid A,X)\right] \\
&=& \E\left[\ell_{01} \cdot f(A,X) + \{1-(1+\ell_{01})f(A,X)\} \Pr(Y(0)= 1\mid A,X)\right].
\end{eqnarray*}
Plugging the bounds on $\Pr(Y(0)=1\mid A=a,X)$ into the expression, we have the bounds on $R(\ell_{01};D^\ast)$: 
\begin{eqnarray*}
R(\ell_{01};D^\ast) &\geq& \E\Big[\ell_{01}\cdot f(A,X) 
  + \left\{1-(1+\ell_{01})f(A,X)\right\} \big[g_{f}(A,X)
  \max_{z^\prime} \Pr(Y=1,D=0\mid A, X,Z=z^\prime) \\
  & &\qquad +\{1-g_{f}(A,X)\}
  \{1-\max_{z^\prime} \Pr(Y=0,D=0\mid A, X,Z=z^\prime)\} \big]\\
R(\ell_{01};D^\ast) &\leq& \E\Big[\ell_{01}\cdot f(A,X) 
  + \left\{1-(1+\ell_{01})f(A,X)\right\} \big[g_{f}(A,X)
  \{1-\max_{z^\prime} \Pr(Y=0,D=0\mid A, X,Z=z^\prime)\} \\
  & &\qquad +\{1-g_{f}(A,X)\}
  \max_{z^\prime} \Pr(Y=1,D=0\mid A, X,Z=z^\prime) \big]\Big].
\end{eqnarray*}
where $g_{f}(a,x) = \1\{1-(1+\ell_{01})f(a,x)\geq0\}$.
\qed

\subsection{Proof of Theorem \ref{thm:bounds_each}}

The proof follows immediately from Lemma~\ref{cor:sharp_bounds-p}. 
\qed 

\newpage

\section{Additional Empirical Results}
\label{app:additional}

\begin{figure}[!h]
    \centering
    \includegraphics[width = \textwidth]{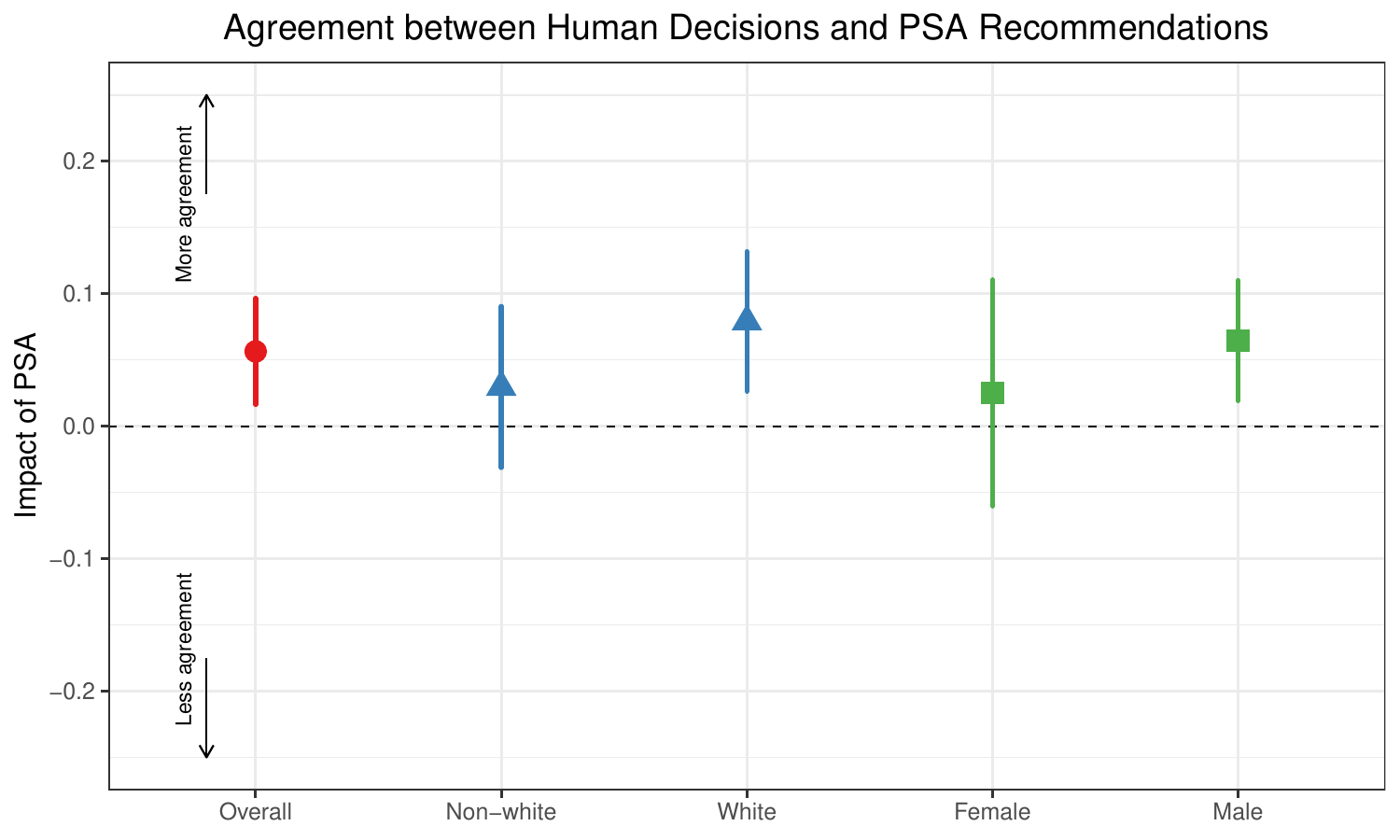}
    \caption{Subgroup Analysis of Estimated Impact of AI Recommendations 
    on Agreement between Human Decisions and AI Recommendations. 
    The figure shows the extent of agreement between judges and 
    AI recommendations when provided to the judges, compared to when it is not. 
    Each panel presents overall and subgroup-specific results using the difference in means estimates of an indicator $\bbone\{D_i = A_i\}$. 
    For each quantity of interest, we report a point estimate and its 
    corresponding 95\% confidence interval for the overall sample (red circle), 
    non-white and white subgroups (blue triangle), and female and male subgroups 
    (green square). The results show that judges agree with AI recommendations 
    more often, especially for white and male arrestees.}
      \label{fig:human_ai_disagreement}
\end{figure}

\begin{figure}[!h]
    \centering
    \includegraphics[width = \textwidth]{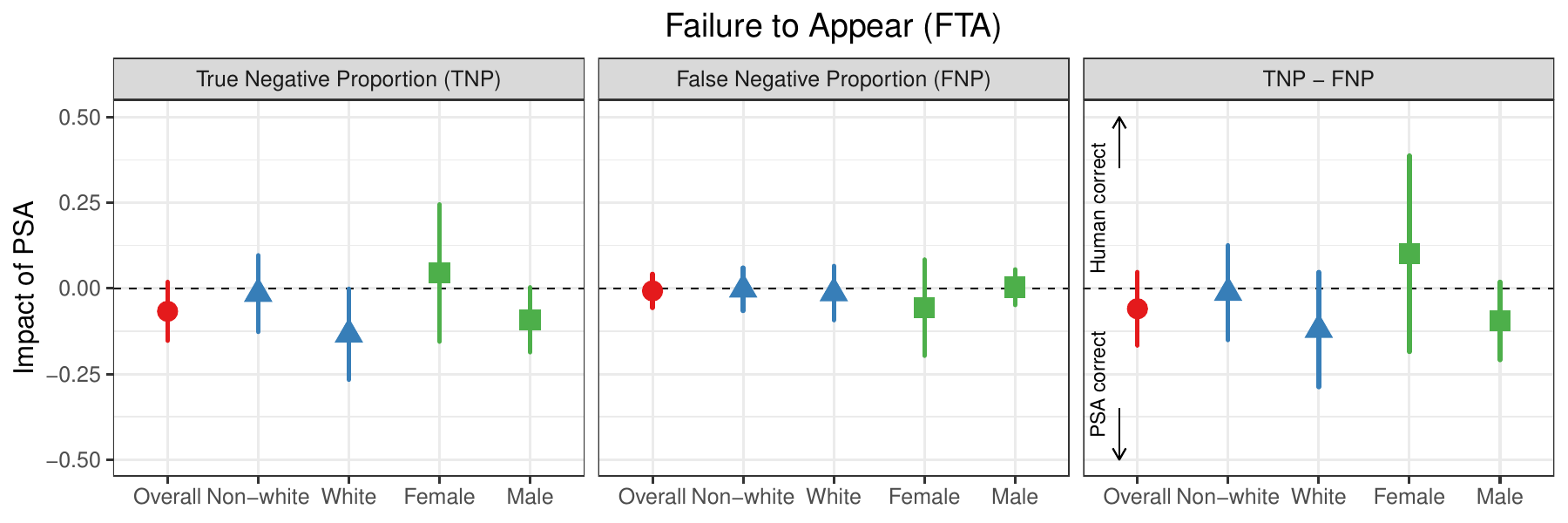}
    \includegraphics[width = \textwidth]{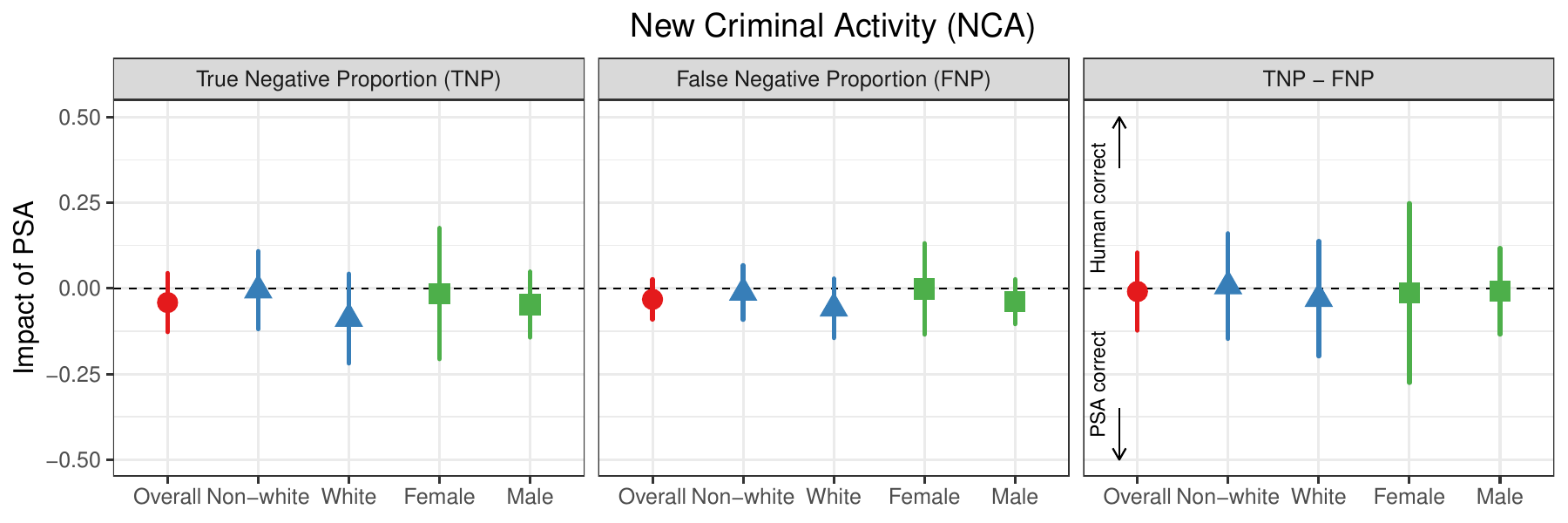}
    \includegraphics[width = \textwidth]{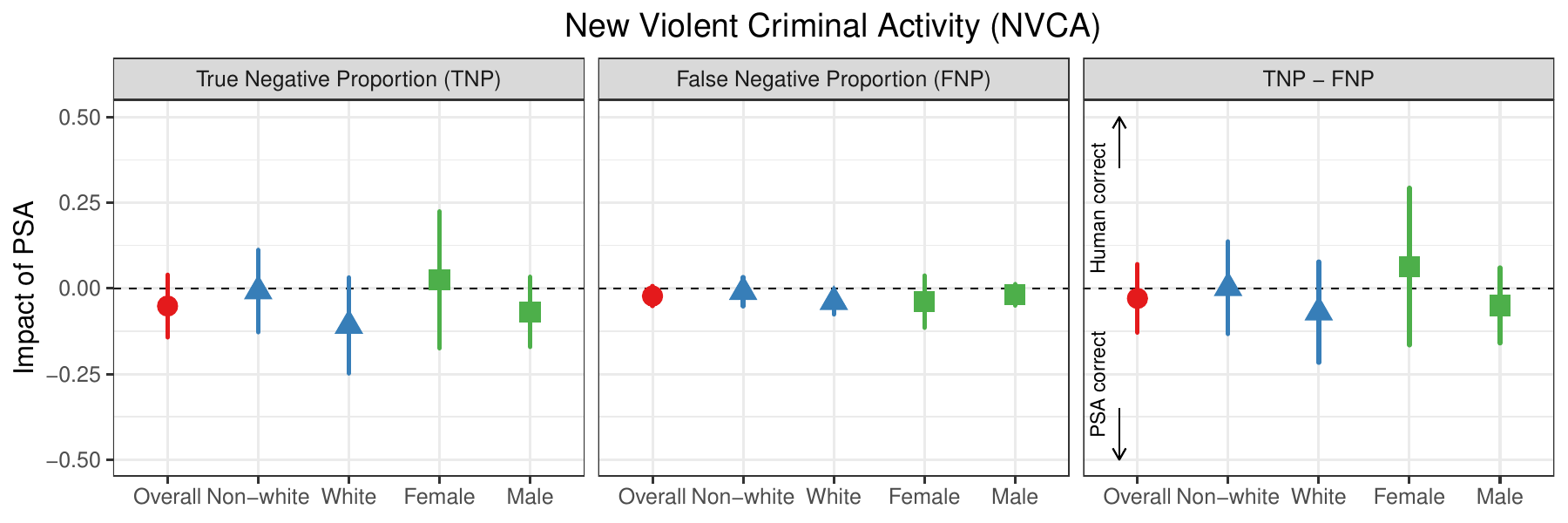}
    \caption{Subgroup Analysis of Estimated Impact of AI
      Recommendations on Human Decisions for the Cases where AI
      Recommends Cash Bail ($A = 1$). The figure shows how the human
      judge overrides the AI recommendation of cash bail in terms of
      true negative proportion (TNP), false negative proportion (FNP),
      and their differences. We adjust for the baseline disagreement
      between the human-alone and AI-alone systems by setting the
      human-alone system as the baseline. Each panel presents the
      overall and subgroup-specific results for a different outcome
      variable. For each quantity of interest, we report a point
      estimate and its corresponding 95\% confidence interval for the
      overall sample (red circle), non-white and white subgroups (blue
      triangle), and female and male subgroups (green square). The
      results shows no statistically significant evidence that the
      judge correctly overrides the AI recommendation of cash bail.}
      \label{fig:human_override_A1}
\end{figure}

\begin{figure}[!h]
    \centering
    \includegraphics[width = \textwidth]{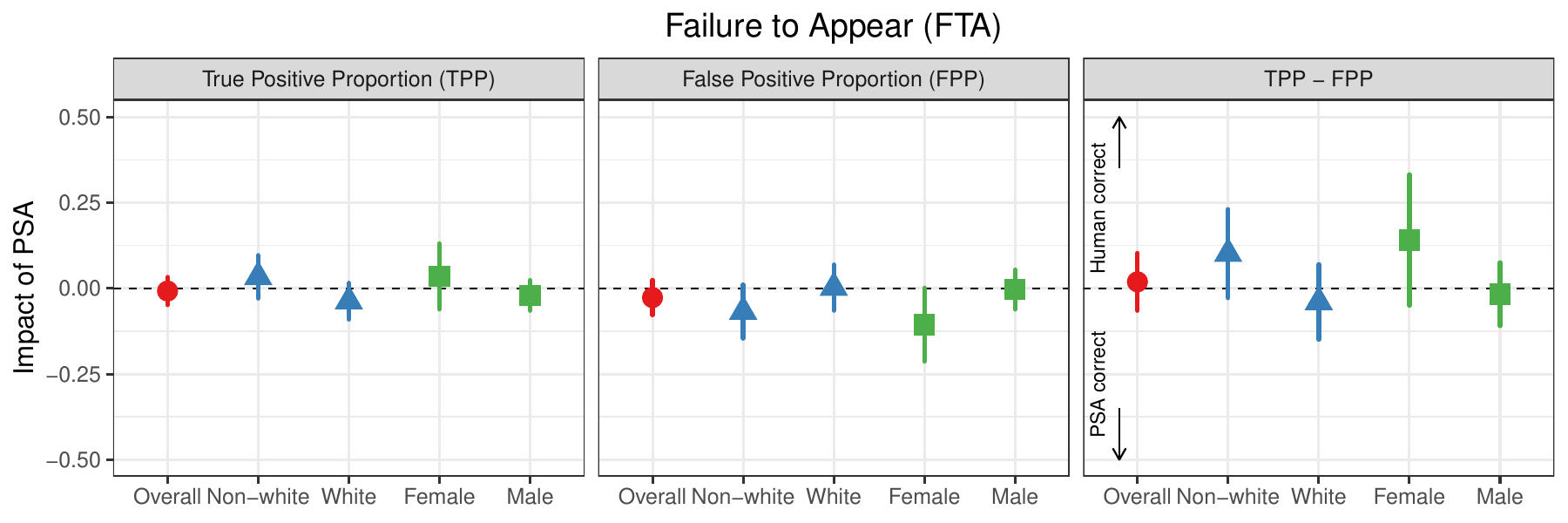}
    \includegraphics[width = \textwidth]{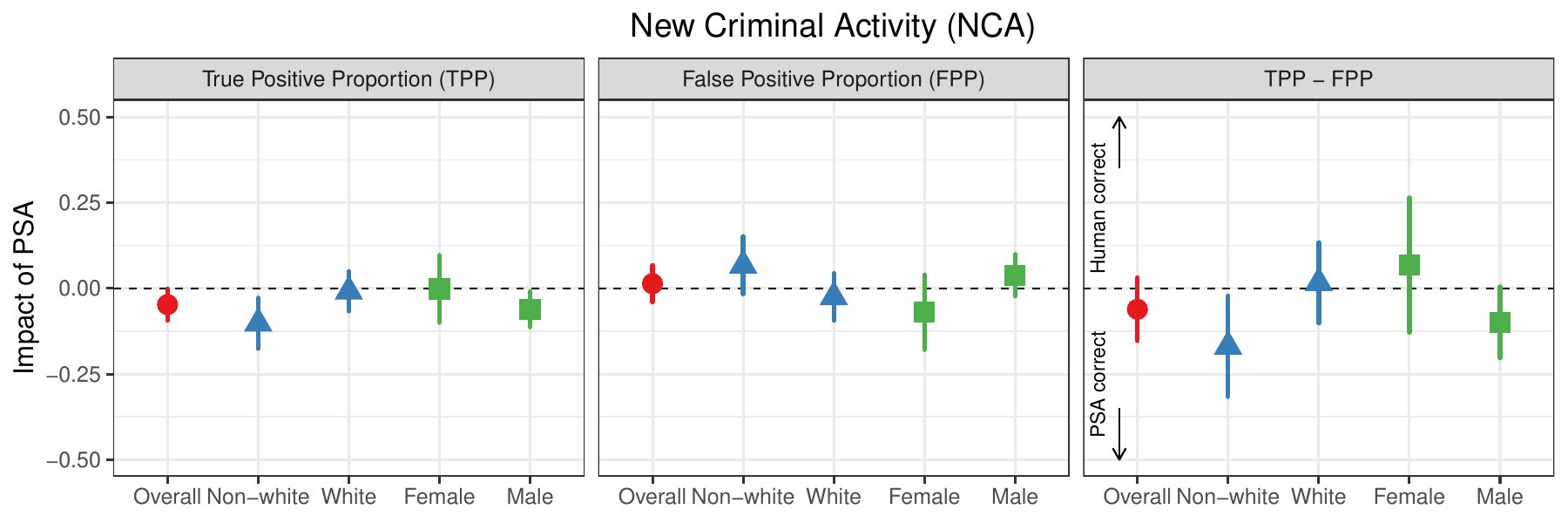}
    \includegraphics[width = \textwidth]{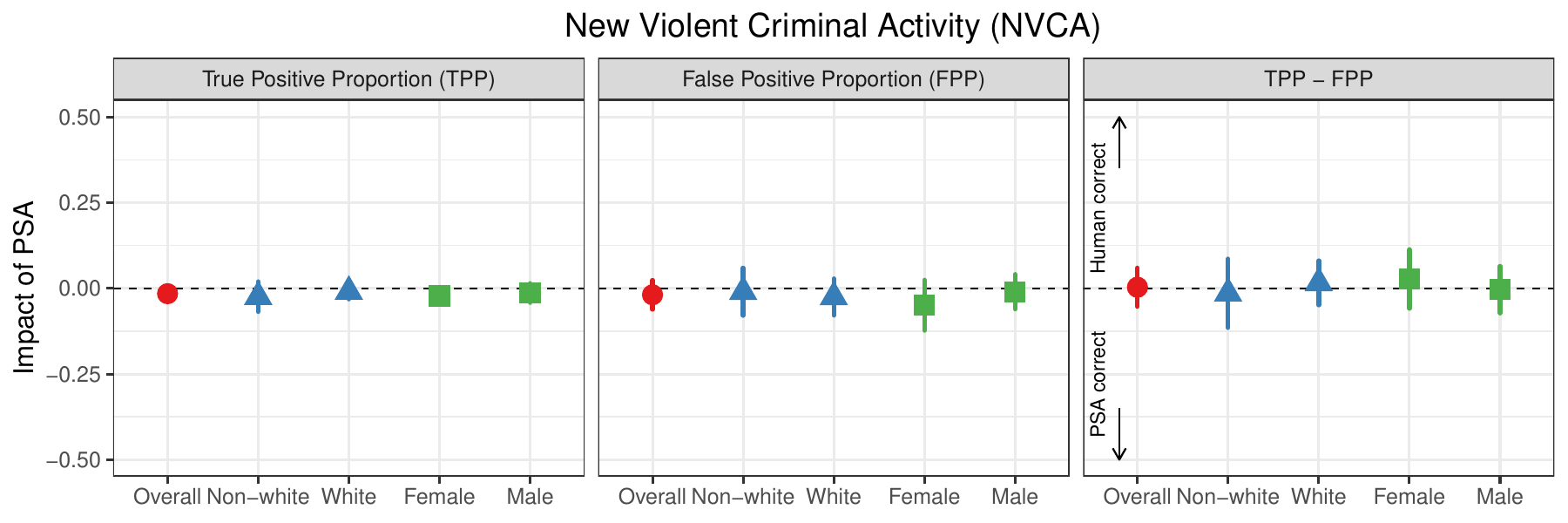}
    \caption{Subgroup Analysis of Estimated Impact of AI
      Recommendations on Human Decisions for the Cases where AI
      Recommends Signature Bond ($A = 0$). The figure shows how human
      judge overrides the AI recommendation of signature bond in terms
      of true positive proportion (TPP), false positive proportion
      (FPP), and their differences. We adjust for the baseline disagreement
      between the human-alone and AI-alone systems by setting the
      human-alone system as the baseline. Each panel presents the overall and
      subgroup-specific results for a different outcome variable. For
      each quantity of interest, we report a point estimate and its
      corresponding 95\% confidence interval for the overall sample
      (red circle), non-white and white subgroups (blue triangle), and
      female and male subgroups (green square). The results show no
      statistically significant evidence that the judge correctly
      overrides the AI recommendation of signature bond.}
      \label{fig:human_override_A0}
\end{figure}

\begin{figure}[!h]
    \centering
    \includegraphics[width = \textwidth]{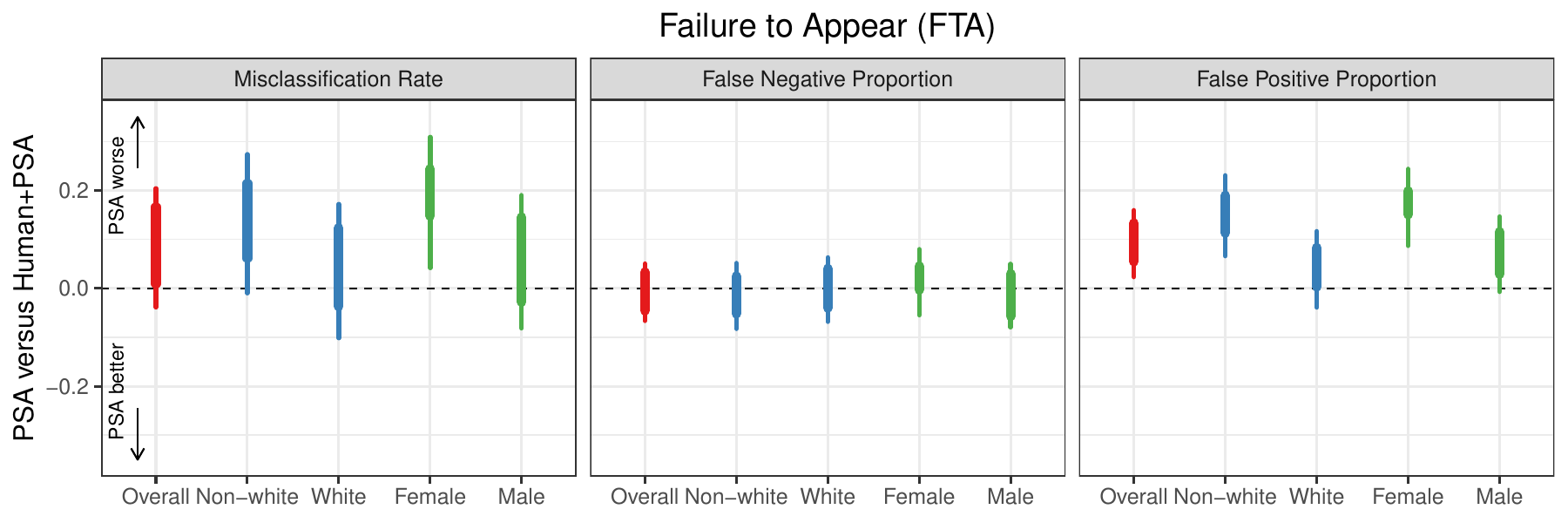}
    \includegraphics[width = \textwidth]{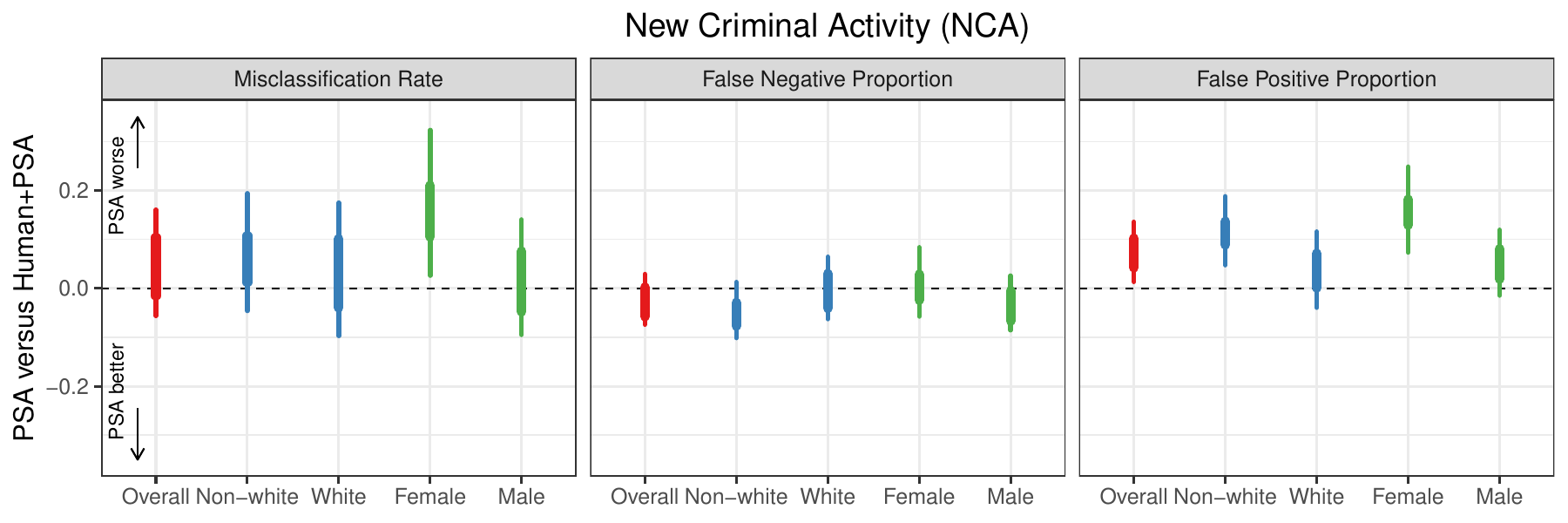}
    \includegraphics[width = \textwidth]{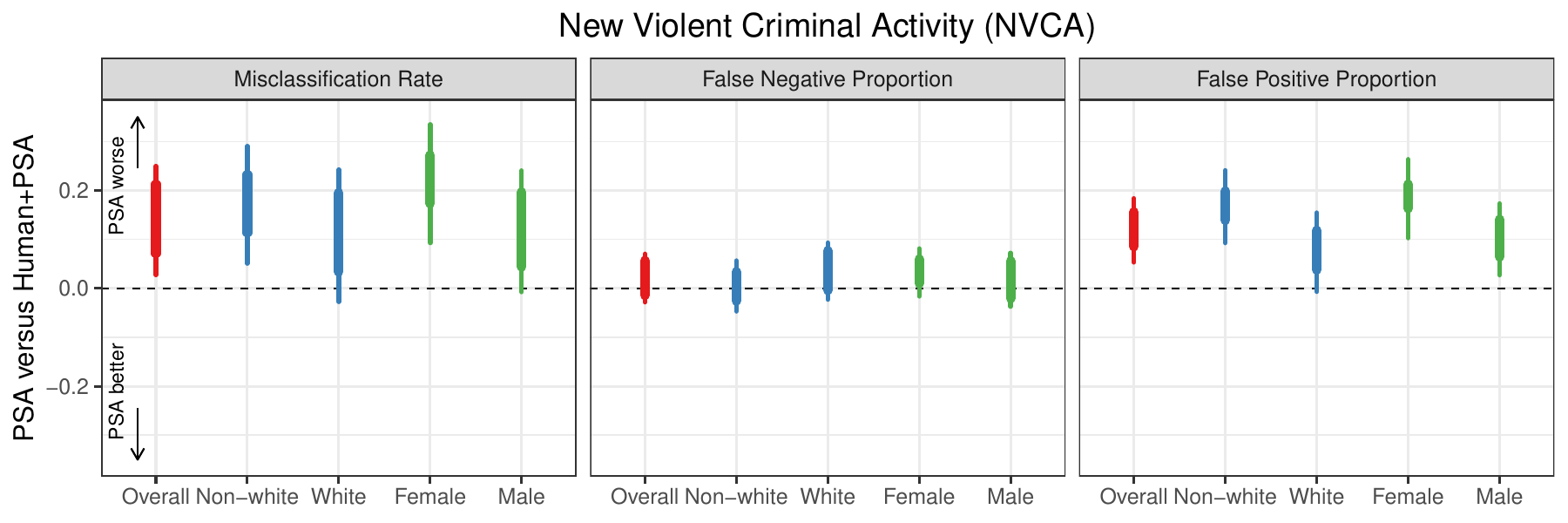}
    \caption{Estimated Bounds on Difference in Classification Ability
      between AI-alone and Human-with-AI Decision Making Systems. The
      figure shows misclassification rate, false negative proportion,
      and false positive proportion. Each panel presents the overall
      and subgroup-specific results for a different outcome variable.
      For each quantity of interest, we report estimated bounds (thick
      lines) and their corresponding 95\% confidence interval (thin
      lines) for the overall sample (red), non-white and white
      subgroups (blue), and female and male subgroups (green). The
      results indicate that AI-alone decisions are less accurate than
      human judge's decisions with AI recommendations in terms of the
      false positive proportion.}
    \label{fig:ai_humanAI_bounds}
\end{figure}

\begin{figure}[p]
    \centering
    \includegraphics[width = \textwidth]{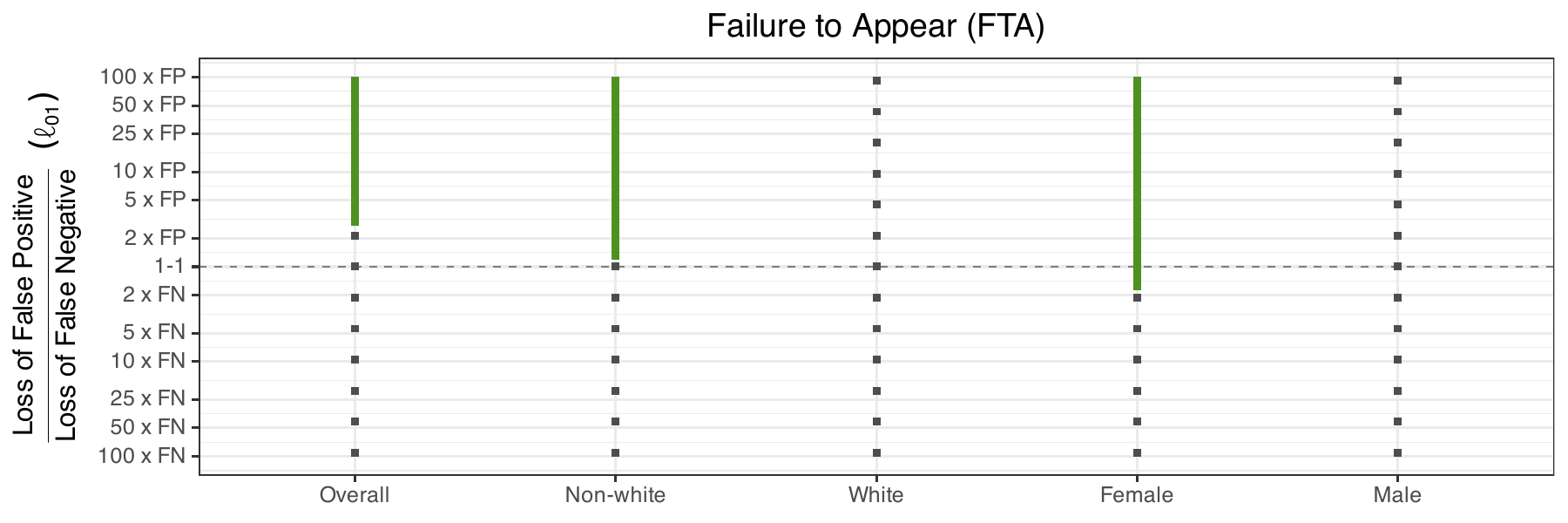}
    \includegraphics[width = \textwidth]{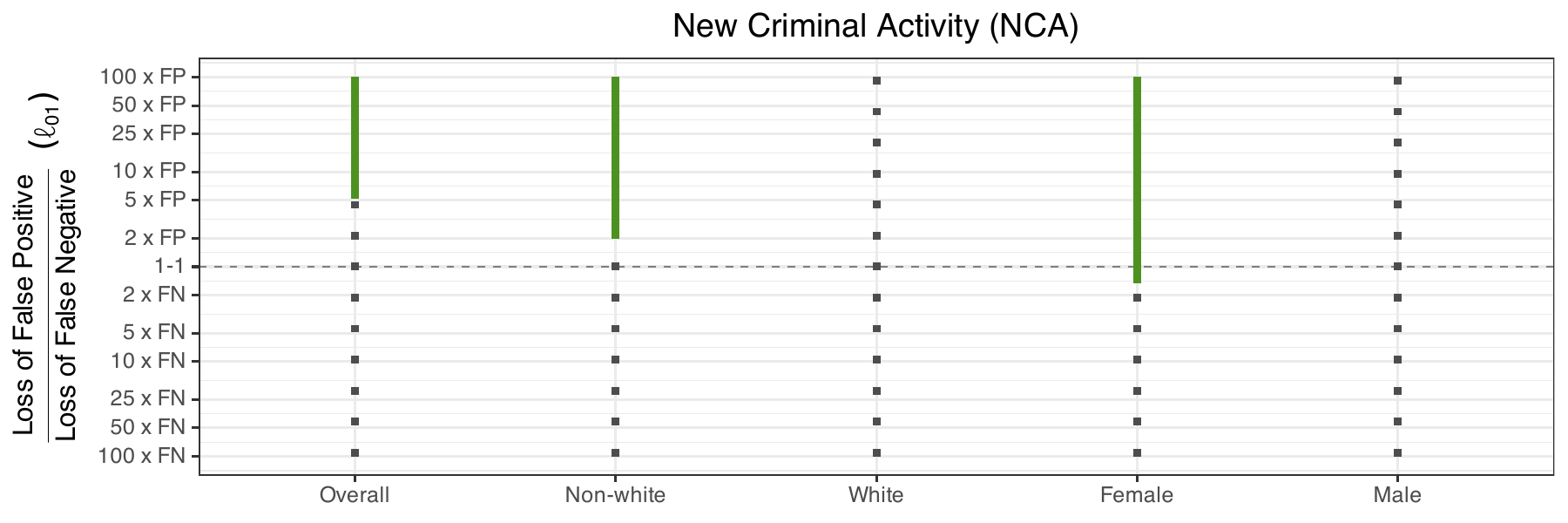}
    \includegraphics[width = \textwidth]{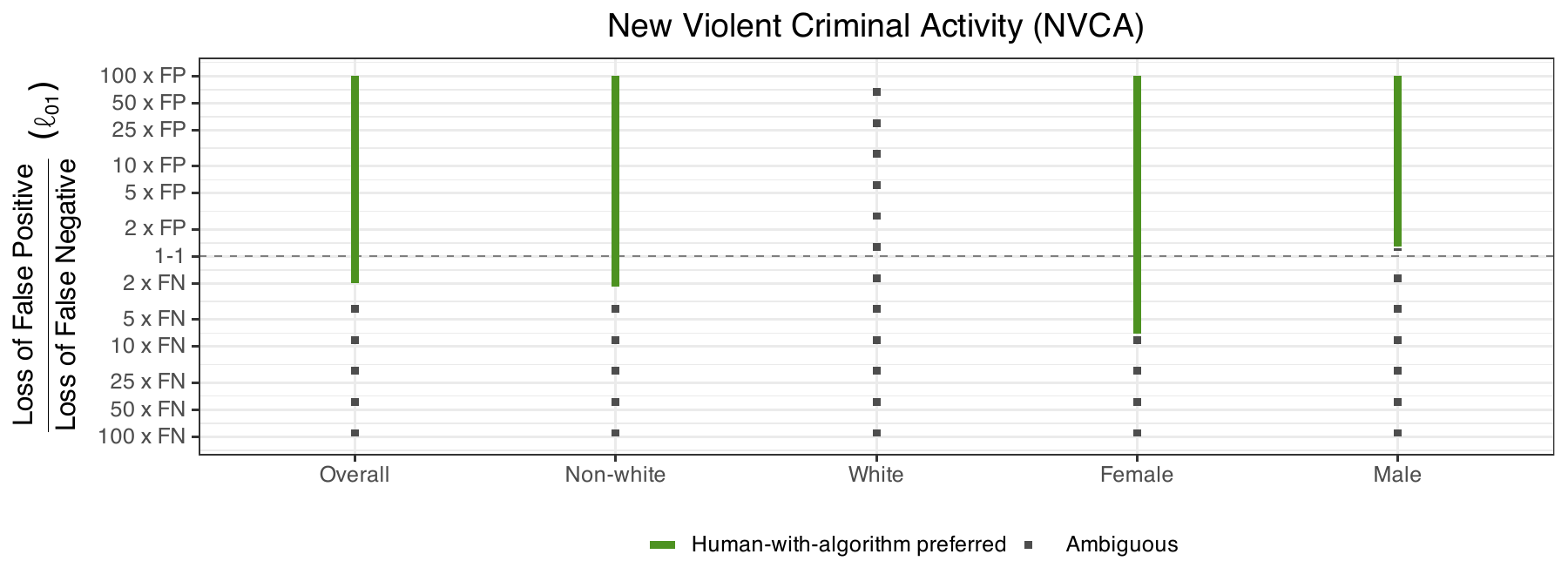}
    \caption{Estimated Preference for Human-with-AI over AI Decision-Making Systems. 
      The figure illustrates the range of the ratio of the loss between false positives and false
      negatives, $\ell_{01}$, for which one decision-making system is
      preferable over the other. A greater value of the ratio
      $\ell_{01}$ implies a greater loss of false positive relative to
      that of false negative. Each panel displays the overall and
      subgroup-specific results for different outcome variables. For
      each quantity of interest, we show the range of $\ell_{01}$ that
      corresponds to the preferred decision-making system; human-with-AI (green lines), and ambiguous (dotted
      lines). The results suggest that the human-with-AI system is
      preferred over the AI-alone system when the loss of false
      positive is about the same as or greater than that of false
      negative.
      The AI-alone system is never preferred within the specified range of $\ell_{01}$.}
    \label{fig:pref_humanAI}
\end{figure}

\begin{table}[ht]

\centering
\begin{tabular}[t]{>{}lrrrr}
\toprule
\multicolumn{1}{c}{ } & \multicolumn{2}{c}{Whether to provide} & \multicolumn{2}{c}{Whether to follow} \\
\cmidrule(l{3pt}r{3pt}){2-3} \cmidrule(l{3pt}r{3pt}){4-5}
Outcome & Increasing & Decreasing & Increasing & Decreasing\\
\midrule
FTA & -0.0012 & \cellcolor{lightgray}-0.0318 & 0 & \cellcolor{lightgray}-0.0014\\
NCA & \cellcolor{lightgray}-0.0101 & -0.0085 & \cellcolor{lightgray}-0.0035 & 0\\
NVCA & -0.0008 & \cellcolor{lightgray}-0.0225 & \cellcolor{lightgray}-0.0002 & 0\\
\bottomrule
\end{tabular}
\caption{Estimated Values of the Empirical Risk Minimization Problem under the Optimal Policy. 
The table presents the estimated values of the empirical risk minimization problem as described in Equation~\ref{eq:optimal_recommendation_rule} for the second and third columns, and in Equation~\ref{eq:optimal_ai_rule} for the fourth and fifth columns.
The second and fourth columns correspond to the results regarding policy class with an increasing monotonicity constraint, while the third and fifth columns represent those with a decreasing monotonicity constraint.
For instance, for the NCA as an outcome, the optimal policy regarding whether to provide PSA recommendations with the increasing monotonicity constraint results in a $0.0101$ decrease in the difference in misclassification rate relative to not providing PSA recommendations.
}
\label{tbl:optimal}
\end{table}

\end{document}